\documentclass[table,xcdraw]{template/arxiv_v2/ai2style/ai2}

\DeclareTextFontCommand{\textbf}{\rmfamily\bfseries}
\renewcommand{\title}[1]{%
  \newcommand{\titlelist}{{\Huge\rmfamily\bfseries #1}}%
}

\titleformat*{\section}{\Large\rmfamily\bfseries}
\titleformat*{\subsection}{\large\rmfamily\bfseries}
\titleformat*{\subsubsection}{\slightlylarger\rmfamily\bfseries}
\titleformat*{\paragraph}{\slightlysmaller\rmfamily\bfseries}

\usepackage{microtype}
\usepackage{hyperref}
\usepackage{url}
\usepackage{booktabs, tabularx} 
\usepackage{graphicx}
\usepackage{lineno}
\usepackage{enumitem}
\usepackage{listings} 

\definecolor{ darkblue}{rgb}{0, 0, 0.5}
\hypersetup{colorlinks=true, citecolor=darkblue, linkcolor=darkblue, urlcolor=darkblue}

\newcommand{\modelfull}{WorldOracle\xspace}
\newcommand{\metricsv}{WorldOracle-SV\xspace}

\newcommand{\physical}{physical state correctness\xspace}
\newcommand{\keyarea}{key interaction area fidelity\xspace}
\newcommand{\multiview}{multi-view action coherence\xspace}

\newcommand{\spatial}{interaction spatial alignment\xspace}

\definecolor{darkgreen}{rgb}{0.29, 0.64, 0.35} 
\newcommand{\cmark}{\textcolor{green}{\ding{51}}} 
\newcommand{\xmark}{\textcolor{red}{\ding{55}}} 

\renewcommand{\paragraph}[1]{\vspace{0.1cm}\noindent\textbf{#1}}

\definecolor{MyDarkBlue}{rgb}{0,0.08,1}
\definecolor{MyAqua}{rgb}{0,0.7,0.7}
\definecolor{MyDarkGreen}{rgb}{0.02,0.6,0.02}
\definecolor{MyDarkRed}{rgb}{0.8,0.02,0.02}
\definecolor{MyDarkOrange}{rgb}{0.40,0.2,0.02}
\definecolor{MyPurple}{RGB}{111,0,255}
\definecolor{MyRed}{rgb}{1.0,0.0,0.0}
\definecolor{MyGold}{rgb}{0.75,0.6,0.12}
\definecolor{MyDarkgray}{rgb}{0.66, 0.66, 0.66}

\definecolor{Cardinal}{rgb}{0.549,0.082,0.082}

\usepackage{fontawesome5}
\usepackage{mathtools}
\usepackage{amssymb}
\usepackage{multirow}
\usepackage{bigdelim}
\usepackage{todonotes}
\usepackage{longtable}
\usepackage{tabularray}
\usepackage{wrapfig}
\usepackage[most]{tcolorbox}
\usepackage{url}
\usepackage{xspace}
\usepackage[absolute]{textpos} 

\DeclareMathOperator*{\argmax}{arg\,max}

\usepackage[utf8]{inputenc} 
\usepackage{url}            
\usepackage{booktabs}       
\usepackage{amsfonts}       
\usepackage{nicefrac}       
\usepackage{microtype}      
\usepackage[table,xcdraw]{xcolor}         
\usepackage{amsmath}
\usepackage[most]{tcolorbox}
\usepackage{csquotes}
\usepackage{tablefootnote}
\usepackage{ragged2e}
\usepackage{graphicx}
\usepackage{arydshln}
\usepackage{wrapfig}
\usepackage{enumitem}

\usepackage{bm,placeins,hyphenat}
\definecolor{ourcolor}{RGB}{230,240,245}

\definecolor{baselinecolor}{gray}{.9}

\usepackage{multirow}
\usepackage{graphicx}
\usepackage{listings}
\lstdefinelanguage{Prompt}{
  morestring=[b]",
}
\definecolor{codebg}{RGB}{245,245,245}
\usepackage{template/arxiv_v2/algorithm}
\usepackage{template/arxiv_v2/algorithmic}
\definecolor{codekw}{rgb}{0.00,0.00,0.60}
\definecolor{codestr}{rgb}{0.58,0.00,0.13}
\definecolor{codecmt}{rgb}{0.30,0.50,0.30}
\definecolor{codenum}{rgb}{0.55,0.55,0.55}
\lstdefinestyle{pystyle}{
  language=Python,
  backgroundcolor=\color{codebg},
  basicstyle=\ttfamily\footnotesize,
  keywordstyle=\color{codekw}\bfseries,
  commentstyle=\color{codecmt}\itshape,
  stringstyle=\color{codestr},
  numberstyle=\tiny\color{codenum},
  showstringspaces=false,
  breaklines=true,
  breakatwhitespace=true,
  numbers=left,
  numbersep=6pt,
  frame=none,
  framexleftmargin=4pt,
  xleftmargin=12pt,
  columns=fullflexible,
  keepspaces=true,
  belowskip=4pt, aboveskip=6pt,
  emph={self,True,False,None},
  emphstyle=\color{codekw},
}

\usepackage{multirow}
\usepackage{xspace}
\usepackage{adjustbox}
\usepackage{pifont}
\usepackage{caption}
\usepackage{makecell}
\usepackage{subcaption}
\usepackage{bold-extra}
\usepackage{url}

\usepackage{pgf-pie}

\usepackage{hyperref}
\definecolor{linkcolor}{RGB}{0, 0, 128}
\hypersetup{
     colorlinks   = true,
     citecolor    = linkcolor,
     linkcolor    = linkcolor,
     urlcolor     = linkcolor,
}
\usepackage{pifont}
\usepackage{listings}

\setlist[itemize]{leftmargin=*,itemsep=0em,parsep=0.3em,topsep=0.3em}

\ifdefined\DeclareUnicodeCharacter
\DeclareUnicodeCharacter{2212}{\ensuremath{-}}
\fi

\definecolor{maroon}{HTML}{F26035}
\definecolor{yellow}{HTML}{FDBC42}
\definecolor{lavender}{HTML}{734f96}
\definecolor{darkergrey}{HTML}{444444}
\definecolor{midgrey}{HTML}{e6eded}
\definecolor{ai2pink}{HTML}{f0529c}
\definecolor{ai2midpink}{HTML}{fad3e5}
\definecolor{ai2lightpink}{HTML}{fbecf3}
\definecolor{ai2midwhite}{HTML}{f2e5d9}
\definecolor{ai2offwhite}{HTML}{fbf4ee}
\definecolor{ai2green}{HTML}{0fcb8c}
\definecolor{ai2lightgreen}{HTML}{e7f9f3}
\definecolor{ai2darkgreen}{HTML}{105257}
\definecolor{ai2purple}{HTML}{B932EB}
\definecolor{ai2lightpurple}{HTML}{f7e8fc}
\definecolor{neutralEight}{HTML}{343434}
\definecolor{neutralFive}{HTML}{838383}
\definecolor{neutralThree}{HTML}{bebebe}
\definecolor{neutralOne}{HTML}{dedede}
\definecolor{lightgrey}{HTML}{fafcfc}
\definecolor{plum}{rgb}{0.56,0.27,0.52}

\usepackage{tikz}

\definecolor{maroon}{HTML}{F26035}
\definecolor{yellow}{HTML}{FDBC42}
\definecolor{darkred}{RGB}{156, 39, 33}
\definecolor{darkblue}{RGB}{31, 90, 153}
\definecolor{forestgreen}{rgb}{0.13, 0.55, 0.13}
\definecolor{brickred}{rgb}{0.8, 0.25, 0.33}
\definecolor{olmoDarkBlue}{HTML}{012e59}
\definecolor{olmoBlue}{HTML}{265ed4}
\definecolor{olmoLightBlue}{HTML}{012e59}
\definecolor{olmoTeal}{HTML}{00d5ff}
\definecolor{olmoYellow}{HTML}{ffbb00}
\definecolor{olmoOrange}{HTML}{ff9100}

\renewcommand{\paragraph}[1]{\vspace{0.5mm}\noindent\textbf{#1}}

\definecolor{molmocolor}{RGB}{240, 82, 156}
\definecolor{tablegray}{RGB}{223, 242, 252}
\definecolor{tablegreen}{RGB}{15, 203, 150}
\definecolor{tableyellow}{RGB}{250, 242, 233}
\definecolor{tableblue}{RGB}{240, 82, 156}
\definecolor{darkpink}{RGB}{139, 14, 98}

\usepackage{setspace}

\newcolumntype{L}[1]{>{\raggedright\let\newline\\\arraybackslash\hspace{0pt}}m{#1}}
\newcolumntype{C}[1]{>{\centering\let\newline\\\arraybackslash\hspace{0pt}}m{#1}}
\newcolumntype{R}[1]{>{\raggedleft\let\newline\\\arraybackslash\hspace{0pt}}m{#1}}
\newcolumntype{P}[1]{>{\centering\let\newline\\\arraybackslash\columncolor{ai2lightpink}}m{#1}}
\makeatletter
\DeclareRobustCommand\onedot{\futurelet\@let@token\@onedot}
\def\@onedot{\ifx\@let@token.\else.\null\fi\xspace}

\makeatother

\definecolor{titlegA}{HTML}{3FA838}  
\definecolor{titlegB}{HTML}{6FB833}  
\definecolor{titlegC}{HTML}{9FC72E}  
\definecolor{titlegD}{HTML}{CFD629}  
\definecolor{titlegE}{HTML}{F0CB1F}  

\definecolor{abstractblue}{RGB}{240,246,254}

\usepackage[most]{tcolorbox}
\tcbuselibrary{listings,breakable}
\usepackage{listings}

\lstdefinestyle{promptstyle}{
  basicstyle=\ttfamily\scriptsize,
  breaklines=true,
  breakatwhitespace=true,
  columns=fullflexible,
  keepspaces=true,
  showstringspaces=false,
  escapeinside={(*@}{@*)},
  xleftmargin=0pt,
  xrightmargin=0pt,
  breakindent=0pt,
  breakautoindent=false,
  resetmargins=true
}

\newtcblisting{promptbox}[1]{%
  enhanced,
  listing only,
  breakable,
  colback=white,
  colframe=gray!50,
  boxrule=0.5pt,
  arc=2pt,
  left=3pt,
  right=3pt,
  top=6pt,
  bottom=6pt,
  title={\textbf{#1}},
  listing options={style=promptstyle}
}

\title{%
\fontsize{16pt}{19pt}\selectfont
\raisebox{-0.18\height}{%
    \includegraphics[height=1.35em]{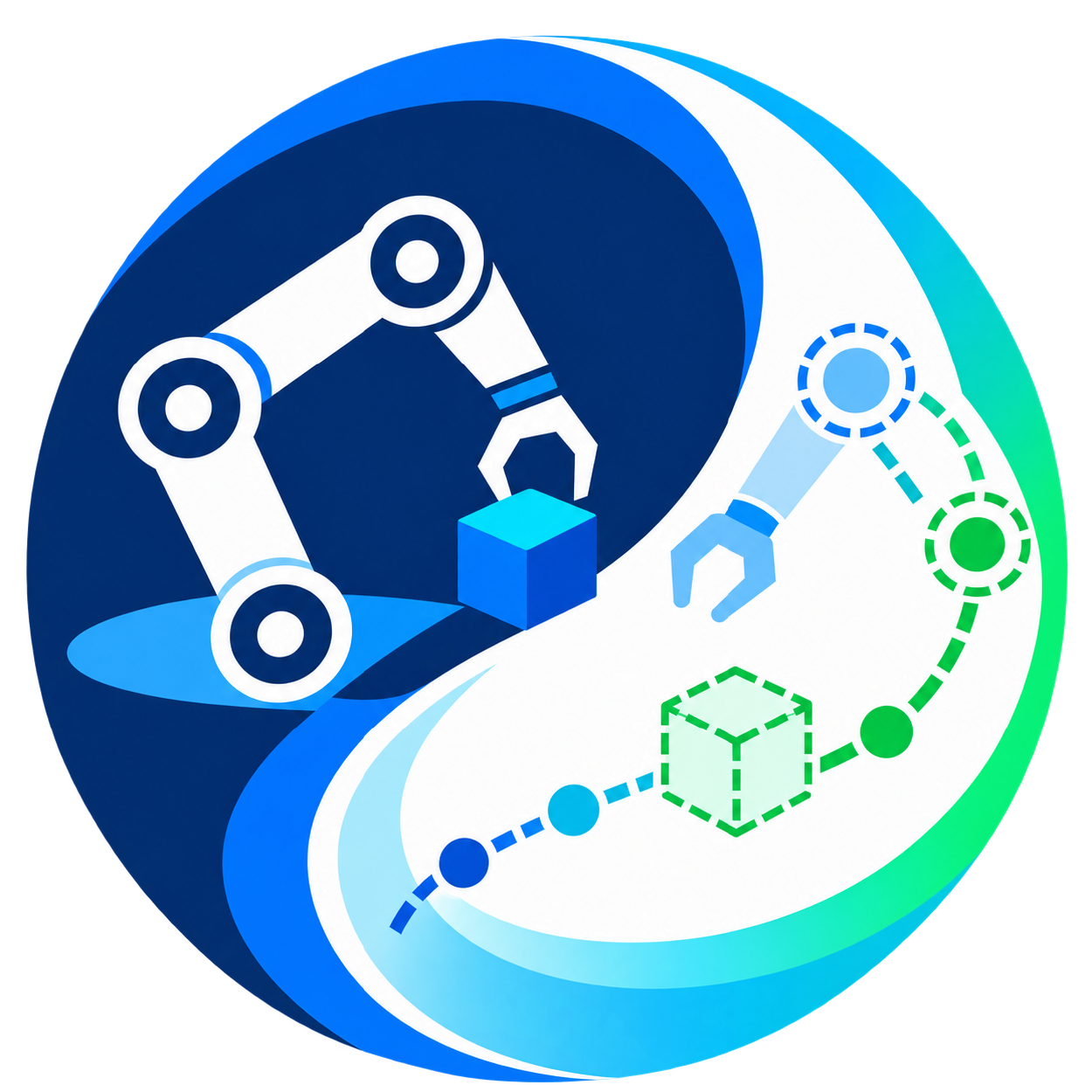}%
}\hspace{0.35em}%
\textbf{WorldOracle: Benchmarking Action-Conditioned}\\
\textbf{World Models against Real-World Executions}
}

\authorOne[1,3]{Haoyi Duan}
\authorOne[1]{Liliang Chen}
\authorOne[1]{Nan Wang}
\authorOne[1]{Yang Zhang}
\authorOne[2]{Yuhang Lu}
\authorTwo[2]{Xiaotong Chen}
\authorTwo[2,5]{Wei-Cheng Tseng}
\authorTwo[2]{Shenyuan Gao}
\authorTwo[2]{Zhaoshuo Li}
\authorTwo[2]{Jingyi Jin}
\authorThree[2]{Pooya Jannaty}
\authorThree[3]{Dhruv Shah}
\authorThree[2,4]{Yuke Zhu}
\authorThree[2]{Ming-Yu Liu}
\authorThree[1]{Maoqing Yao}
\authorThree[1]{Guanghui Ren}

\affiliation[1]{AgiBot}
\affiliation[2]{NVIDIA}
\affiliation[3]{Princeton University}
\affiliation[4]{University of Texas at Austin}
\affiliation[5]{University of Toronto}

\newcommand{\institutionlogos}{%
  \makebox[\textwidth]{%
    \raisebox{-0.15\height}{\includegraphics[height=0.9cm]{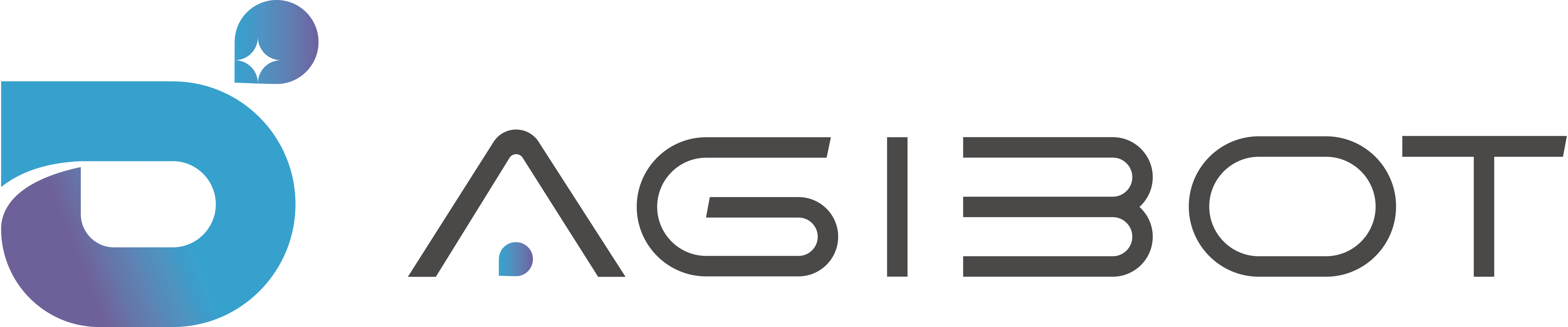}}%
    \hfill
    \raisebox{-0.15\height}{\includegraphics[height=0.9cm]{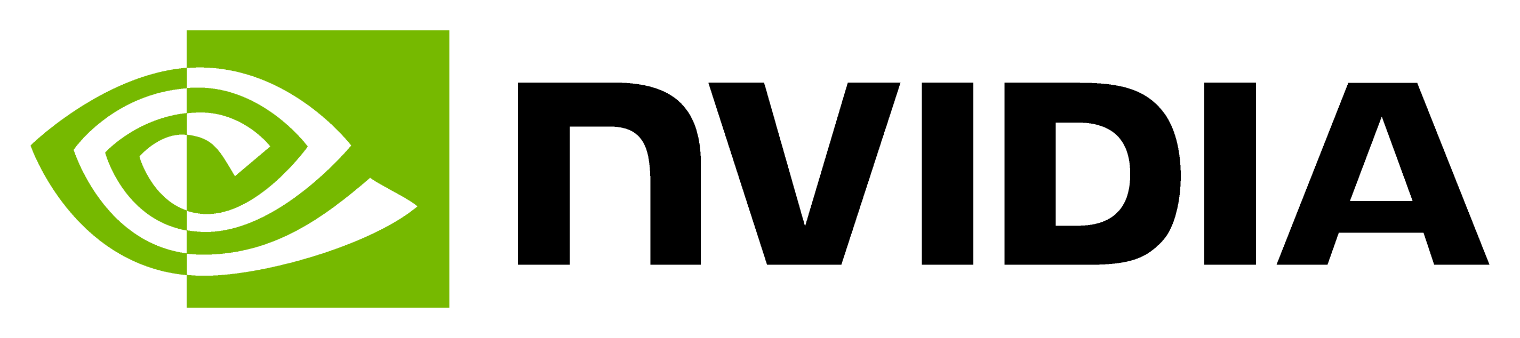}}%
  }%
}

\newlength{\headrulegap}
\renewcommand{\headrule}{%
  \vskip\headrulegap%
  \hrule height\headrulewidth width\headwidth%
  \vskip-\headrulewidth%
}
\fancypagestyle{firstpage}{%
  \fancyhf{}%
  \fancyhead[C]{\institutionlogos}%
  \renewcommand{\headrulewidth}{1.4pt}%
  \renewcommand{\footrulewidth}{0pt}%
}
\fancypagestyle{withtitle}{%
  \fancyhf{}%
  \fancyhead[C]{\small\sffamily WorldOracle: Benchmarking Action-Conditioned World Models Against Real-World Executions}%
  \fancyfoot[C]{\small\thepage}%
  \renewcommand{\headrulewidth}{1.2pt}%
  \renewcommand{\footrulewidth}{0pt}%
}
\definecolor{baselinecolor}{gray}{.9}

\usepackage[utf8]{inputenc}
\usepackage[most]{tcolorbox}
\tcbuselibrary{listings,breakable}

\definecolor{molmocolor}{RGB}{240, 82, 156}
\definecolor{tablegray}{RGB}{223, 242, 252}
\definecolor{tablegreen}{RGB}{15, 203, 150}
\definecolor{tableyellow}{RGB}{240,246,254}
\definecolor{tableblue}{RGB}{240, 82, 156}
\definecolor{darkpink}{RGB}{139, 14, 98}
\definecolor{baselinecolor}{gray}{.9}

\newlength{\teaserurlgap}
\newlength{\teaserbottomgap}
\newcommand{\projectdataurl}{https://huggingface.co/WorldOracle}
\newcommand{\projectcodeurl}{https://github.com/WorldOracle-benchmark/WorldOracle}
\newcommand{\projectleaderboardurl}{https://worldoracle-benchmark.github.io/leaderboard.html}
\newcommand{\projectwebsiteurl}{https://worldoracle-benchmark.github.io}

\newcommand{\projectresource}[3]{%
  \noindent
  \makebox[1.5em][c]{\textcolor{ai2accent}{#1}}%
  \textbf{#2:}\;
  \href{#3}{\nolinkurl{#3}}\par
}

\renewcommand{\teaserfigure}{%
  \begingroup
    \centering

    \includegraphics[width=\textwidth]{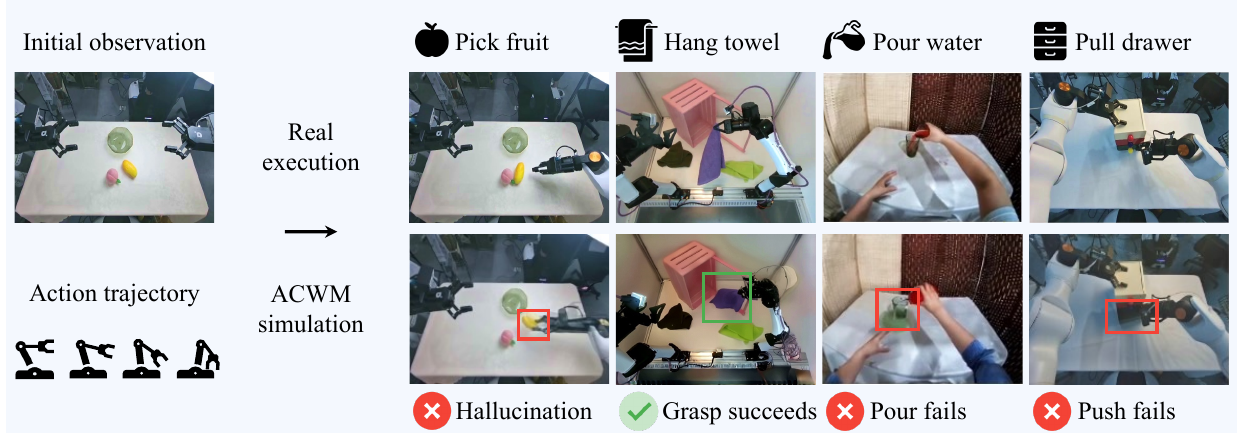}\par

    \begin{minipage}{0.96\textwidth}
      \captionsetup{hypcap=false}%
      \captionof{figure}{%
        \textbf{Does the predicted future match reality?} Given the same initial observation and action trajectory, an action-conditioned world model (ACWM) can produce a visually plausible rollout that does not match what actually happens in the real world. In the ``pick fruit'' example, the robot fails to grasp the fruit, while the model hallucinates a successful grasp.%
      }
      \label{fig:motivation}
    \end{minipage}

  \endgroup

  \vspace{0.5em}
  \begin{center}
    \begin{minipage}{0.9\textwidth}
      \small
      \setlength{\parskip}{3pt}

      \projectresource{\faDatabase}
        {Data}
        {\projectdataurl}

      \projectresource{\faGithub}
        {Code}
        {\projectcodeurl}

      \projectresource{\faTrophy}
        {Leaderboard}
        {\projectleaderboardurl}

      \projectresource{\faGlobe}
        {Website}
        {\projectwebsiteurl}

    \end{minipage}
  \end{center}
  \vspace*{\teaserbottomgap}

  \begin{center}
    \begin{tcolorbox}[
      enhanced,
      frame hidden,
      width=0.9\textwidth,
      left=0.5cm,
      right=0.5cm,
      top=0.5cm,
      bottom=0.5cm,
      arc=3.5pt,
      colback=abstractblue,
      before skip=0pt
    ]
      {\Large\sffamily\bfseries Abstract\par}
      \vspace{0.3cm}
      We introduce \modelfull, a benchmark specifically designed to evaluate action-conditioned world models (ACWMs) against paired real-world executions. Given the same initial observation and low-level action sequence, \modelfull asks whether a predicted rollout reproduces what happens in the real world, including physical consequences and task-state transitions. We evaluate models along four complementary dimensions: \physical, \spatial, \keyarea, and \multiview. To support this evaluation, we curate real-world training and evaluation suites that cover different interaction dynamics and embodiments. The evaluation suite contains long, multi-segment episodes with both successful and failed executions, enabling evaluation across varied action outcomes and behavior distributions. Experiments on seven representative ACWMs reveal distinct model strengths and failure modes. \modelfull provides a systematic foundation for measuring and advancing action-conditioned world modeling for manipulation. We release the datasets, evaluation code, and leaderboard: \url{https://worldoracle-benchmark.github.io}.

    \end{tcolorbox}
  \end{center}
}

\lstdefinestyle{datapromptstyle}{
 style=promptstyle,
 literate=
 {×}{{$\times$}}1 {â}{{\^a}}1 {é}{{\'e}}1 {í}{{\'i}}1
 {–}{{-}}1 {—}{{---}}3 {’}{{'}}1 {“}{{"}}1 {”}{{"}}1
 {…}{{...}}3 {→}{{$\rightarrow$}}2 {↔}{{$\leftrightarrow$}}2
 {★}{{*}}1 {⛔}{{[EXCLUDE]}}9 {✅}{{[PASS]}}6 {❌}{{[REJECT]}}8
 {王者荣耀}{{Wangzhe Rongyao}}{14}
 {📂}{{[CATEGORIES]}}{12} {📐}{{[FORMAT]}}8 {🔑}{{[KEY]}}5
 {🚨}{{[IMPORTANT]}}{11}
}
\newtcblisting{datapromptbox}[1][]{
 listing only, breakable,
 colback=gray!3, colframe=gray!60, boxrule=0.4pt, arc=1pt,
 left=3pt, right=3pt, top=4pt, bottom=4pt,
 title={#1}, listing options={style=datapromptstyle}
}

\hypersetup{
  pdftitle={WorldOracle: Benchmarking Action-Conditioned World Models against Real-World Executions},
  pdfauthor={Haoyi Duan, Liliang Chen, Nan Wang, Yang Zhang, Yuhang Lu, Xiaotong Chen, Wei-Cheng Tseng, Shenyuan Gao, Zhaoshuo Li, Jingyi Jin, Pooya Jannaty, Dhruv Shah, Yuke Zhu, Ming-Yu Liu, Maoqing Yao, Guanghui Ren}
}

\begin{document}

\maketitle
\thispagestyle{firstpage}  

\section{Introduction}

Recent advances in video generation~\citep{nvidia2025cosmos,openai2024sora,deepmind2025veo3,wan2025wan,yang2025cogvideox} have opened a new path toward learned world simulation for robotic manipulation. Trained on large-scale interaction data, these models can capture a wide range of scenes, objects, and contact-rich behaviors that are difficult to model with hand-crafted simulators. Recent work has therefore adapted video generators into action-conditioned world models (ACWMs)~\citep{qiu2026gesim2.0,jiang2025enerverseac,gao2026dreamdojo,guo2025ctrlworld,zhu2024irasim,tseng2026sc3eval}. Given an initial observation and a low-level robot action sequence, an ACWM predicts how the scene will evolve as the robot executes those actions. Such models can potentially support data generation, scalable policy evaluation, and closed-loop policy learning. But for these applications, producing a realistic future is not enough. The predicted future should reproduce what actually happens when the same actions are executed in the real world. As illustrated in Fig.~\ref{fig:motivation}, a rollout can look visually convincing while still getting the outcome wrong. In the ``pick fruit'' example, the real gripper misses the fruit, but the model hallucinates a successful grasp.

We refer to the agreement between a predicted rollout and its paired real-world execution as \emph{execution coherence}. A reliable ACWM should maintain this coherence under three basic stress tests. First, \emph{causal obedience} asks whether the prediction follows the given action sequence, including incorrect actions and failed attempts. Second, \emph{distribution transport} asks whether this agreement holds across different embodiments and action distributions produced by human operators or autonomous policies. Third, \emph{temporal endurance} asks whether this coherence can be maintained over long rollouts, rather than only over short prediction horizons.

Existing benchmarks do not jointly test these properties against physical execution. General world-model benchmarks~\citep{duan2025worldscore,li2026worldmodelbench} assess visual quality, semantic alignment, controllability, and physical plausibility, but do not target embodied action-conditioned prediction. Embodied world-model benchmarks~\citep{yue2025ewmbench,deng2026rethinkingrbench,chen2026abotezsbench,zhou2026pai,fan2026wow,jiang2026robowm} evaluate manipulation videos conditioned on text or task-level instructions, but do not directly test whether a predicted future matches the physical consequences of a specific low-level action trajectory. More recent action- and executability-oriented benchmarks~\citep{shang2026worldarena,shang2026worldarena2} evaluate embodied world models as data engines, policy evaluators, planners, or interactive learning environments. These evaluations, however, rely mostly on simulation. As a result, they do not directly measure whether an ACWM reproduces what happens when the same actions are executed on a real robot, especially over long trajectories that include both successful and failed or counterfactual behaviors.

To close this gap, we introduce \modelfull, a benchmark for evaluating ACWMs against paired real-world executions. Its central question is simple: given the same initial observation and low-level action trajectory, does the predicted future match what actually happens in the real world?

\begin{figure}[t]
    \begin{center}
\centerline{\includegraphics[width=1\linewidth]{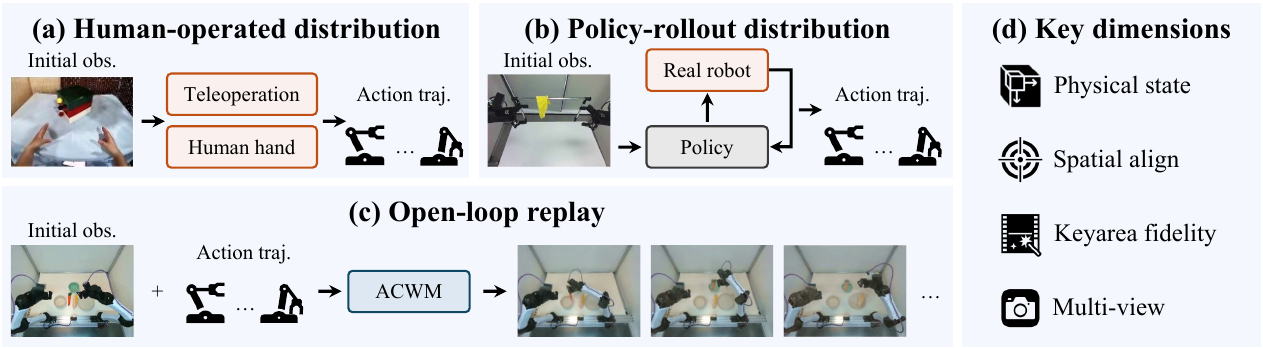}}
    \caption{\textbf{Overview of \modelfull.} \textbf{(a)} The human-operated trajectories are collected through robot teleoperation or direct human-hand execution. \textbf{(b)} The policy-rollout trajectories are collected by running a policy in closed loop on a real robot. \textbf{(c)} For both distributions, the recorded action trajectory is replayed open loop by the ACWM, yielding a generated rollout. \textbf{(d)} The predicted rollout is compared with the paired real execution along four dimensions: \physical, \spatial, \keyarea, and \multiview.}
    \label{fig:framework}
    \end{center}
\end{figure}

As illustrated in Fig.~\ref{fig:framework}, \modelfull is designed around the three stress tests introduced above: causal obedience is tested by replaying the same action trajectory in the ACWM and comparing it with the paired real execution; distribution transport is tested across embodiments and action sources; and temporal endurance is tested over long, multi-segment rollouts. \textbf{Causal obedience.}
We evaluate whether an ACWM follows the given action sequence even when the execution is imperfect or unsuccessful. Each episode is divided into temporally ordered \emph{interaction segments}, where each segment corresponds to one manipulation attempt and ends at an \emph{outcome keyframe} that records its physical result, such as a successful lift, an empty grasp, or a dropped object (Fig.~\ref{fig:motivation}). This lets us compare the predicted and real outcomes of individual attempts, including failures. \textbf{Distribution transport.}
We evaluate whether execution coherence transfers across embodiments and action distributions. Our benchmark includes AgiBot G2, YAM, and human-hand embodiments, covering different robot morphologies and interaction styles. We also include a policy-rollout split on AgiBot G2, where actions are generated by a closed-loop policy. These rollouts introduce a different action distribution, including hesitation, jitter, and corrections. \textbf{Temporal endurance.}
We evaluate long-duration rollouts, testing whether execution coherence is maintained as prediction errors accumulate over time.

For each episode, the ACWM receives the recorded initial observation and low-level action trajectory and replays the actions open loop to generate a predicted rollout. We then compare this rollout with its paired real-world reference along four dimensions. \textit{Physical state correctness} measures whether the predicted outcome matches the real physical state. \textit{Interaction spatial alignment} measures whether the robot and manipulated objects remain aligned with their real-world trajectories. \textit{Key interaction area fidelity} evaluates visual fidelity in the interaction regions, including manipulated objects and robot arms or human hands. For models that support multi-view outputs, \textit{\multiview} tests whether the predicted robot execution is aligned across synchronized camera views.

To support this evaluation, we curate real-world training and evaluation suites across AgiBot G2, YAM, and human-hand embodiments. The training suite includes a newly collected AgiBot G2 dataset, supplemented with the public ABC-130K dataset~\citep{allshire2026scalableabc130k}. The evaluation suite contains human-operated rollouts for all three embodiments and a policy-rollout split on AgiBot G2. As shown in Fig.~\ref{fig:motivation}, the benchmark covers four tasks with different interaction dynamics: fruit picking for rigid-body manipulation, towel hanging for deformable objects, water pouring for fluid dynamics, and drawer pulling for articulated objects. Each episode contains multiple consecutive interaction segments, enabling evaluation of repeated manipulation attempts over long rollouts. The datasets also include both successful and failed interaction segments. This lets \modelfull test whether ACWMs follow the given actions across different outcomes, embodiments, and rollout lengths.

We also build an annotation pipeline for the evaluation suite. As shown in Fig.~\ref{fig:annotation}, human annotators mark interaction-segment boundaries and their outcome keyframes. A VLM then generates physical-state criteria for each keyframe, such as ``the green cup is grasped by the gripper''. These criteria are used to evaluate \physical. We use SAM3~\citep{carion2026sam3} to obtain frame-wise masks of manipulated objects, robot arms, and human hands. These masks focus the evaluation on action-relevant regions instead of the background, and are used for spatial matching and local visual comparison in \spatial and \keyarea. We apply the same annotation procedure to synchronized camera views for \multiview.

We evaluate seven representative models across metrics, tasks, embodiments, and rollout lengths. The experiments reveal several failure modes. Visually plausible rollouts can misrepresent object dynamics and interaction outcomes, even when robot-arm motion is accurately reproduced. Strong short-horizon performance also does not consistently translate into reliable long-horizon prediction. We also observe performance patterns associated with different modeling choices, which we discuss as hypotheses rather than causal conclusions. In summary, our contributions are threefold:
\begin{itemize}[
    leftmargin=*,
    nosep
]
    \item We introduce \modelfull, a benchmark for evaluating ACWM execution coherence against paired real-world rollouts under the same initial observations and low-level action trajectories. 
    \item We curate real-world training and evaluation suites spanning diverse interaction dynamics, multiple embodiments, human-operated and policy-rollout trajectories, and long episodes with both successful and failed executions. 
    \item We provide an evaluation toolkit and leaderboard covering four complementary metrics, and use them to identify recurring failure modes and performance patterns across current ACWMs.
\end{itemize}

\section{Related Work}
\label{sec:related_work}

\subsection{Embodied World Models}
Recent video generators have motivated embodied world models (EWMs) that serve as learned visual simulators for robotic interaction~\citep{nvidia2025cosmos,wan2025wan,deepmind2025veo3}. Early EWMs predict future observations from an initial image and a high-level instruction, whereas action-conditioned world models (ACWMs) introduce low-level actions to control the generated rollouts~\citep{zhu2024irasim,guo2025ctrlworld,nvidia2025cosmos,gao2026dreamdojo,qiu2026gesim2.0}. 

ACWMs differ primarily in their action injection mechanisms. IRASim~\citep{zhu2024irasim} aligns frame-level action embeddings with generated observations. Ctrl-World~\citep{guo2025ctrlworld} supports joint multi-view rollouts with memory retrieval, while ORV~\citep{yang2025orv} and Kinema4D~\citep{xu2026kinema4d} introduce occupancy- and kinematics-based geometric guidance. DreamDojo~\citep{gao2026dreamdojo} transfers latent actions learned from human videos to continuous robot actions, and Cosmos 3~\citep{nvidia2025cosmos} unifies video and action across heterogeneous spaces. GE-Sim 2.0~\citep{qiu2026gesim2.0} further advances closed-loop simulation by jointly predicting proprioceptive states. 

\modelfull provides an evaluation suite that allows each ACWM to use its native conditioning interface while being evaluated against paired real-world execution references.

\begin{table*}[t]
\resizebox{\textwidth}{!}{%
\centering
\small
\setlength{\tabcolsep}{4.5pt} 
\begin{tabular}{lccccccccc}
\toprule
Benchmark & \# Samples & Action. & Real. & Fail. & Long. & Multi-Embod. & Multi-view. & Policy Dist. \\
\midrule
EWMBench~\citep{yue2025ewmbench}       & 100  & \xmark & \cmark & \xmark & \xmark & \xmark(AgiBot G1) & \xmark & \xmark \\
PAI-Bench~\citep{zhou2026pai}  & 2808  & \xmark & \cmark & \xmark & \xmark & \cmark(AgiBot G1, WidowX, Human hand) & \xmark & \xmark \\
WoW-World-Eval~\citep{fan2026wow}                & 609  & \xmark & \cmark & \xmark & \xmark & \cmark(Franka, UR5e, AgileX, humanoid) & \xmark & \xmark \\
RBench~\citep{deng2026rethinkingrbench}           & 650  & \xmark & \cmark & \xmark & \xmark & \cmark(single-, dual-arm,humanoid, quadruped) & \xmark & \xmark \\
WorldArena~\citep{shang2026worldarena} & 500 & \cmark & \xmark & \xmark & \xmark & \xmark(RoboTwin) & \xmark & \cmark \\
EZSBench~\citep{chen2026abotezsbench} & 196  & \xmark & \cmark & \xmark & \xmark & \cmark(real/synthetic robot morphologies) & \xmark & \xmark \\
RoboWM-Bench~\citep{jiang2026robowm} & 160  & \xmark & \cmark & \cmark & \xmark & \cmark(Franka, Human hand) & \xmark & \xmark \\
WorldArena 2.0~\citep{shang2026worldarena2}  & 500+ & \cmark & \xmark & \xmark & \xmark & \cmark(RoboTwin, LIBERO/Panda, AgileX) & \xmark & \cmark \\
\textbf{\modelfull (Ours)} & 335 & \cmark & \cmark & \cmark & \cmark & \cmark(AgiBot G2, YAM, Human hand) & \cmark & \cmark \\
\bottomrule
\end{tabular}
}
\caption{\textbf{Comparison of Embodied World-Model Benchmarks.} \textit{\# Samples}: number of samples for evaluation. \textit{Action.}: evaluation of models conditioned on action sequences. \textit{Real.}: real-world data. \textit{Fail.}: data with failed executions. \textit{Long.}: long-duration execution traces with multiple interaction segments. \textit{Multi-Embod.}: evaluation across multiple embodiments. \textit{Multi-view.}: evaluation across multiple camera views. \textit{Policy Dist.}: action trajectories generated by autonomous policy rollouts.
}
\label{tab:benchmark_comparison}
\end{table*}

\subsection{Embodied World Model Benchmarks}
A growing number of benchmarks evaluate generated videos in embodied settings~\citep{yue2025ewmbench,zhou2026pai,fan2026wow,deng2026rethinkingrbench,shang2026worldarena,shang2026worldarena2,chen2026abotezsbench,jiang2026robowm}. EWMBench~\citep{yue2025ewmbench}, RBench~\citep{deng2026rethinkingrbench}, and EZS-Bench ~\citep{chen2026abotezsbench} assess properties such as visual quality, motion, physical consistency, task completion, and cross-embodiment generalization. PAI Bench~\citep{zhou2026pai} and WoW-World-Eval~\citep{fan2026wow} further consider physical understanding, planning, plausibility, and executability. However, these benchmarks primarily evaluate videos conditioned on initial image and a task-level instruction.

More recent benchmarks consider embodied execution. WorldArena~\citep{shang2026worldarena} evaluates world models as data engines, policy evaluators, and action planners, but conducts its functional evaluation in the simulator. WorldArena 2.0~\citep{shang2026worldarena2} extends this setting to visuotactile prediction, online policy learning, and cross-platform transfer, while remaining predominantly simulation-based. RoboWM-Bench~\citep{jiang2026robowm} evaluates physical executability by extracting actions from generated videos and replaying them in real-to-sim environments. Despite these advances, none directly assess whether action-conditioned predictions are coherent with corresponding real-world executions. We show a detailed comparison with existing benchmarks in Table~\ref{tab:benchmark_comparison}.

\section{The \modelfull Benchmark}
\label{sec:method}

\paragraph{Problem formulation.} We represent episode $i$ as $\mathcal{E}_i=\left(\mathbf{O}_i,\mathbf{A}_i\right)$. The observation sequence $\mathbf{O}_i=\left(\{O_{i,t}^{v}\}_{v\in\mathcal{V}}\right)_{t=0}^{T_i}$
contains synchronized observations from the camera views
$\mathcal{V}=\{\texttt{top\_head},\allowbreak\texttt{hand\_left},\allowbreak\texttt{hand\_right}\}$. We use \texttt{top\_head} as the primary view, while \texttt{hand\_left} and \texttt{hand\_right} are the left and right hand-mounted camera views. The corresponding low-level action sequence is denoted by
$\mathbf{A}_i=(A_{i,t})_{t=1}^{T_i}$. Each episode is divided into $N_i$ temporally ordered interaction segments
$\{\mathcal{S}_{i,n}\}_{n=1}^{N_i}$.
Each segment corresponds to one manipulation attempt and spans
$\mathcal{S}_{i,n}
=
[t_{i,n}^{\mathrm{start}},t_{i,n}^{\mathrm{end}}]$.
We define its outcome keyframe as the final observation of the segment: $K_{i,n}^{v}
\coloneqq
O_{i,t_{i,n}^{\mathrm{end}}}^{v}
$. The outcome keyframe captures the observable physical result of the interaction. For each evaluation episode $i$, \modelfull provides the initial observation $\left\{O_{i,0}^{v}\right\}_{v\in\mathcal{V}}$ and the recorded real-world action sequence $\mathbf{A}_i$. An ACWM $\mathcal{W}_{\theta}$ generates the predicted rollout:
\begin{equation}
\hat{\mathbf{O}}_i=\mathcal{W}_{\theta}\left(\left\{O_{i,0}^{v}\right\}_{v\in\mathcal{V}},\mathbf{A}_i\right),
\end{equation}
which is compared with the paired real-world rollout $\mathbf{O}_i$. Because both rollouts start from the same observation and follow the same action sequence, this comparison directly tests whether the ACWM reproduces the physical consequences of those actions correctly.

\paragraph{Design overview.} \modelfull is designed to support ACWMs with various low-level action conditioning and injection mechanisms. We design benchmark tasks based on two principles. First, they should cover different types of interaction dynamics, including rigid-body, deformable-object, fluid, and articulated-object interactions. Second, both successful and failed executions should lead to physical outcomes that are directly observable and visually unambiguous. This makes it possible to evaluate task outcomes reliably with a VLM. These principles lead to four representative tasks shown in Fig.~\ref{fig:motivation}: \textit{pick fruit}, \textit{hang towel}, \textit{pour water}, and \textit{pull drawer}.

We next describe the training and evaluation suites, the annotation pipeline, and the four evaluation metrics used to quantify execution coherence.

\begin{figure}[t]
    \begin{center}
\centerline{\includegraphics[width=1\linewidth]{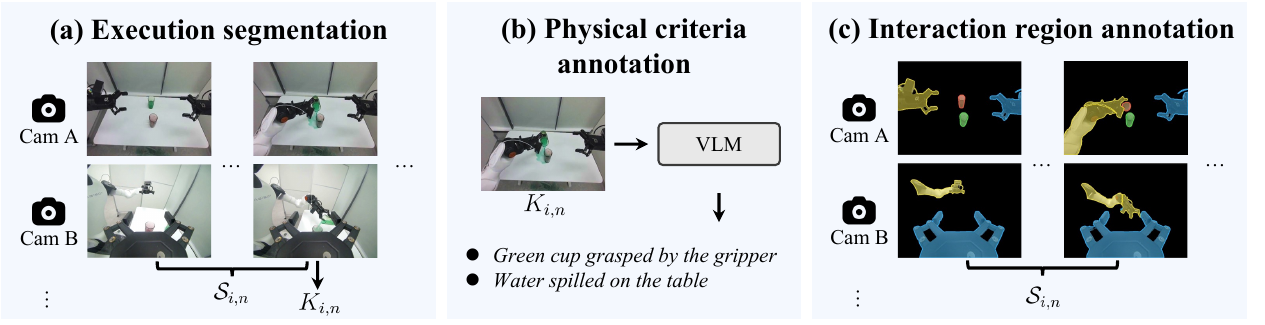}}
    \caption{\textbf{\modelfull annotation pipeline.} \textbf{(a)} Execution segmentation: each synchronized multi-view episode is divided into interaction segments $\mathcal{S}_{i,n}$, and the final frame of each segment is the outcome keyframe $K_{i,n}$ capturing the observable physical outcome. \textbf{(b)} Physical criteria annotation: a VLM generates physical-state criteria for each outcome keyframe. \textbf{(c)} Interaction region annotation: manipulated objects, robot arms, and human hands are annotated frame-wise across camera views.}
    \label{fig:annotation}
    \end{center}
\end{figure}

\subsection{Data Curation}
\label{sec:data_curation}

\paragraph{Training suite.} The training suite is used to adapt models to the target embodiments before evaluation on the held-out test set. For AgiBot G2, we collect a new real-world dataset with 1,779 episodes across the aforementioned four benchmark tasks. Each episode lasts about one minute and contains multiple interaction segments. Around $20\%$ of the segments in each episode are failed or suboptimal executions. This composition enables ACWMs to learn the actual physical consequences of erroneous actions instead of associating every action sequence with successful task completion. For YAM, we use ABC-130K~\citep{allshire2026scalableabc130k}, a large public dataset of bimanual manipulation collected on YAM stations. We do not provide adaptation data for the human-hand embodiment; models with egocentric world-modeling capabilities are evaluated zero-shot instead. More details are provided in App.~\ref{sec:appendix_training_suite} and Table~\ref{tab:app_training_set_statistics}.

\paragraph{Human-operated distribution.} The human-operated evaluation split contains trajectories collected through robot teleoperation or direct human-hand execution (Fig.~\ref{fig:framework}(a)). We collect held-out real-world episodes for AgiBot G2, YAM, and human-hand embodiments, with 100, 75, and 100 episodes, respectively. Each episode lasts about two minutes and contains multiple consecutive interaction segments, including both successful and failed manipulation attempts. Compared with the training suite, the evaluation suite is longer and contains a higher proportion of failures, with approximately $40\%$ of the segments corresponding to failed executions for each episode. This increased failure frequency tests whether an ACWM can reproduce the consequences of non-ideal actions instead of defaulting to successful task completion.

We design each interaction segment to end in a physical outcome that is clearly visible and easy for a VLM to judge. For example, the VLM evaluator can directly determine whether a fruit is held by the gripper or placed in a target container, whether liquid is transferred or spilled, whether a drawer is open or closed, and whether a towel is hanging over the rack or remains on the table. This enables reliable VLM-assisted assessment during the evaluation. Details are provided in App.~\ref{sec:appendix_human_distribution} and Table~\ref{tab:app_evaluation_set_statistics}.

\paragraph{Policy-rollout distribution.} We additionally construct a policy-rollout evaluation split on the physical AgiBot G2 robot (Fig.~\ref{fig:framework}(b)). A $\pi_{0.5}$ policy~\citep{intelligence2025pi05} trained on our curated training data interacts with the real robot in closed loop. During evaluation, each recorded action trajectory is replayed open loop from its paired initial observation (Fig.~\ref{fig:framework}(c)). This split contains 60 episodes. More details are provided in App.~\ref{sec:appendix_policy_distribution} and App.~\ref{app:training_implementation_details}.

\paragraph{Annotation pipeline.} We build a systematic annotation pipeline for evaluation suites. As illustrated in Fig.~\ref{fig:annotation}, each evaluation episode is temporally divided into interaction segments. Each segment terminates at an outcome keyframe that captures an observable successful or failed physical outcome. For physical criteria annotation, a VLM-assisted pipeline generates physical-state criteria for each outcome keyframe. For interaction region annotation, we use the segmentation method to obtain frame-wise masks of manipulated objects and robot arms or human hands. (See App.~\ref{app:annotation_pipeline} for details).

\begin{figure}[!t]
    \begin{center}
\centerline{\includegraphics[width=1\linewidth]{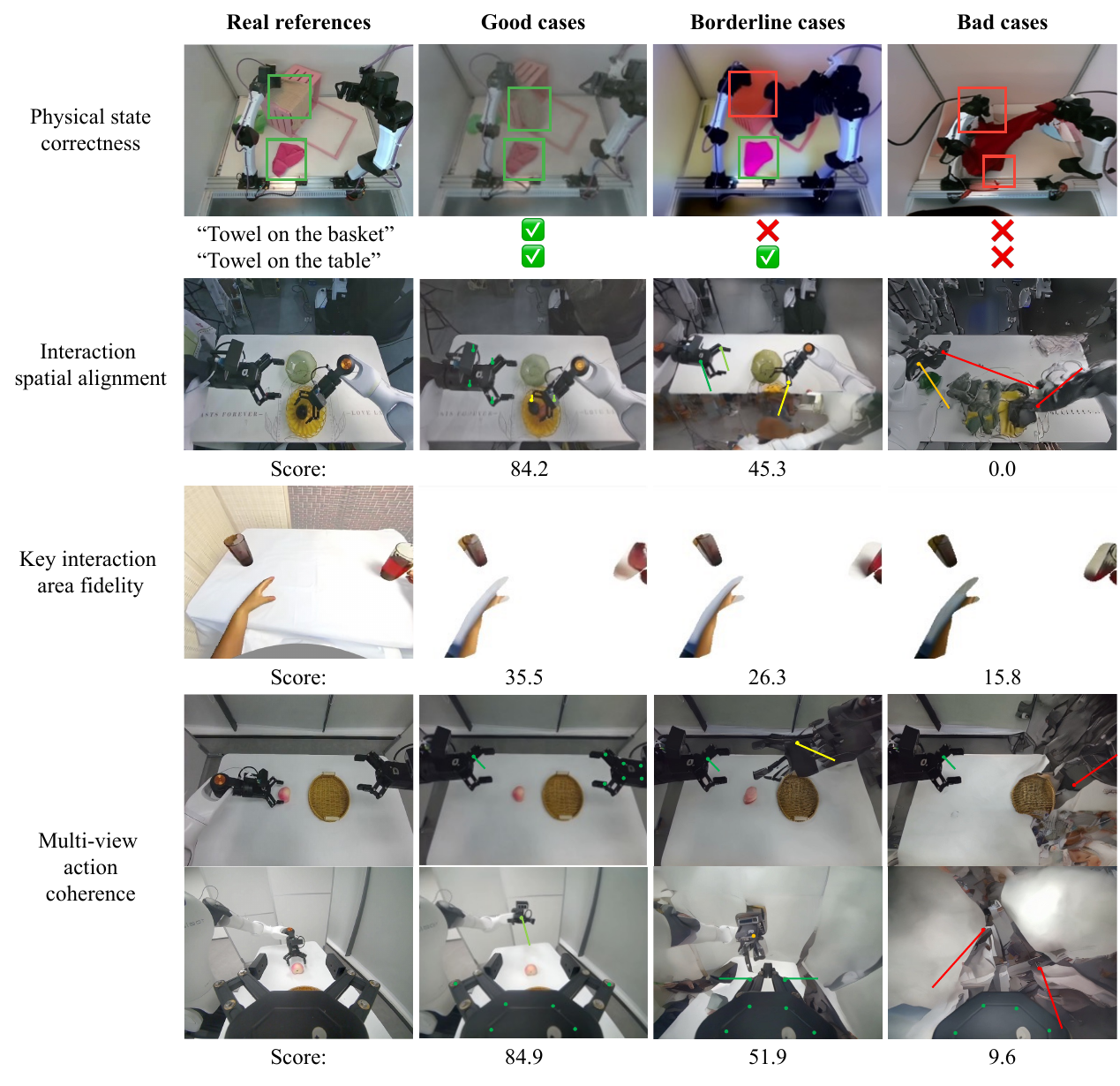}}
\caption{\textbf{Qualitative examples.} Each row compares a real-world reference with high-, intermediate-, and low-scoring predictions. \textit{Physical state correctness} checks whether the prediction satisfies the criteria. The bad case predicts states of both towels wrong. \textit{Interaction spatial alignment.} Points indicate matched visual features and lines show their displacement to reference. Point color denotes matching confidence from green (high) to red (low), while shorter lines indicate better spatial alignment. \textit{Key interaction area fidelity.} Within the interaction regions, the predicted entities of the bad example exhibit degraded appearance and structural distortion. \textit{Multi-view action coherence.} In the bad example, the predicted robot arm is poorly aligned in the synchronized camera views.
}
    \label{fig:qualitative}
    \end{center}
\end{figure}

\subsection{The \modelfull Metrics}
\label{sec:metrics}

Given a real observation sequence $\mathbf{O}_i=(\mathbf{O}_{i,t})_{t=0}^{T_i}$ and a prediction $\hat{\mathbf{O}}_i=(\hat{\mathbf{O}}_{i,t})_{t=0}^{T_i}$ conditioned on the same initial observation and action trajectory, \modelfull evaluates execution coherence along four dimensions. Physical state correctness, \spatial, and \keyarea are evaluated on the primary \texttt{top\_head} view. Multi-view action coherence evaluates all three synchronized views in $\mathcal{V}$. Full metric definitions and details are provided in App.~\ref{app:metric_details}.

\paragraph{Physical state correctness.} A rollout can look realistic while still predicting the wrong physical outcome. For example, the model may hallucinate a successful grasp of an object even when the given action trajectory causes the real gripper to close without picking up the object (Fig.~\ref{fig:motivation}). Therefore, for each interaction segment, we evaluate the predicted outcome keyframe against physical-state criteria annotated from the paired real-world keyframe (Fig.~\ref{fig:annotation}(b)). A VLM assigns a binary judgement to each criterion. The resulting binary criterion scores are averaged across criteria for each segment, and the episode score is obtained by averaging over its execution segments. This metric evaluates whether the predicted execution reaches the same physical outcome as the real execution.

\paragraph{Interaction spatial alignment.} Matching the final physical outcome does not necessarily mean that an ACWM follows the correct motion throughout the execution. We therefore introduce \spatial, which measures whether robot arms or human hands and manipulated entities remain spatially aligned with their real-world trajectories. At each sampled frame, DINOv3 features~\citep{simeoni2025dinov3} within the annotated arm or hand and object masks (Fig.~\ref{fig:annotation}(c)) are matched to features in the predicted observations. The score combines the spatial displacement of these matches with their correspondence confidence, assigning a high score when the action-relevant entities are reliably identified near their real-world positions over time.

\paragraph{Key interaction area fidelity.} Whole-frame perceptual metrics can be dominated by background pixels and overlook errors in regions where interactions actually happen. We therefore use masks from the real execution to identify the expected regions of manipulated objects and robot arms or human hands (as shown in Fig.~\ref{fig:annotation}(c)), and evaluate fidelity only within these regions. At each sampled frame, the corresponding masked regions are cropped from the predicted and real observations. We then evaluate their pixel-level, perceptual, and semantic appearance fidelity using PSNR~\citep{huynh2008scopepsnr}, LPIPS~\citep{zhang2018unreasonablelpips}, and V-JEPA similarity~\citep{assran2025vjepa}, respectively. After normalization, the three components are combined using a geometric mean, so a severe error in any component lowers the final score. A high score requires the relevant entity to appear in the expected region and to match the paired real execution in appearance. This also penalizes spatial drift outside the expected region. The metric captures local differences such as object deformation, gripper artifacts, and incorrect contact geometry.

\paragraph{Multi-view action coherence.} For multi-view models, \multiview measures robot execution alignment with the paired real execution across all synchronized camera views. It is reference-grounded rather than a direct measure of intrinsic geometric consistency between generated views, as structure-from-motion estimates for robot arms are unreliable. We therefore compute robot-arm alignment independently in each view using a similar correspondence strategy as \spatial, and then aggregate the per-view scores. A frame receives a high score only when the predicted robot motion is well aligned in every view. This penalizes cases where the robot appears correctly aligned in one view but is missing or spatially misplaced in another.

\paragraph{Score normalization and aggregation.} All four dimensions are reported on a common $[0,100]$ scale, using the metric-specific scoring and normalization procedures described in App.~\ref{app:metric_details}. Let $S_{\mathrm{phys}}$, $S_{\mathrm{isa}}$, $S_{\mathrm{key}}$, $S_{\mathrm{mv}}$ denote the scores for \physical, \spatial, \keyarea, and \multiview, respectively. The single-view track aggregates the three dimensions evaluated on the primary \texttt{top\_head} view:
\begin{equation}
\mathrm{WorldOracle\text{-}SV}
=
\frac{
S_{\mathrm{phys}}
+
S_{\mathrm{isa}}
+
S_{\mathrm{key}}
}{3}.
\label{eq:acwmscore_sv}
\end{equation}
For multi-view predictions, the multi-view track additionally includes \multiview:
\begin{equation}
\mathrm{WorldOracle\text{-}MV}
=
\frac{
S_{\mathrm{phys}}
+
S_{\mathrm{isa}}
+
S_{\mathrm{key}}
+
S_{\mathrm{mv}}
}{4}.
\label{eq:acwmscore_mv}
\end{equation}
All dimensions receive equal weight within their respective tracks. We further report scores under two temporal regimes. The short-duration regime includes sampled frames and outcome keyframes within the first 30 seconds. We choose this boundary based on the empirical duration distribution of existing manipulation datasets; details are provided in Sec.~\ref{app:short_horizon}. The long-episode regime evaluates all available frames and outcome keyframes. Models with synchronized multi-view outputs are evaluated on both tracks, while single-view models are evaluated only on \metricsv.

\section{Results}
\label{sec:exp}

\begin{table*}[t]
\centering
\small
\setlength{\tabcolsep}{4pt}
\setlength{\aboverulesep}{0pt}
\setlength{\belowrulesep}{0pt}
\renewcommand{\arraystretch}{1.15}

\caption{\textbf{Results on AgiBot G2.}
The 30s and Long rows evaluate the first 30 seconds and the complete episode.
All metrics are reported on a $[0,100]$ scale; higher is better.}
\label{tab:agibot_results}

\resizebox{\textwidth}{!}{%
\begin{tabular}{
llcccc
>{\columncolor[RGB]{240,246,254}}c
>{\columncolor[RGB]{240,246,254}}c
}
\toprule
\multirow{2}{*}{Model}
& \multirow{2}{*}{Regime}
& \multirow{2}{*}{\shortstack{Physical\\Correctness}}
& \multirow{2}{*}{\shortstack{Spatial\\Alignment}}
& \multirow{2}{*}{\shortstack{Key Area\\Fidelity}}
& \multirow{2}{*}{\shortstack{Multi-View\\Coherence}}
& \multicolumn{2}{>{\columncolor[RGB]{240,246,254}}c}{WorldOracle} \\
\cmidrule(lr){7-8}
& & & & & & -SV & -MV \\
\midrule

\multirow{2}{*}{IRASim~\citep{zhu2024irasim}}
& 30s
& $\underline{55.64}\pm3.16$
& $53.96\pm2.10$
& $38.00\pm1.26$
& --
& $49.20\pm1.31$
& -- \\
& Long
& $\underline{42.06}\pm2.03$
& $\underline{37.60}\pm1.72$
& $\underline{25.47}\pm1.00$
& --
& $\underline{35.04}\pm0.95$
& -- \\

\multirow{2}{*}{Ctrl-World~\citep{guo2025ctrlworld}}
& 30s
& $51.15\pm3.47$
& $57.08\pm2.40$
& $40.53\pm1.51$
& $\underline{65.57}\pm2.05$
& $49.59\pm1.47$
& $\underline{53.58}\pm1.19$ \\
& Long
& $38.31\pm2.96$
& $29.44\pm2.34$
& $23.45\pm1.23$
& $\underline{39.31}\pm1.88$
& $30.40\pm1.32$
& $\underline{32.62}\pm1.10$ \\

\multirow{2}{*}{DreamDojo~\citep{gao2026dreamdojo}}
& 30s
& $53.32\pm3.63$
& $53.96\pm1.71$
& $\underline{44.39}\pm1.31$
& --
& $50.56\pm1.41$
& -- \\
& Long
& $37.97\pm2.30$
& $18.10\pm1.01$
& $25.18\pm0.92$
& --
& $27.08\pm0.87$
& -- \\

\multirow{2}{*}{Cosmos 3~\citep{nvidia2025cosmos}}
& 30s
& $53.12\pm3.07$
& $\underline{67.41}\pm1.52$
& $41.44\pm1.08$
& --
& $\underline{53.99}\pm1.17$
& -- \\
& Long
& $33.43\pm1.71$
& $25.26\pm1.05$
& $21.10\pm0.84$
& --
& $26.60\pm0.72$
& -- \\

\multirow{2}{*}{GE-Sim 2.0~\citep{qiu2026gesim2.0}}
& 30s
& $\mathbf{72.54}\pm3.52$
& $\mathbf{81.40}\pm1.34$
& $\mathbf{67.62}\pm1.04$
& $\mathbf{97.53}\pm0.25$
& $\mathbf{73.86}\pm1.27$
& $\mathbf{79.77}\pm0.95$ \\
& Long
& $\mathbf{60.26}\pm2.79$
& $\mathbf{73.49}\pm1.77$
& $\mathbf{56.24}\pm0.90$
& $\mathbf{96.96}\pm0.39$
& $\mathbf{63.33}\pm1.11$
& $\mathbf{71.74}\pm0.85$ \\

\bottomrule
\end{tabular}%
}

\end{table*}

\begin{table*}[t]
\centering
\small
\setlength{\tabcolsep}{8pt}
\setlength{\aboverulesep}{0pt}
\setlength{\belowrulesep}{0pt}
\renewcommand{\arraystretch}{1.15}

\caption{
\textbf{Results on the AgiBot G2 policy-rollout distribution.}
Each rollout is evaluated over its complete duration.
All scores are reported on a $[0,100]$ scale, and higher is better.
}
\label{tab:agibot_policy_results}

\resizebox{\textwidth}{!}{%
\begin{tabular}{
lcccc
>{\columncolor[RGB]{240,246,254}}c
>{\columncolor[RGB]{240,246,254}}c
}
\toprule
\multirow{2}{*}{Model}
& \multirow{2}{*}{\shortstack{Physical\\Correctness}}
& \multirow{2}{*}{\shortstack{Spatial\\Alignment}}
& \multirow{2}{*}{\shortstack{Key Area\\Fidelity}}
& \multirow{2}{*}{\shortstack{Multi-View\\Coherence}}
& \multicolumn{2}{>{\columncolor[RGB]{240,246,254}}c}{WorldOracle} \\
\cmidrule(lr){6-7}
& & & & & -SV & -MV \\
\midrule

IRASim~\citep{zhu2024irasim}
& $\mathbf{62.18}\pm7.04$
& $51.62\pm3.74$
& $32.30\pm2.19$
& --
& $\underline{48.70}\pm2.67$
& -- \\

Ctrl-World~\citep{guo2025ctrlworld}
& $27.22\pm6.52$
& $42.95\pm3.81$
& $33.16\pm2.52$
& $\underline{62.91}\pm3.69$
& $34.44\pm2.69$
& $\underline{41.56}\pm2.20$ \\

DreamDojo~\citep{gao2026dreamdojo}
& $44.44\pm7.86$
& $52.88\pm3.53$
& $\underline{36.57}\pm2.52$
& --
& $44.63\pm2.97$
& -- \\

Cosmos 3~\citep{nvidia2025cosmos}
& $34.29\pm6.78$
& $\underline{59.81}\pm3.02$
& $34.31\pm2.81$
& --
& $42.80\pm2.60$
& -- \\

GE-Sim 2.0~\citep{qiu2026gesim2.0}
& $\underline{57.03}\pm6.89$
& $\mathbf{78.70}\pm1.84$
& $\mathbf{58.33}\pm1.39$
& $\mathbf{96.64}\pm0.35$
& $\mathbf{64.69}\pm2.37$
& $\mathbf{72.67}\pm1.79$ \\

\bottomrule
\end{tabular}%
}

\end{table*}

\begin{table*}[t]
\centering
\small
\setlength{\tabcolsep}{4pt}
\setlength{\aboverulesep}{0pt}
\setlength{\belowrulesep}{0pt}
\renewcommand{\arraystretch}{1.15}

\caption{\textbf{Results on YAM.}
The 30s and Long rows report short-duration and full-episode evaluation,
respectively. All metrics are reported on a $[0,100]$ scale, and higher is
better.}
\label{tab:yam_results}

\resizebox{\textwidth}{!}{%
\begin{tabular}{
llcccc
>{\columncolor[RGB]{240,246,254}}c
>{\columncolor[RGB]{240,246,254}}c
}
\toprule
\multirow{2}{*}{Model}
& \multirow{2}{*}{Regime}
& \multirow{2}{*}{\shortstack{Physical\\Correctness}}
& \multirow{2}{*}{\shortstack{Spatial\\Alignment}}
& \multirow{2}{*}{\shortstack{Key Area\\Fidelity}}
& \multirow{2}{*}{\shortstack{Multi-View\\Coherence}}
& \multicolumn{2}{>{\columncolor[RGB]{240,246,254}}c}{WorldOracle} \\
\cmidrule(lr){7-8}
& & & & & & -SV & -MV \\
\midrule

\multirow{2}{*}{IRASim~\citep{zhu2024irasim}}
& 30s
& $65.23\pm4.69$
& $51.52\pm1.96$
& $39.67\pm1.13$
& --
& $52.14\pm1.74$
& -- \\
& Long
& $\underline{53.39}\pm2.75$
& $\underline{38.71}\pm1.60$
& $\underline{30.35}\pm0.93$
& --
& $\underline{40.82}\pm1.11$
& -- \\

\multirow{2}{*}{Ctrl-World~\citep{guo2025ctrlworld}}
& 30s
& $56.39\pm4.88$
& $53.43\pm2.70$
& $35.93\pm1.58$
& $\underline{41.87}\pm2.15$
& $48.59\pm1.91$
& $\underline{46.91}\pm1.53$ \\
& Long
& $39.03\pm3.02$
& $20.71\pm2.04$
& $18.97\pm1.35$
& $\underline{15.52}\pm1.34$
& $26.24\pm1.28$
& $\underline{23.56}\pm1.02$ \\

\multirow{2}{*}{DreamDojo~\citep{gao2026dreamdojo}}
& 30s
& $59.14\pm4.31$
& $56.35\pm2.78$
& $\underline{45.84}\pm1.65$
& --
& $53.78\pm1.79$
& -- \\
& Long
& $48.15\pm3.09$
& $19.70\pm2.11$
& $28.11\pm1.13$
& --
& $31.99\pm1.27$
& -- \\

\multirow{2}{*}{Cosmos 3~\citep{nvidia2025cosmos}}
& 30s
& $\mathbf{67.84}\pm4.13$
& $\underline{68.93}\pm2.14$
& $40.71\pm1.18$
& --
& $\underline{59.16}\pm1.60$
& -- \\
& Long
& $\mathbf{62.85}\pm2.52$
& $34.61\pm1.83$
& $16.04\pm0.74$
& --
& $37.83\pm1.08$
& -- \\

\multirow{2}{*}{GE-Sim 2.0~\citep{qiu2026gesim2.0}}
& 30s
& $\underline{67.78}\pm5.35$
& $\mathbf{74.97}\pm1.97$
& $\mathbf{54.64}\pm1.11$
& $\mathbf{64.10}\pm2.17$
& $\mathbf{65.80}\pm1.92$
& $\mathbf{65.37}\pm1.54$ \\
& Long
& $49.37\pm3.17$
& $\mathbf{59.78}\pm2.37$
& $\mathbf{44.41}\pm1.22$
& $\mathbf{48.54}\pm1.56$
& $\mathbf{51.19}\pm1.39$
& $\mathbf{50.53}\pm1.11$ \\

\bottomrule
\end{tabular}%
}

\end{table*}

\begin{table*}[t]
\centering
\small
\setlength{\tabcolsep}{8pt}
\setlength{\aboverulesep}{0pt}
\setlength{\belowrulesep}{0pt}
\renewcommand{\arraystretch}{1.15}

\caption{\textbf{Results on Human Hand.}
The 30s and Long rows report short-duration and full-episode evaluation,
respectively. All metrics are reported on a $[0,100]$ scale, and higher is
better.}
\label{tab:human_hand_results}

\resizebox{0.98\textwidth}{!}{%
\begin{tabular}{
llccc
>{\columncolor[RGB]{240,246,254}}c
}
\toprule
\multirow{2}{*}{Model}
& \multirow{2}{*}{Regime}
& Physical
& Spatial
& Key Area
& \\
&
& Correctness
& Alignment
& Fidelity
& \multirow{-2}{*}{WorldOracle-SV} \\
\midrule

\multirow{2}{*}{
JointControlVideo~\citep{zhang2026controllableJointControlVideo}}
& 30s
& $\mathbf{55.47}\pm4.24$
& $\mathbf{57.24}\pm2.46$
& $20.50\pm1.40$
& $\mathbf{44.40}\pm1.69$ \\
& Long
& $\mathbf{56.33}\pm3.02$
& $\mathbf{23.79}\pm1.66$
& $8.10\pm0.67$
& $\mathbf{29.41}\pm1.18$ \\

\multirow{2}{*}{Cosmos 3~\citep{nvidia2025cosmos}}
& 30s
& $47.46\pm3.88$
& $33.70\pm2.33$
& $\underline{22.21}\pm1.06$
& $34.46\pm1.58$ \\
& Long
& $30.23\pm2.22$
& $9.43\pm0.87$
& $\underline{11.44}\pm0.72$
& $17.03\pm0.82$ \\

\multirow{2}{*}{OSCAR~\citep{wu2026oscar}}
& 30s
& $\underline{53.18}\pm3.83$
& $\underline{47.48}\pm2.34$
& $\mathbf{27.46}\pm1.08$
& $\underline{42.71}\pm1.55$ \\
& Long
& $\underline{48.88}\pm2.35$
& $\underline{16.07}\pm1.04$
& $\mathbf{17.82}\pm0.76$
& $\underline{27.59}\pm0.90$ \\

\bottomrule
\end{tabular}%
}

\end{table*}

\paragraph{Validation.} We validate \modelfull through human studies and controlled analyses. The metric-induced preferences agree closely with human judgements in 88.1--93.2\% of pairwise comparisons, with Spearman correlations ranging from $0.72$ to $0.82$. We show more details in Sec.~\ref{sec:human_metric_alignment}. We also verify the annotated physical criteria on their real outcome keyframes. The VLM judge marks all criteria as satisfied, yielding a criterion--reference agreement of $100\%$. Independent human annotators further check the criteria against the reference images, to avoid relying only on VLM self-consistency. Controlled examples show that \physical and \spatial capture complementary failure modes, while \keyarea decreases monotonically under increasing corruption inside the interaction regions but remains unchanged under background-only corruption. Further validation details are provided in Sec.~\ref{app:metric_validation}.

\paragraph{Models and implementation details.} We evaluate seven ACWMs: IRASim, Ctrl-World, DreamDojo, Cosmos 3 Nano, GE-Sim 2.0, JointControlVideo, and OSCAR. For each model, we start from the official pretrained checkpoint and follow its original training setup, action representation, and conditioning mechanism. The robot-oriented models are adapted to AgiBot or YAM using our AgiBot training suite or the public ABC-130K dataset, while models with egocentric world-modeling capabilities are evaluated zero-shot on the human-hand embodiment. To collect the policy-rollout evaluation split, we initialize a $\pi_{0.5}$ policy from its official DROID checkpoint, fine-tune it on the AgiBot training suite, and deploy it on the physical robot. Complete training and implementation details are provided in App.~\ref{app:training_implementation_details}.

\subsection{Observations and Challenges}

The AgiBot G2, YAM, and human-hand evaluations should not be interpreted as a controlled comparison of embodiment difficulty. The models are adapted differently across settings: AgiBot G2 uses a benchmark-specific training suite, YAM relies on the public ABC-130K dataset and therefore has a larger train--test gap, and the human-hand models are evaluated zero-shot. We therefore compare models primarily within each setting. Results are shown in Tables~\ref{tab:agibot_results}--\ref{tab:human_hand_results}. Each score is reported as the mean $\pm$ 95\% bootstrap confidence interval, estimated from 5,000 episode resamples. More results and analyses are provided in Sec.~\ref{app:extended_results}.

\paragraph{Spatially grounded conditioning is associated with stronger action alignment.} GE-Sim~2.0 achieves the highest \spatial scores on both robot embodiments: $81.40/73.49$ in the 30s/long AgiBot setting and $74.97/59.78$ on YAM. It also reaches $78.70$ on the AgiBot policy-rollout distribution. One possible explanation is its use of pixel-aligned pose maps, which explicitly connect end-effector motion to image locations. However, the evaluated models also differ in pretraining, architecture, memory, and pretrained embodiment. The present comparison therefore does not establish that pixel-aligned conditioning alone causes the improvement.

\paragraph{Execution coherence degrades over long autoregressive rollouts.} All evaluated models decline as the rollout horizon increases, although at different rates. As shown in Table~\ref{tab:agibot_results}, the \metricsv scores of DreamDojo and Cosmos~3 decrease from $50.56$ to $27.08$ and from $53.99$ to $26.60$, respectively. GE-Sim~2.0 degrades less, retaining $63.33$ from $73.86$. The results show that strong short-horizon performance does not necessarily carry over to long rollouts.

\paragraph{Strong action alignment does not guarantee the correct physical outcome.} A model can follow the robot trajectory closely while still predicting the wrong interaction outcome. In the full-episode evaluation in Table~\ref{tab:agibot_results}, GE-Sim~2.0 obtains $73.49$ in \spatial and $96.96$ in \multiview, but lower scores of $60.26$ in \physical and $56.24$ in \keyarea. This suggests that accurate action alignment and faithful physical prediction remain separate challenges. In particular, a rollout may track the robot motion well while differing from the paired real execution in grasp state, contact, object motion, or task outcome.

\begin{figure*}[ht]
    \centering
    \includegraphics[width=\textwidth]{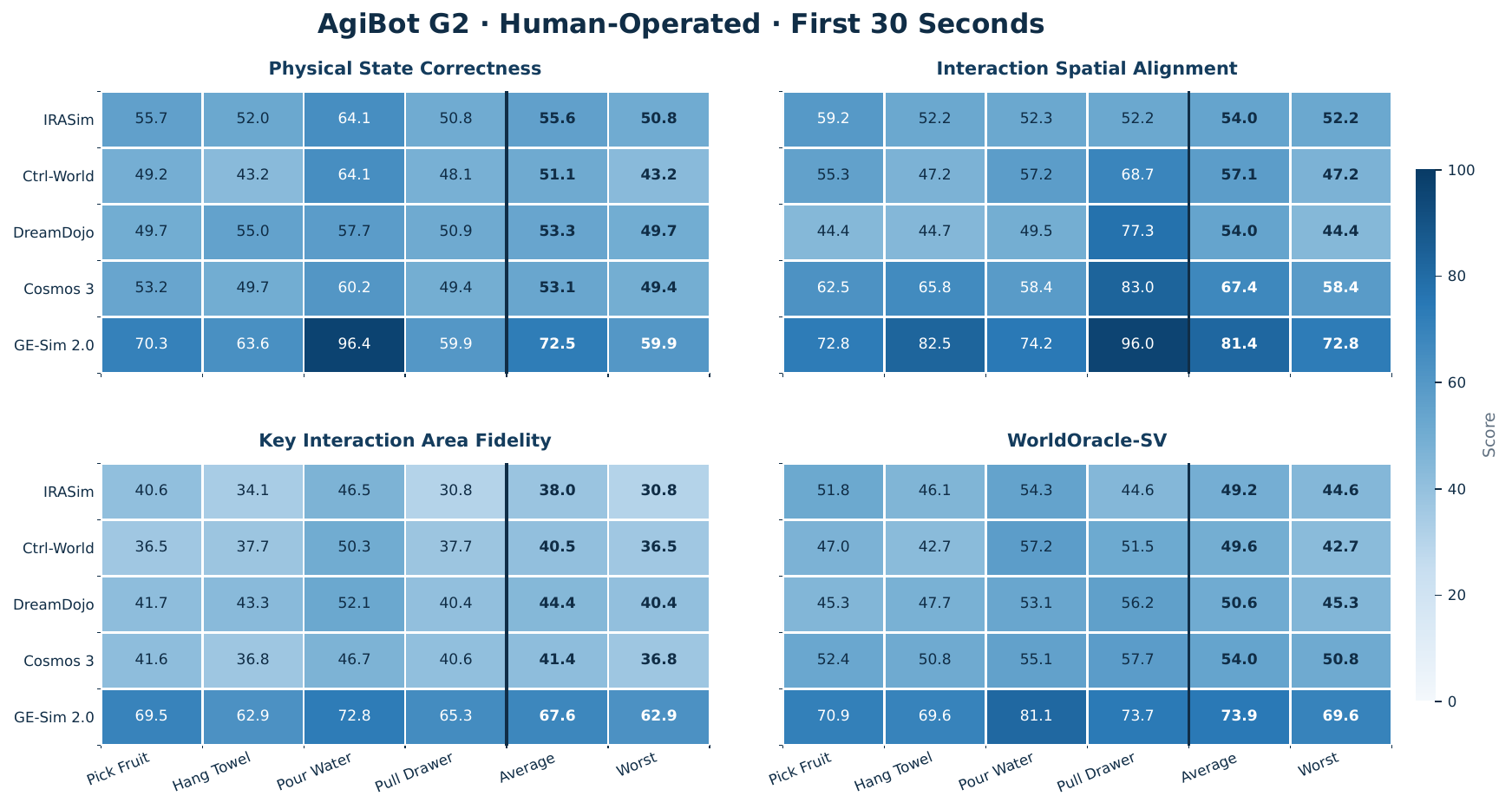}
    \caption{
    \textbf{Task-wise interaction-dynamics analysis on human-operated
    AgiBot G2 split.} Results are evaluated over the first 30 seconds on \physical, \spatial, \keyarea, and WorldOracle-SV. In each heatmap, \emph{Average} denotes the arithmetic mean across tasks, while \emph{Worst} denotes the lowest task score.
    }
    \label{fig:interaction_dynamics_agibot_human}
\end{figure*}

\section{Extended Results and Analyses}
\label{app:extended_results}

This section provides detailed result breakdowns across tasks, action distributions, embodiments, and evaluation horizons. The models differ in architecture, action representation, training data, and training embodiment. Moreover, adaptation is not matched across the three embodiments: AgiBot G2 uses benchmark-specific training data, YAM has a larger train--test gap, and the human-hand models are evaluated zero-shot. \textbf{The following results therefore describe performance patterns and hypotheses rather than controlled causal effects.}

\begin{figure*}[!t]
    \centering
    \includegraphics[width=\textwidth]{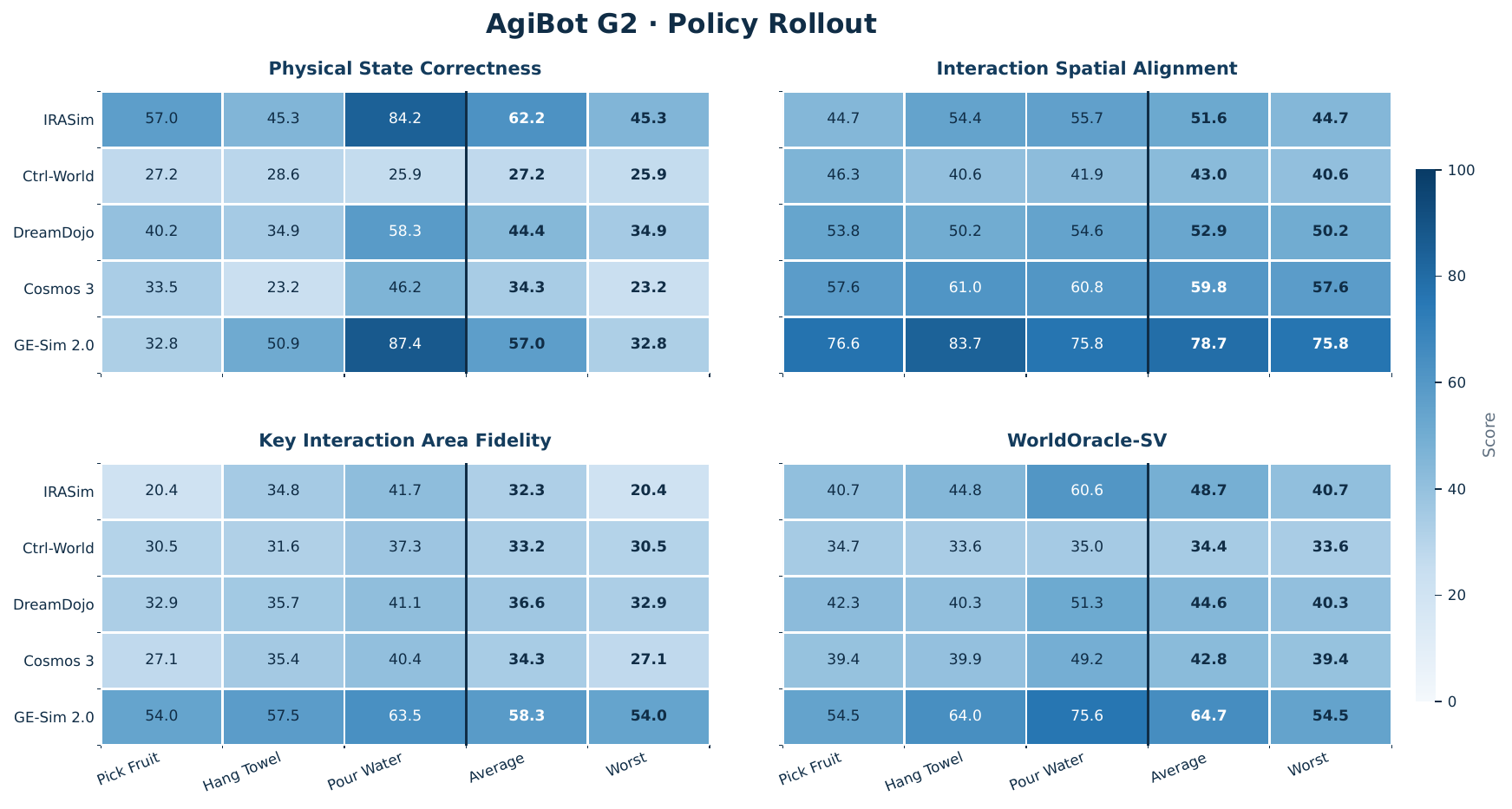}
    \caption{
    \textbf{Task-wise interaction-dynamics analysis on AgiBot G2
    policy-rollout split.}
    We report results on \physical, \spatial, \keyarea, and WorldOracle-SV. In each heatmap, \emph{Average} denotes the arithmetic mean across tasks, while \emph{Worst} denotes the lowest task score.
    }
    \label{fig:interaction_dynamics_agibot_policy}
\end{figure*}
\begin{figure*}[!t]
    \centering
    \includegraphics[width=\textwidth]{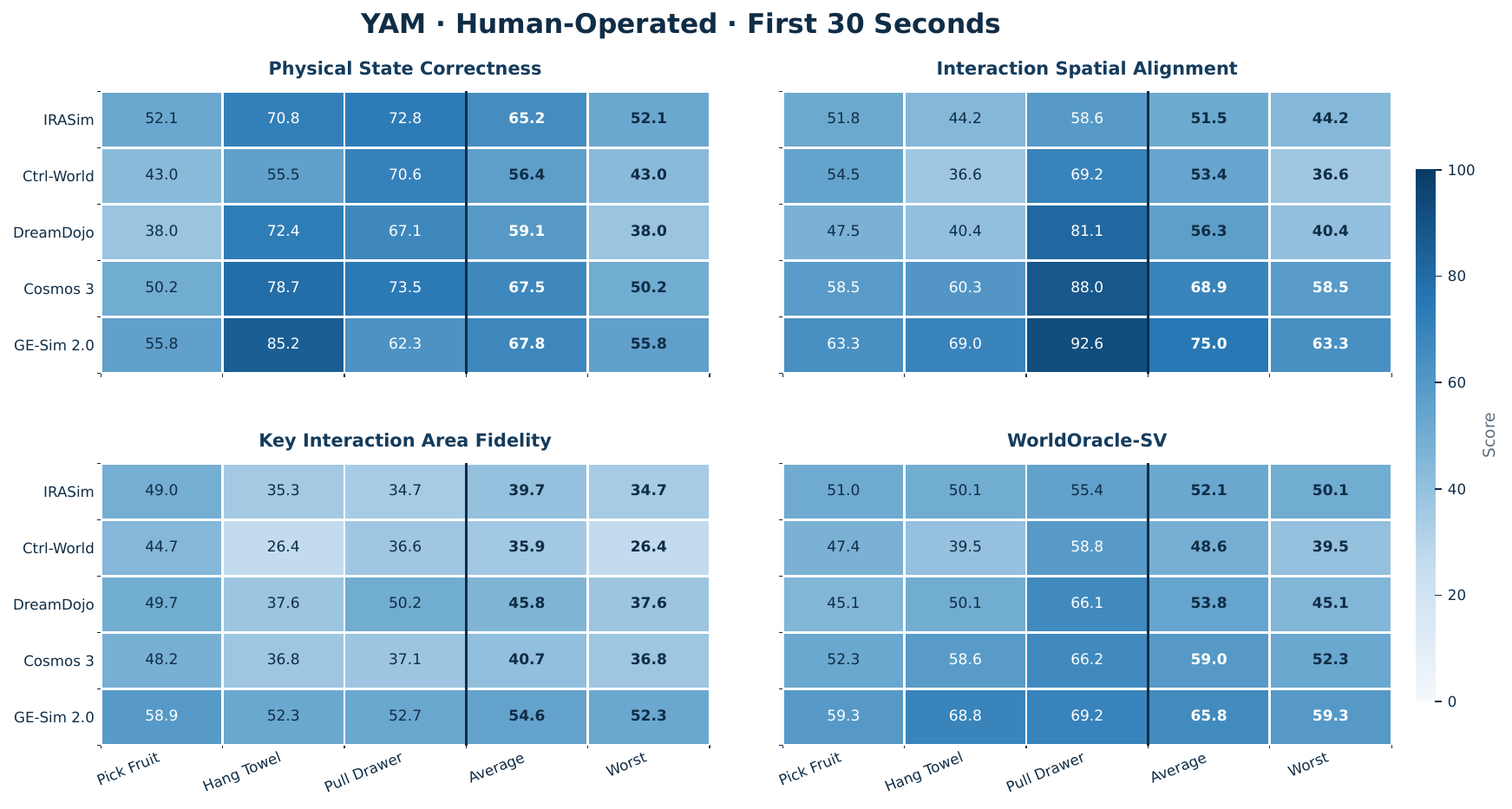}
    \caption{
    \textbf{Task-wise interaction-dynamics analysis on human-operated YAM split.}
    Results are evaluated over the first 30 seconds on \physical, \spatial, \keyarea, and WorldOracle-SV. In each heatmap, \emph{Average} denotes the arithmetic mean across tasks, while \emph{Worst} denotes the lowest task score.
    }
    \label{fig:interaction_dynamics_yam}
\end{figure*}
\begin{figure*}[!t]
    \centering
    \includegraphics[width=\textwidth]{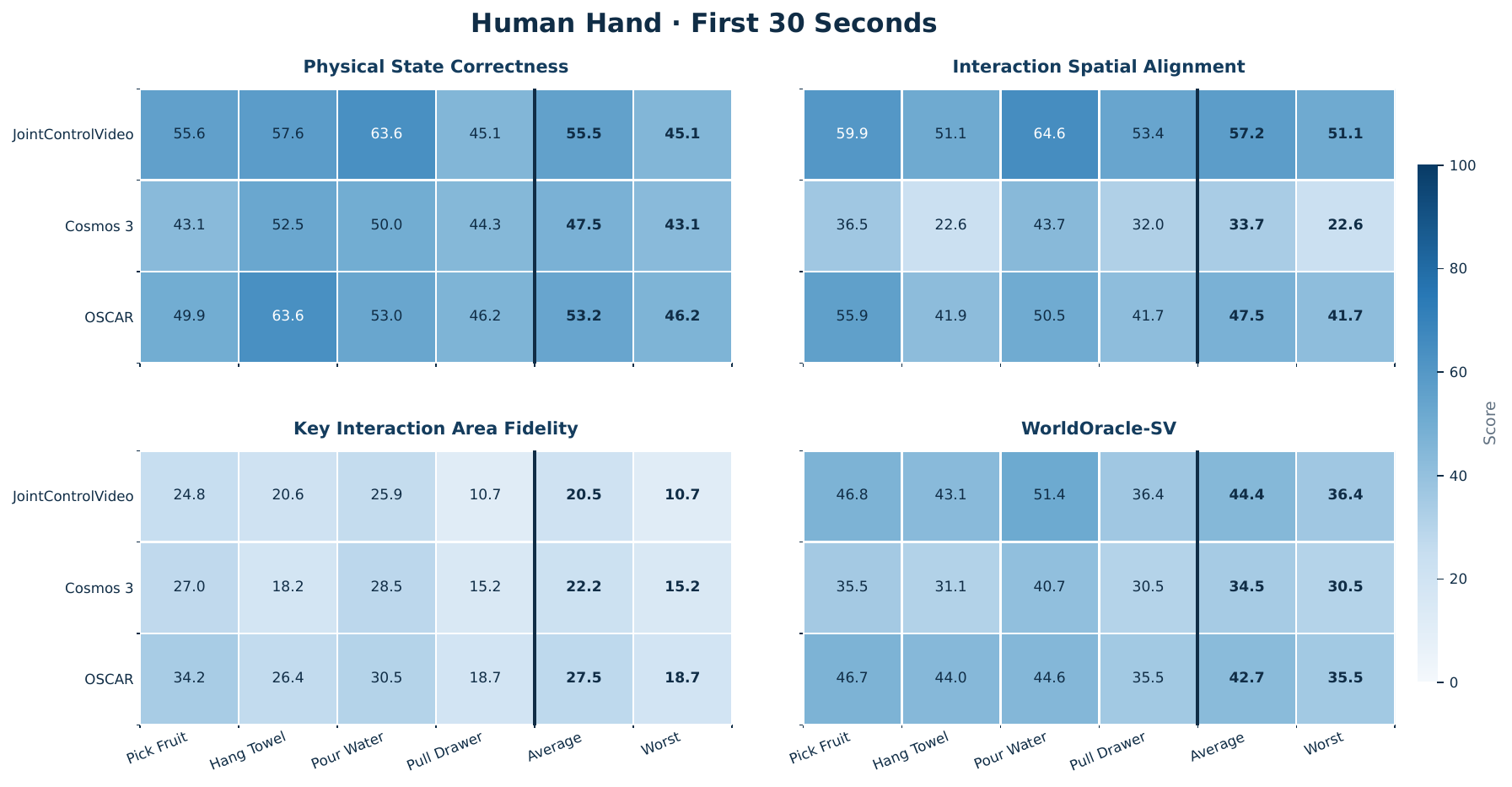}
    \caption{
    \textbf{Task-wise interaction-dynamics analysis for human-hand
    split.}
    Results are evaluated over the first 30 seconds on \physical, \spatial, \keyarea, and WorldOracle-SV. In each heatmap, \emph{Average} denotes the arithmetic mean across tasks, while \emph{Worst} denotes the lowest task score.
    }
    \label{fig:interaction_dynamics_human_hand}
\end{figure*}
\begin{figure*}[!t]
    \centering
    \includegraphics[width=\textwidth]{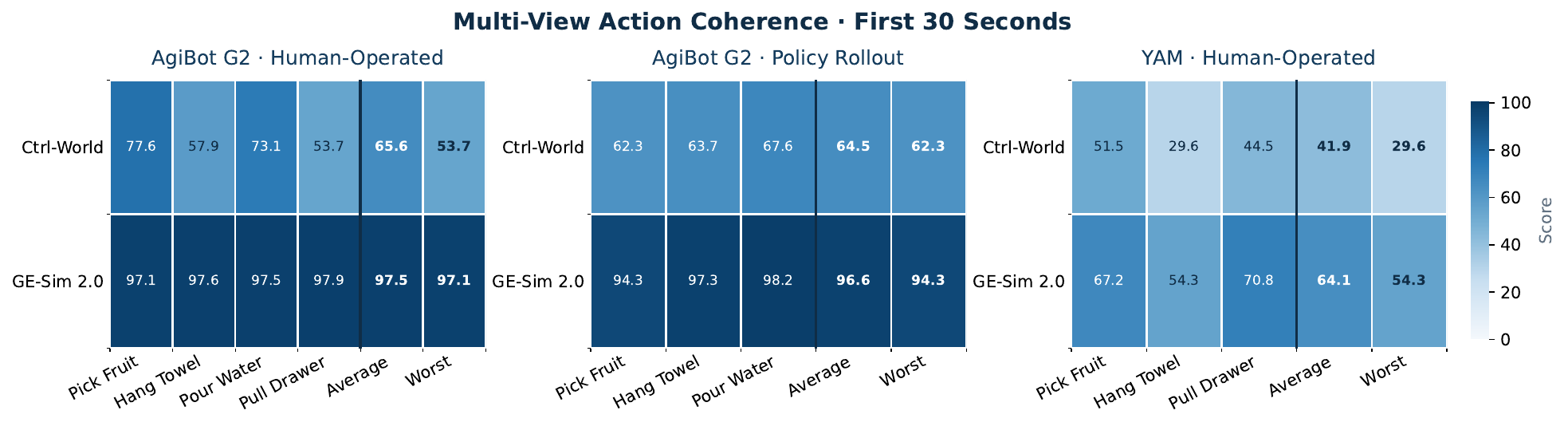}
    \caption{
    \textbf{Task-wise interaction-dynamics analysis for \multiview.}
    Results are evaluated over the first 30 seconds. In each heatmap, \emph{Average} denotes the arithmetic mean across tasks, while \emph{Worst} denotes the lowest task score.
    }
    \label{fig:interaction_dynamics_multiview_30s}
\end{figure*}

\begin{figure*}[!t]
    \centering
    \includegraphics[width=\textwidth]{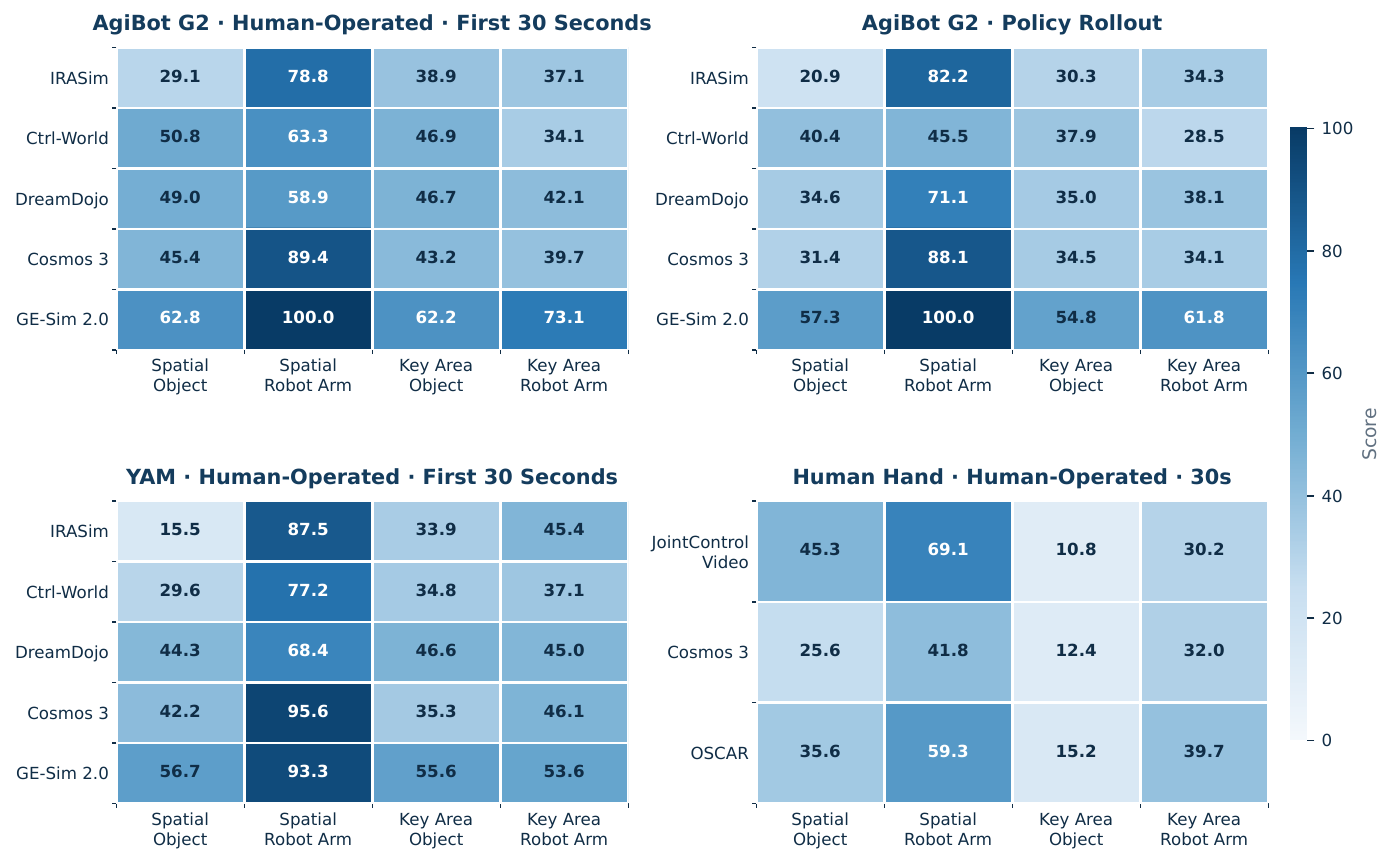}
    \caption{
    \textbf{Entity-centric error decomposition.}
    \spatial and \keyarea are decomposed into scores computed over the object and robot-arm/hand masks, separately. In the figure, \emph{Spatial} denotes \spatial, while \emph{Key Area} denotes \keyarea. Each cell reports the score for the corresponding metric and mask category.
    }
    \label{fig:entity_error_decomposition}
\end{figure*}

\begin{figure}[t]
    \centering
    \includegraphics[width=\linewidth]{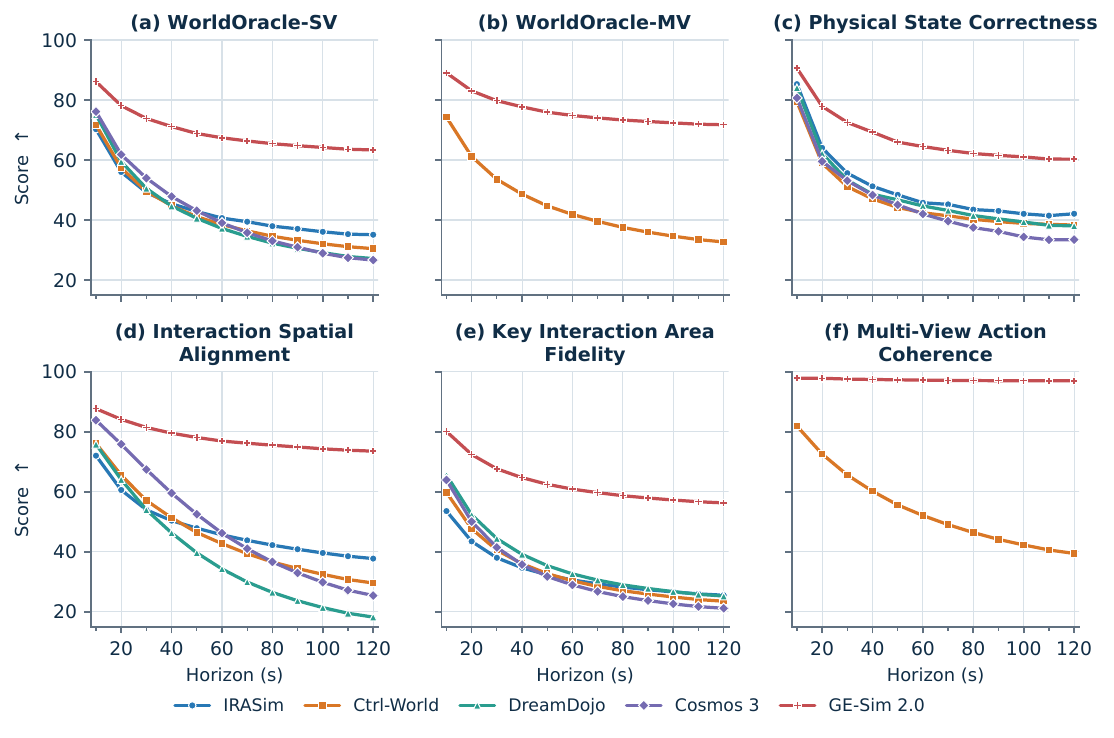}
    \caption{
    \textbf{Performance across cumulative rollout horizons on the
    human-operated AgiBot G2 distribution.}
    Each point reports the corresponding metric evaluated over the cumulative interval $[0,T)$, for horizons $T$ ranging from 10 to 120 seconds. Higher scores indicate better performance.
    }
    \label{fig:worldoracle_agibot}
\end{figure}
\begin{figure}[t]
    \begin{center}
\centerline{\includegraphics[width=1\linewidth]{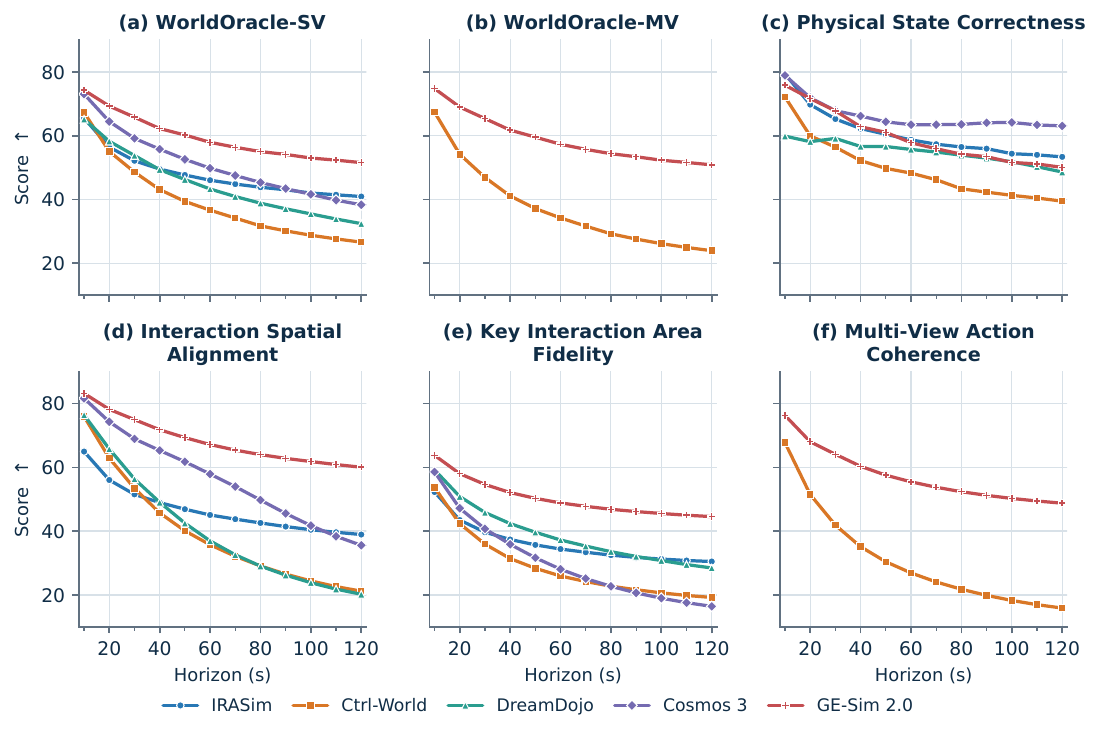}}
    \caption{\textbf{Performance across cumulative rollout horizons on the human-operated YAM distribution.}
    Each point reports the corresponding metric evaluated over the cumulative interval $[0,T)$, for horizons $T$ ranging from 10 to 120 seconds. Higher scores indicate better performance.
    }
    \label{fig:worldoracle_yam}
    \end{center}
\end{figure}

\begin{figure}[t]
    \begin{center}
\centerline{\includegraphics[width=1\linewidth]{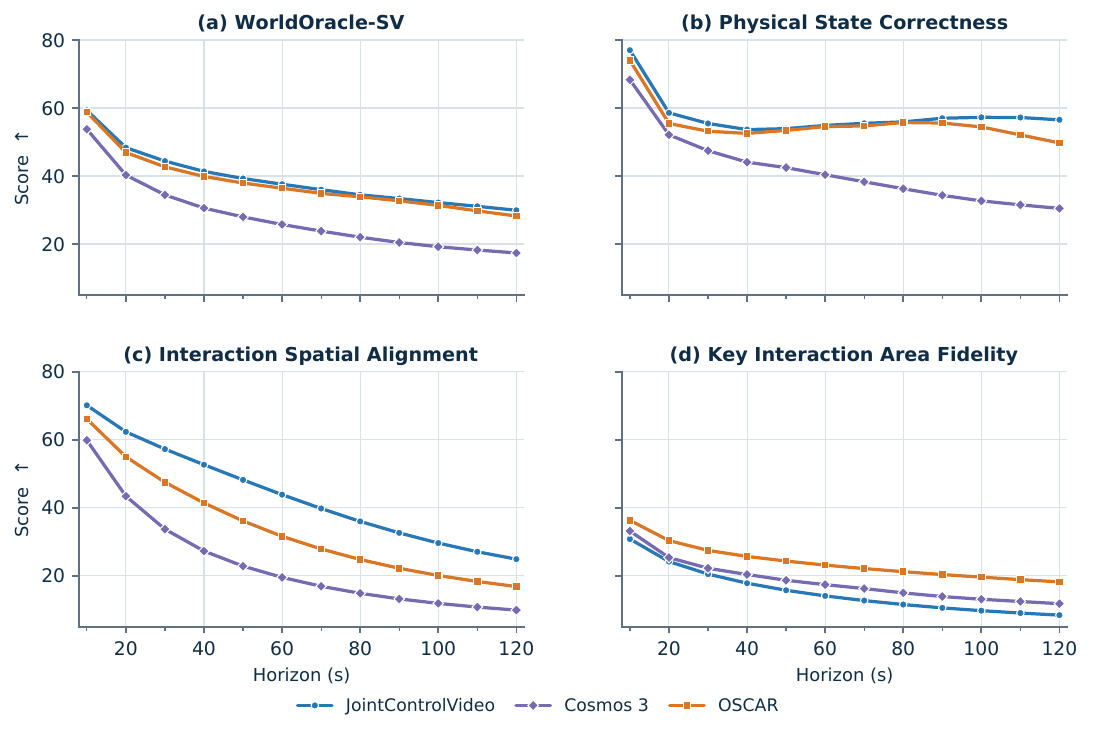}}
    \caption{\textbf{Performance across cumulative rollout horizons on the human-hand distribution.} Each point reports the corresponding metric evaluated over the cumulative interval $[0,T)$, for horizons $T$ ranging from 10 to 120 seconds. Higher scores indicate better performance.
    }
    \label{fig:worldoracle_ego}
    \end{center}
\end{figure}

\subsection{Analysis across Interaction Dynamics}
\label{sec:appendix_interaction_dynamics}

We further decompose model performance across tasks with different interaction dynamics. We evaluate the first 30 seconds for the human-operated distribution and complete episodes for the policy-rollout distribution. Takeaways are described below.

\paragraph{Pixel-aligned conditioning is associated with stronger action following.} Fig.~\ref{fig:interaction_dynamics_agibot_human} shows that GE-Sim~2.0
maintains a clear advantage across all 4 AgiBot tasks. Its average \metricsv score reaches $73.9$, compared with $54.0$ for the strongest alternative. The advantage is particularly large in \spatial ($81.4$) and \keyarea ($67.6$). One possible explanation is its pixel-aligned action conditioning mechanism, which provides an explicit correspondence between the specified action and the image regions in which the interaction should happen.

The task-level results also reveal differences among interaction dynamics. Models perform generally well on \textit{pour water} in terms of \physical, \keyarea, and \metricsv, and yield higher \spatial scores on \textit{pull drawer}. Although liquid transfer is intrinsically challenging, the relatively large cup and contact region may simplify grasp localization compared with tasks involving smaller objects or more precise contact. Similarly, while articulated-object dynamics introduce additional physical complexity, the visually salient drawer and its constrained motion may make the interaction region easier to localize.

\paragraph{Accurate action following does not guarantee correct physical interactions.} The results in Fig.~\ref{fig:interaction_dynamics_agibot_policy} reveal a distinction
between following the commanded trajectory and producing the intended
physical outcome. IRASim achieves the highest average \physical score
($62.2$), slightly exceeding GE-Sim~2.0 ($57.0$). In contrast, GE-Sim~2.0 substantially outperforms IRASim in \spatial ($78.7$ versus $51.6$). This pattern is consistent with pixel-aligned action maps providing useful spatial grounding for robot-trajectory following. However, reaching the correct interaction region does not ensure that the physical interaction is coherent with the real world. 

We also find that \textit{pick fruit} is the weakest task for GE-Sim~2.0 and other models. This suggests that precise interaction with small objects remains challenging even under pixel-aligned control.

\paragraph{The advantage narrows across embodiments.} Fig.~\ref{fig:interaction_dynamics_yam} shows that GE-Sim~2.0 remains the
strongest model after transfer to the YAM embodiment, achieving an average \metricsv score of $65.8$. However, its margin over the strongest alternative decreases from $19.9$ points on AgiBot G2 (Fig.~\ref{fig:interaction_dynamics_agibot_human}) to $6.8$ points on
YAM. This reduction is consistent with the domain shift. The GE-Sim~2.0
backbone is pretrained on AgiBot G1, whose morphology and visual appearance are closer to those of AgiBot G2 than YAM. Nevertheless, GE-Sim~2.0 retains the highest \spatial ($75.0$) and \keyarea ($54.6$) scores, suggesting that pixel-aligned action conditioning might be beneficial across embodiments. The narrower advantage indicates that action grounding alone does not fully resolve embodiment shift and motivates broader embodiment pretraining or explicit cross-embodiment adaptation.

\paragraph{Human-hand models trade off action following and local fidelity.} Fig.~\ref{fig:interaction_dynamics_human_hand} reveals different failure modes among the human-hand models. JointControlVideo achieves the highest average \spatial score ($57.2$), indicating comparatively strong action following, but obtains the lowest \keyarea score ($20.5$), suggesting poor visual fidelity within the interaction regions. Cosmos~3 exhibits the opposite tendency. Its \keyarea score ($22.2$) slightly exceeds that of JointControlVideo, but its substantially lower \spatial score ($33.7$) indicates weak correspondence between the conditioning signal and the generated interaction. OSCAR provides the best balance, combining a relatively strong \spatial score ($47.5$) with the highest \keyarea score ($27.5$).

\paragraph{Multi-view coherence degrades under domain shift.} Fig.~\ref{fig:interaction_dynamics_multiview_30s} shows that both models perform substantially worse on YAM than on AgiBot G2. GE-Sim~2.0 decreases from an average \multiview of $97.5$ on human-operated AgiBot trajectories to $64.1$ on YAM, while Ctrl-World decreases from $65.6$ to $41.9$. This degradation likely reflects the substantial domain gap between the ABC-130K training data and the YAM evaluation distribution, including differences in tasks, scene settings, and camera observations. These results indicate that maintaining synchronized multi-view predictions remains challenging under cross-embodiment transfer, especially when there is a huge domain gap.

\subsection{Entity-Centric Error Decomposition}
\label{sec:entity_centric_decomposition}

We decompose \spatial and \keyarea scores by the manipulated object and the robot arm (or human hand). We report results over the first 30 seconds for the human-operated distribution and over the complete episode for the policy-rollout distribution. Takeaways are described below.

\paragraph{Object dynamics are harder than arm motion.} The \spatial scores in Fig.~\ref{fig:entity_error_decomposition} reveal a consistent gap between robot-arm and object alignment. For example, on human-operated AgiBot trajectories, IRASim scores $78.8$ on the arm but only $29.1$ on the object, while Cosmos~3 scores $89.4$ and $45.4$, respectively. This pattern holds for all the evaluated models across all the embodiments and data distributions. This indicates that a model may accurately follow or render the specified action sequence without reproducing the plausible object motion. Modeling physical interactions and outcomes that are consistent with the real world remains a central challenge.

\paragraph{Human-hand interactions exhibit weaker interaction grounding.} Compared with the robot embodiments, all human-hand models achieve substantially lower scores, particularly for \keyarea. Even the strongest human-hand model reaches only $15.2$ on object fidelity and $39.7$ on hand fidelity, while the best robot models obtain considerably higher scores. Human hands introduce highly articulated motion, deformable appearance, and fine-grained contact, which may make action grounding and local interaction rendering more difficult than in the robot settings. Accurately modeling hand--object interactions therefore remains especially challenging for current world models.

\subsection{Temporal Degradation across Cumulative Rollout Horizons}
\label{sec:stagewise_temporal_endurance}

We evaluate temporal degradation by computing each metric over cumulative intervals $[0,T)$, where the rollout horizon $T$ increases from 10 to 120 seconds. A robust world model should retain high scores as autoregressive generation extends to longer horizons. Based on results shown in Fig.~\ref{fig:worldoracle_agibot}, Fig.~\ref{fig:worldoracle_yam}, and Fig.~\ref{fig:worldoracle_ego}, we summarize the key observations below.

\paragraph{AgiBot G2.}
Fig.~\ref{fig:worldoracle_agibot} shows that GE-Sim~2.0 maintains a clear lead across all cumulative horizons. Its \metricsv score decreases from $86.1$ at 10 seconds to $63.4$ at 120 seconds, while other methods finish between $26.7$ and $35.1$. Among the remaining models, Cosmos~3 achieves the highest \metricsv score through 60 seconds but degrades rapidly thereafter. IRASim overtakes it at 70 seconds. This indicates that strong short-horizon performance does not necessarily imply temporal endurance.

\paragraph{YAM.} Overall, models perform worse on YAM than on AgiBot G2, indicating greater sensitivity to the train--test domain gap. For YAM, we rely on public training data whose scenes, visual appearance, and trajectory distribution differ substantially from our evaluation set. In contrast, the AgiBot training data are collected specifically for the benchmark and therefore more closely match the corresponding evaluation distribution. These results suggest that world-model performance decreases as the domain gap between training and evaluation increases.

GE-Sim~2.0 exhibits an additional cross-embodiment gap, particularly in
\physical and \multiview. Its backbone is pretrained on AgiBot G1, whose morphology and visual appearance are closer to AgiBot G2 than to YAM. Therefore, a larger embodiment shift is associated with weaker performance in this setting. This result indicates that strong performance within a closely related robot family does not necessarily transfer to a morphologically different embodiment.

\paragraph{Human hand.}
As shown in Fig.~\ref{fig:worldoracle_ego}, human-hand models perform worse than models evaluated on the two robot embodiments. We attribute this gap to two factors. First, the models are evaluated zero-shot on the human-hand distribution. Second, human-hand interactions involve highly articulated motion, deformable appearance, and fine-grained contact, making them intrinsically more difficult to model than fixed-morphology robot interactions.

Across all three models, \physical is considerably more durable than
\spatial and \keyarea. For example, JointControlVideo retains a \physical score of $56.5$ at 120 seconds, whereas its \spatial and \keyarea scores decrease to $24.9$ and $8.4$, respectively. Thus, although current models can preserve coarse physical plausibility over long horizons, accurately grounding and rendering fine-grained hand--object interactions remains a key challenge.
\begin{figure}[!t]
    \centering
    \includegraphics[width=0.98\linewidth]{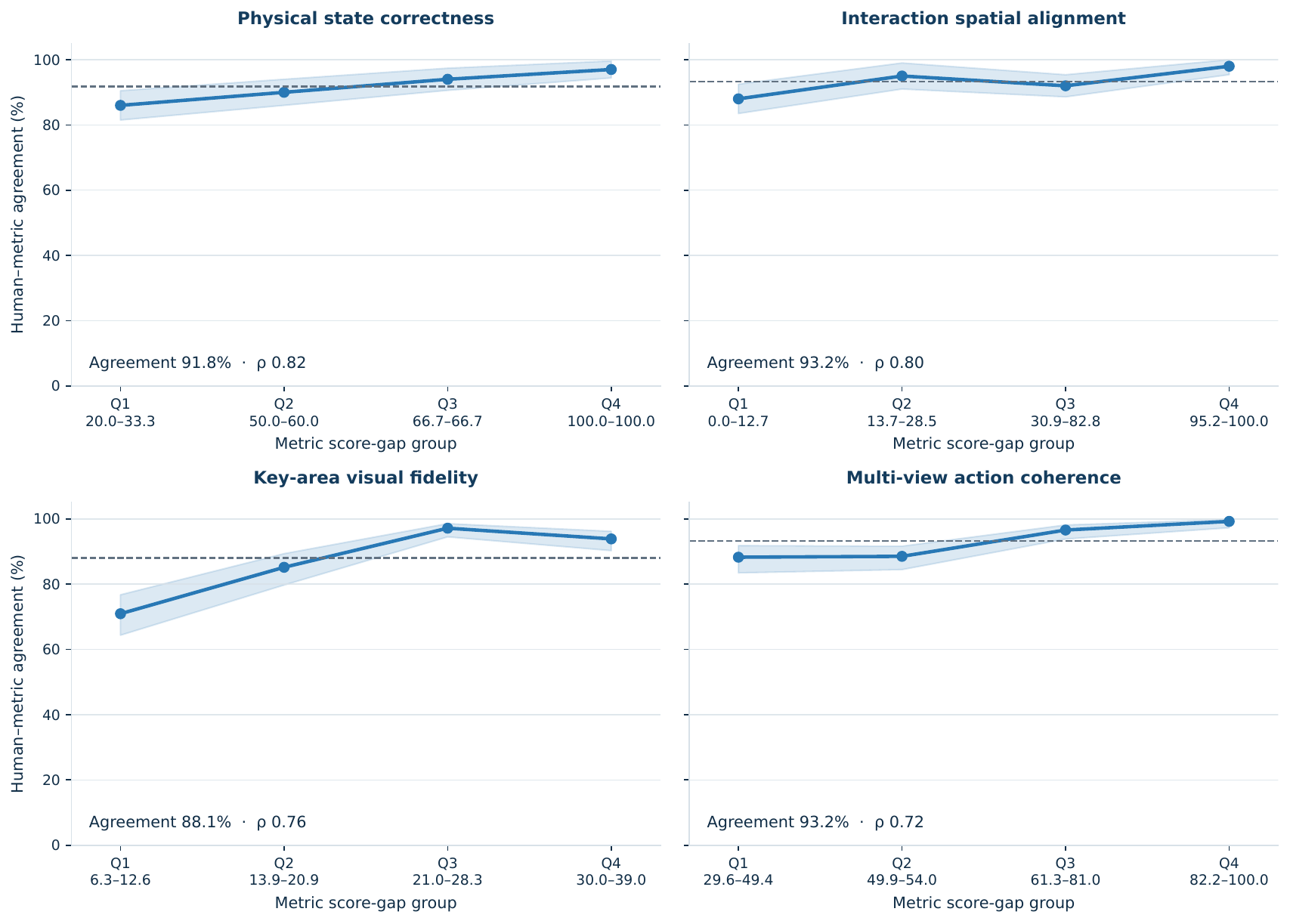}
    \caption{
    \textbf{Alignment between WorldOracle metrics and human judgments.}
    Each panel reports the fraction of human choices that agree with the
    preference induced by the corresponding metric, grouped by metric score
    gap. Shaded regions denote $95\%$ Wilson confidence intervals, and dashed
    horizontal lines indicate overall agreement. The reported Spearman
    coefficient $\rho$ measures the pair-level association between the signed
    metric-score difference and the human preference rate.
    }
    \label{fig:human_alignment}
\end{figure}

\section{Validation of WorldOracle Metrics}
\label{app:metric_validation}

\subsection{Human--Metric Alignment Study}
\label{sec:human_metric_alignment}

We conduct a human study to examine whether preferences induced by the \modelfull metrics agree with human judgements. The study contains 128 prediction pairs spanning the AgiBot human-operated, AgiBot policy-rollout, YAM, and human-hand distributions. We collect responses from 119 participants. Each trial presents two anonymous predictions in randomized order together with the real-world reference, and participants select the prediction that better satisfies the corresponding evaluation dimension.

\paragraph{WorldOracle metrics align consistently with human judgments.}
As shown in Fig.~\ref{fig:human_alignment}, all 4 WorldOracle metrics
achieve high overall agreement with human preferences, ranging from $88.1\%$ to $93.2\%$. Importantly, strong agreement is observed across different metric score-gap groups, indicating that the alignment is not driven only by a small number of easily distinguishable prediction pairs. Even within the group with small score gap, agreement remains substantially above chance for all 4 evaluation dimensions.

The strong pair-level Spearman correlations ($\rho=0.72$--$0.82$) further show that larger metric differences correspond to more decisive human preferences. Together, these results demonstrate that \modelfull metrics align with human preference across both difficult and clearly distinguishable comparisons.

\begin{figure*}[!t]
    \centering
    \includegraphics[width=\textwidth]{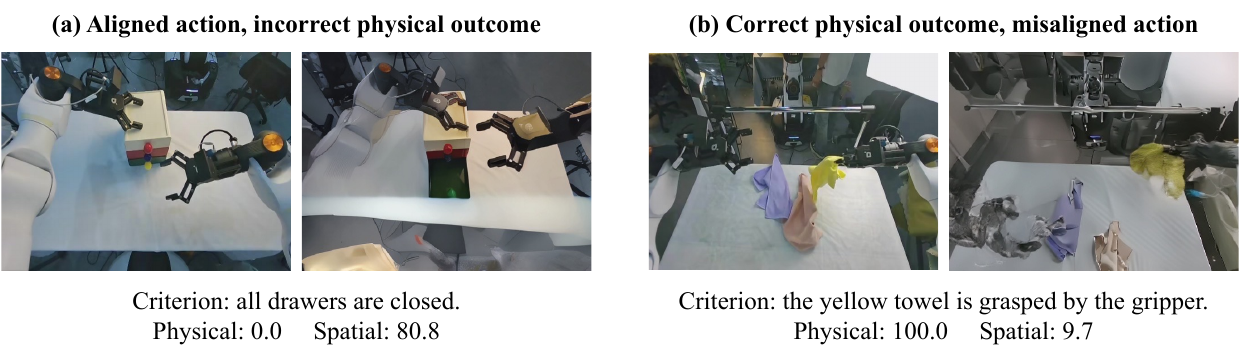}
    \caption{
    \textbf{Complementary failure modes captured by \physical and \spatial.}
    Each panel compares the real and predicted images at the annotated
    outcome frame. \emph{Physical} is the criterion-level \physical score at this frame, while \emph{Spatial} is the \spatial score computed over a four-second local window centered on it. In (a), the prediction approximately follows the robot motion but incorrectly leaves the green drawer open when the real execution closes all drawers. In (b), the model predicts that the yellow towel is grasped, satisfying the stated physical criterion, but renders the arm and towel along a trajectory that differs from the real execution.
    }
    \label{fig:physical_spatial_case_studies}
\end{figure*}

\subsection{Metric Case Studies}
\label{sec:metric_case_studies}

\paragraph{Outcome correctness and action alignment are complementary.} Fig.~\ref{fig:physical_spatial_case_studies} illustrates both directions of the distinction between \physical and \spatial. In the pull-drawer example, the predicted arm motion remains well aligned with the real execution (\spatial $=80.8$), but the generated drawer state contradicts the annotated outcome that all drawers are closed (\physical $=0.0$). Conversely, in the hang-towel example, the prediction satisfies the criterion that the yellow towel is grasped (\physical $=100.0$), while the generated arm and towel trajectory is poorly aligned with the real execution (\spatial $=9.7$). Thus, neither metric subsumes the other. Spatial action following does not guarantee the correct physical consequence, and matching an outcome criterion does not guarantee a faithful execution trajectory.

\begin{figure*}[t]
    \centering
    \includegraphics[width=\textwidth]{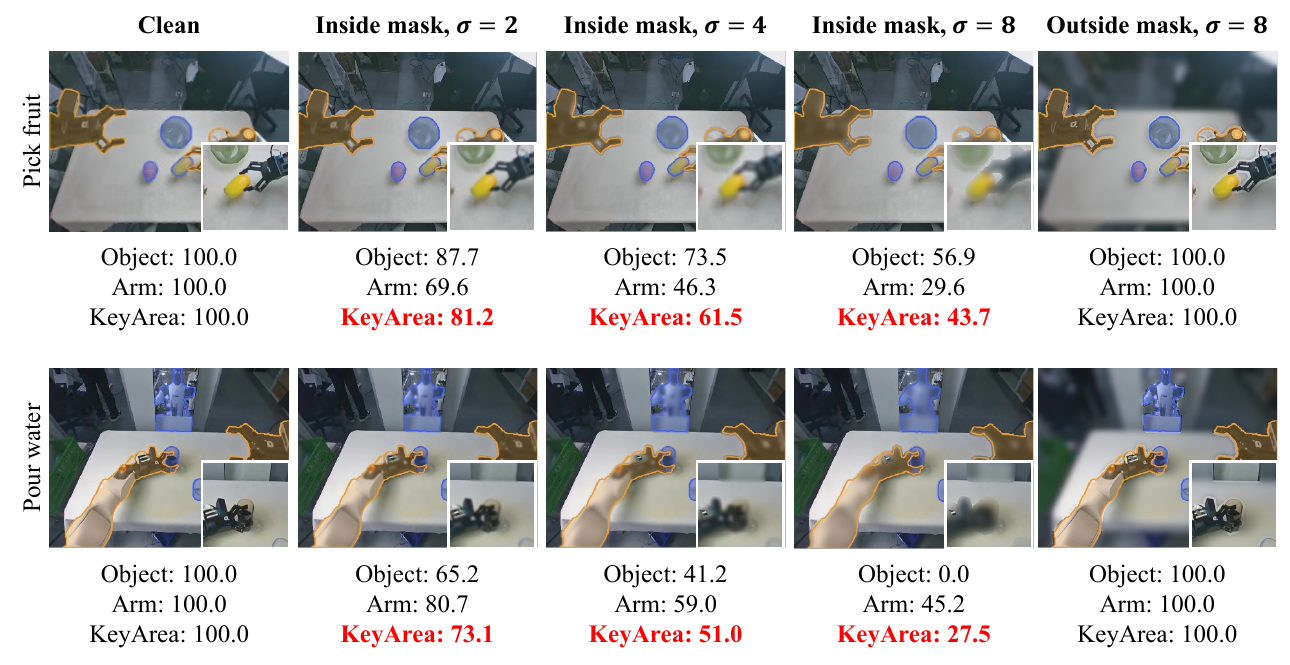}
    \caption{
    \textbf{Controlled case study of \keyarea.}
    Starting from clean real interaction frames, we apply Gaussian blur either within the object and robot-arm masks ($\sigma\in\{2,4,8\}$) or only outside this region ($\sigma=8$). Blue and orange contours denote the object and robot-arm masks, respectively. Each panel reports the object score, arm score, and their aggregate \keyarea score. Increasing mask-local blur produces a monotonic decrease
    in \keyarea for both an object-centric Pick Fruit example and a
    contact-centric Pour Water example, whereas applying the strongest blur only to the background leaves the score unchanged.
    }
    \label{fig:keyarea_blur_case_study}
\end{figure*}
\begin{figure}[!t]
    \centering
    \includegraphics[width=\linewidth]{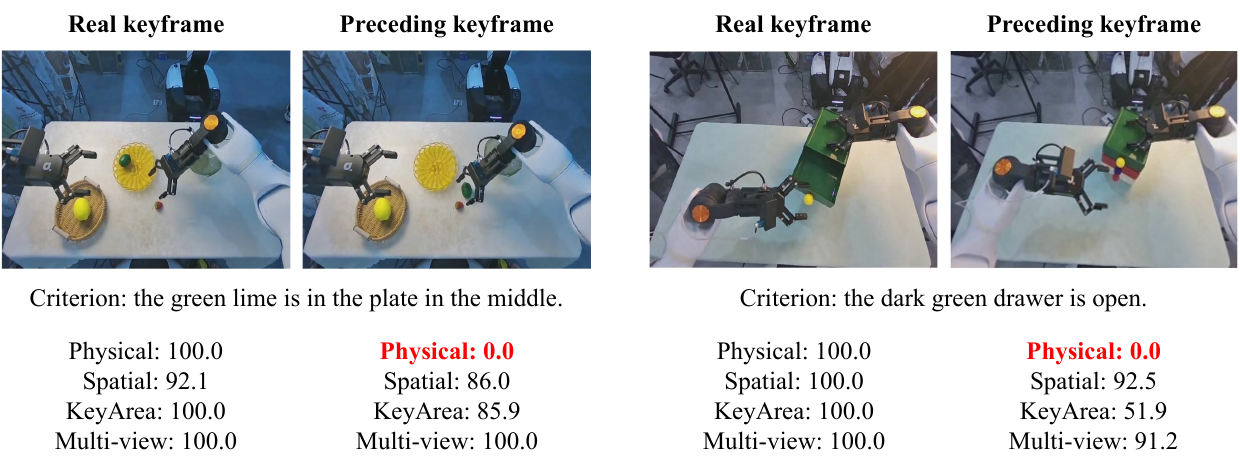}
    \caption{
    \textbf{Qualitative examples of the outcome-swap intervention.}
    Although both frames come from the same episode, the
    preceding keyframe contradicts the newly introduced physical-state criterion. Each example reports all four metric scores.
    }
    \label{fig:outcome_swap_qualitative}
\end{figure}

\paragraph{Key-area fidelity is sensitive to local interaction quality.}
Fig.~\ref{fig:keyarea_blur_case_study} isolates the spatial locality of \keyarea using controlled corruptions of two clean reference frames.
For Pick Fruit, the aggregate score decreases from $100.0$ on the clean
frame to $81.2$, $61.5$, and $43.7$ as the mask-local blur increases from $\sigma=2$ to $4$ and $8$. The Pour Water example exhibits the same
monotonic trend, decreasing from $100.0$ to $73.1$, $51.0$, and $27.5$.
In contrast, severe background-only blur leaves both examples at $100.0$
because the evaluated interaction pixels remain unchanged. These cases show that \keyarea responds to progressively degraded object and contact
appearance while remaining localized to the interaction regions it is
designed to evaluate.

\subsection{Controlled Metric Selectivity}

\paragraph{Outcome swapping primarily affects \physical.}
We construct a balanced event-level set containing ten state transitions from each task. To ensure that the intervention tests an actual state change, we retain only keyframes whose physical criterion differs from the preceding keyframe and evaluate the newly introduced criterion. The perturbed sample replaces the original keyframe with the preceding keyframe from the same episode, while retaining the original outcome keyframe as the reference. Fig.~\ref{fig:outcome_swap_qualitative} shows two representative interventions: the swapped keyframe depicts a similar scene layout and robot trajectory but different fruit placement or drawer-state transition.

Across the 40 events, we compute the relative decrease as
$100(S_{\mathrm{clean}}-S_{\mathrm{swap}})/S_{\mathrm{clean}}$.
The intervention reduces \physical by $67.9\%$ on average, substantially more than \keyarea ($34.3\%$), \spatial ($20.0\%$), or \multiview ($12.3\%$). Replacing an outcome keyframe also changes local object appearance and robot--object geometry. Nevertheless, the largest response occurs in the intended \physical metric, supporting its sensitivity to incorrect interaction outcomes.


\begin{figure*}[t]
  \centering
  \includegraphics[width=\textwidth]{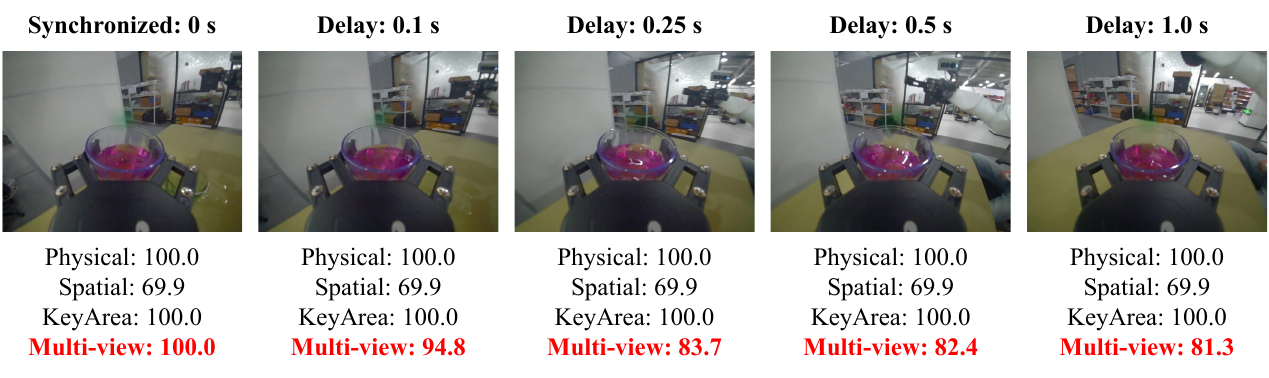}
  \caption{\textbf{Cross-view desynchronization across increasing delays.} We progressively delay only the left auxiliary view while retaining the top-head observation and right view at time $t$. All four scores are shown for each condition. In this example, \multiview decreases monotonically from $100.0$ to $94.8$, $83.7$, $82.4$, and $81.3$ as the delay increases, whereas the three top-head metrics remain unchanged.}
  \label{fig:crossview_desync_selectivity}
\end{figure*}

\paragraph{Cross-view desynchronization.}
We test whether \multiview detects temporal disagreement between camera streams by delaying exactly one auxiliary view while keeping the top-head view and the other auxiliary view synchronized. We independently perturb the left and right auxiliary views in 40 interaction events and average their effects. As the delay increases from $0.1$ to $0.25$, $0.5$, and $1.0$ seconds, the mean relative decrease in \multiview grows monotonically from $0.7\%$ to $1.7\%$, $2.9\%$, and $4.4\%$, respectively. At the medium $0.5$-second delay, the decrease is $2.9\%$. Because \physical, \spatial, and \keyarea use only the unchanged top-head input in this intervention, their scores remain exactly unchanged. This controlled result isolates the intended sensitivity of \multiview while demonstrating invariance of the single-view metrics to an off-target camera perturbation. Qualitative examples are shown in Fig.~\ref{fig:crossview_desync_selectivity}.

\begin{figure*}[!t]
  \centering
  \includegraphics[width=\textwidth]{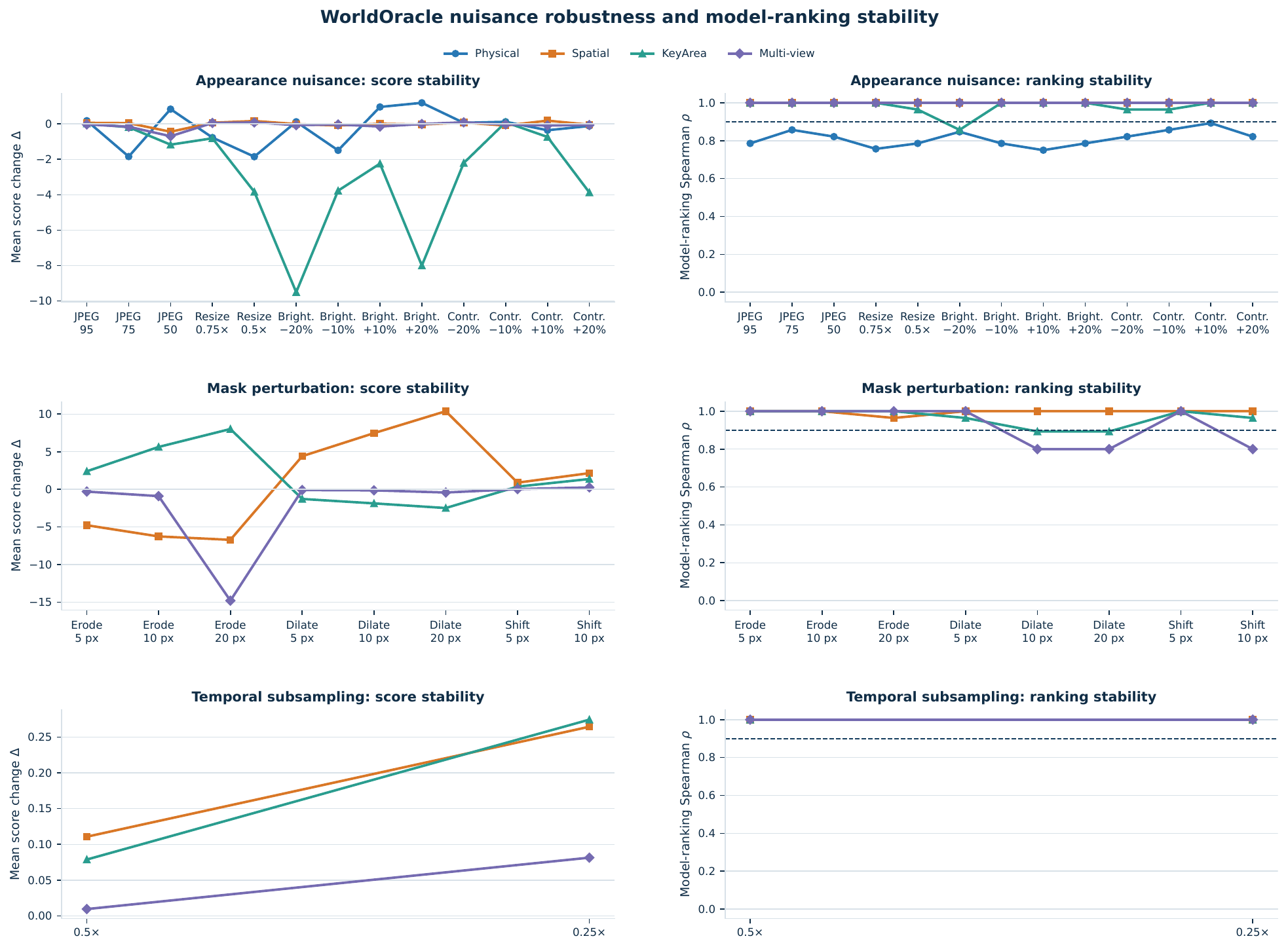}
  \caption{\textbf{Nuisance robustness and sensitivity.}
  Left panels show the signed mean score change relative to the default
  setting; right panels show model-ranking Spearman correlation. Appearance
  perturbations leave geometry-centered metrics nearly unchanged while
  \keyarea appropriately responds to visible degradation. Mask morphology
  changes absolute scores more strongly but generally preserves rankings.
  Temporal subsampling produces negligible score changes and identical
  rankings. The dashed line denotes $\rho=0.9$.}
  \label{fig:metric_nuisance_robustness}
\end{figure*}

\subsection{Metric Robustness and Stability}
\label{sec:metric_robustness}

\subsubsection{Nuisance Robustness and Model-Ranking Stability}

We test whether each metric is robust to irrelevant perturbations while remaining sensitive to the property it is designed to measure. The study contains 40 interaction events from four tasks and seven models, yielding 280 prediction--reference pairs. We report the mean score change from the unperturbed setting and the Spearman correlation between model rankings.

\paragraph{Appearance robustness.}
We apply JPEG compression, downsample--upsample degradation, and brightness and contrast changes. As shown in the top row of
Fig.~\ref{fig:metric_nuisance_robustness}, \spatial and \multiview are highly stable: their mean changes remain within $0.45$ and $0.70$ points, respectively, and their model rankings are unchanged ($\rho=1.00$). Physical state correctness also changes by at most $1.9$ points on average, although its ranking stability is lower ($\rho=0.75$--$0.89$). In contrast, \keyarea appropriately responds to visible degradation: $0.5\times$ resizing reduces its score by $3.8$ points, while $\pm20\%$ brightness changes reduce it by $8.0$--$9.5$ points. Its rankings remain largely stable ($\rho=0.86$--$1.00$).

\paragraph{Mask robustness.}
We perturb object and robot masks through erosion, dilation, and translation. As shown in the middle row of
Fig.~\ref{fig:metric_nuisance_robustness}, translations of 5 pixels change the metric means by at most $0.89$ points,
while 10-pixel translations change them by at most $2.15$ points.
Erosion and dilation have larger effects because they directly remove
interaction pixels or introduce surrounding context. Despite these absolute
score changes, model rankings remain stable for \spatial
($\rho\geq0.96$) and \keyarea ($\rho\geq0.89$). The lowest \multiview correlation is $0.80$.

\paragraph{Temporal sampling.}
As shown in the bottom row of
Fig.~\ref{fig:metric_nuisance_robustness}, reducing the sampling frequency to $0.25\times$ changes the mean scores by
only $0.26$ points for \spatial, $0.27$ for \keyarea, and $0.08$ for
\multiview. All model rankings remain identical to the default evaluation
($\rho=1.00$). The benchmark conclusions are therefore insensitive to
moderate temporal subsampling.

\paragraph{Overall findings.}
The results support the robustness of the proposed metrics: geometry-based metrics remain stable under appearance changes, \keyarea responds to the visual degradation it is intended to measure, and model rankings are generally preserved under mask noise and temporal subsampling.

\begin{figure*}[t]
    \centering
    \includegraphics[width=0.98\textwidth]
    {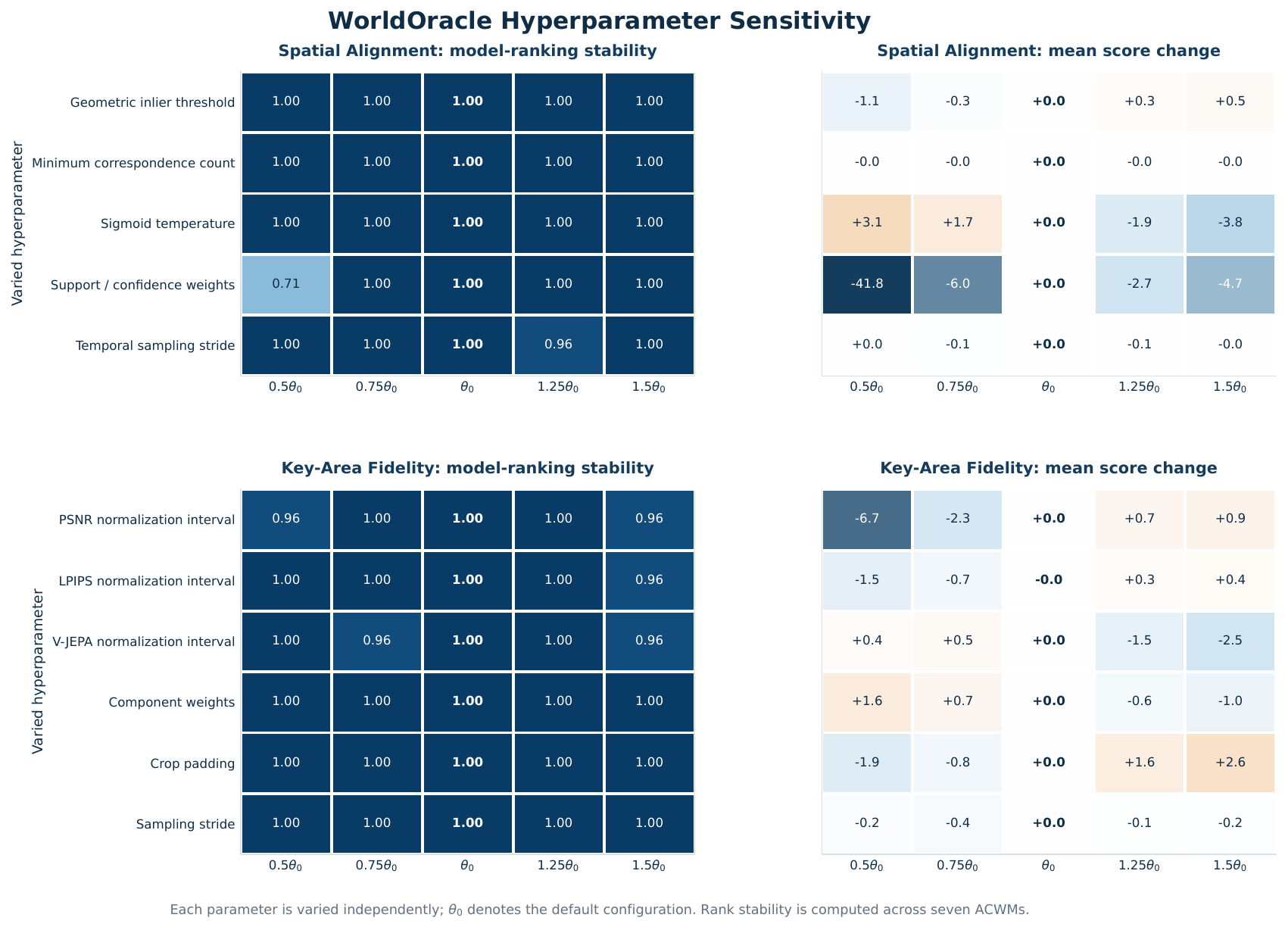}
    \caption{
    \textbf{Hyperparameter sensitivity of Spatial Alignment and
    Key-Area Fidelity.}
    We vary one setting at a time from $0.5\theta_0$ to
    $1.5\theta_0$, where $\theta_0$ is the default.
    The left panels show agreement with the default ranking of seven
    ACWMs; the right panels show the average score change.
    }
    \label{fig:hyperparameter_sensitivity}
\end{figure*}

\subsubsection{Hyperparameter and ranking stability.}

We vary the main thresholds, weights, crop padding, normalization
ranges, and sampling strides of both \spatial and \keyarea metrics. The main study contains 280 model--event pairs from seven ACWMs. Sampling stride is tested on an additional dense subset of 56 pairs.

As shown in Fig.~\ref{fig:hyperparameter_sensitivity}, the model
rankings remain nearly unchanged under most settings. For \spatial (denoted as ``Spatial Alignment'' in the figure), the geometric threshold, correspondence count, and sampling stride change the mean score by at most $1.1$ points. The only clear failure occurs when the feature-evidence weights are halved: the mean drops by $41.8$ points and rank agreement decreases to $\rho=0.71$. At $0.75\theta_0$ and above, the original ranking is fully recovered.

Key interaction area fidelity (denoted as ``Key-Area Fidelity'' in the figure) is also stable, with $\rho\geq0.96$ in every test. Its largest shift is $-6.7$ points when the PSNR normalization range
is halved; component weights, crop padding, and sampling stride all preserve the model ranking.

\begin{figure*}[t]
    \centering
    \includegraphics[width=0.9\textwidth]{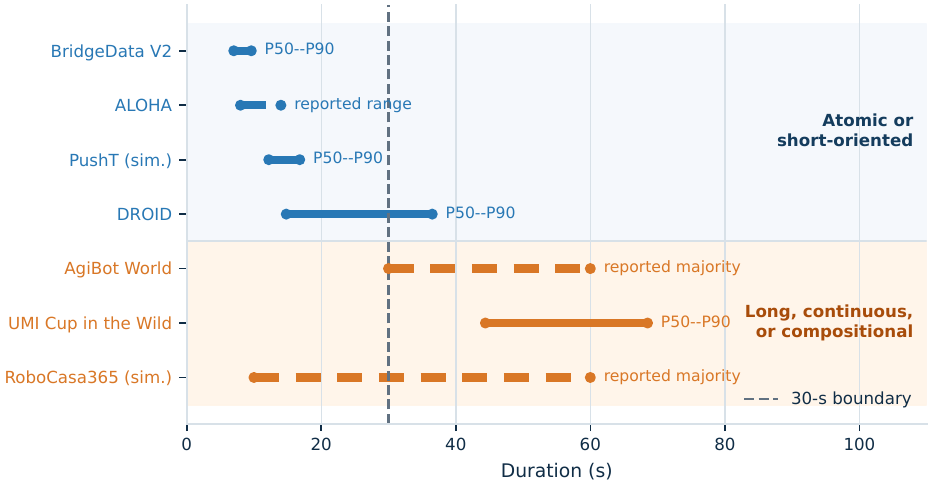}
    \caption{
    \textbf{Duration statistics of public robot-manipulation
    datasets.}
    Blue rows denote short, atomic datasets, whereas orange
    rows denote long, compositional datasets.
    Each row is annotated with its available timing statistic:
    an episode-level P50--P90 interval, or a reported majority range.
    Simulated datasets are marked with ``sim.'' explicitly. The dashed line indicates our selected 30-second short-horizon boundary.
    }
    \label{fig:public_dataset_duration}
\end{figure*}
\section{Selection of the Short-Duration Boundary}
\label{app:short_horizon}

We select the short-duration boundary based on public robot-manipulation datasets. For datasets with episode-level metadata, duration is computed as the number of recorded frames divided by the nominal frame rate. When only aggregate timing information is available, we use the range reported by the corresponding paper.

As shown in Fig.~\ref{fig:public_dataset_duration}, many atomic
manipulation datasets operate primarily below 30 seconds. For example, BridgeData V2~\citep{walke2023bridgedatav2} has a P50--P90 interval of \(7.0\)--\(9.6\) seconds, and ALOHA~\citep{zhao2023learningaloha} reports episodes lasting \(8\)--\(14\) seconds. PushT~\citep{chi2025diffusionpusht} has a P50--P90 interval of \(12.2\)--\(16.8\) seconds. DROID~\citep{khazatsky2024droid} is broader, with a median of \(14.8\) seconds, and \(84.9\%\) of its episodes are no longer than 30 seconds. In contrast, datasets containing longer, continuous, or compositional behavior extend beyond this boundary. AgiBot World~\citep{bu2025agibotworld} reports that most episodes last \(30\)--\(60\) seconds, UMI Cup~\citep{chi2024universalUMI} has a P50--P90 interval of \(44.4\)--\(68.5\) seconds, and most RoboCasa365~\citep{nasiriany2026robocasa365} episodes span \(10\)--\(60\) seconds.

These observations place 30 seconds as an approximate transition
between short, atomic behaviors and longer, compositional sequences. We therefore evaluate the first 30 seconds as a complementary
short-duration regime while retaining full-episode evaluation as the long-horizon setting.

\section{Conclusion}
\label{sec:conclusion}

We introduce \modelfull, a benchmark for evaluating whether action-conditioned world models reproduce the physical consequences observed in the real world, given the same initial observations and action trajectories. Four metrics evaluate physical outcomes, spatial action alignment, interaction-region fidelity, and multi-view action coherence. Across seven ACWMs, we find that visually plausible rollouts can misrepresent object dynamics and interaction outcomes, even when robot-arm motion is accurately reproduced. Strong short-horizon performance also does not consistently translate into reliable long-horizon prediction. Human studies and controlled analyses support the validity and complementary roles of the metrics. Improving ACWMs with persistent action grounding and accurate modeling of object interactions remains an important challenge. We hope \modelfull provides a useful benchmark for measuring progress toward world models that better match what actually happens in the physical world.



\clearpage

\clearpage
\bibliographystyle{abbrvnat}
\bibliography{template/arxiv_v2/main}

\clearpage

\clearpage
\appendix

\renewcommand{\thesection}{\Alph{section}}
\renewcommand{\thesubsection}{\thesection.\arabic{subsection}}
\renewcommand{\thesubsubsection}
{\thesubsection.\arabic{subsubsection}}

\setcounter{figure}{0}
\setcounter{table}{0}
\renewcommand{\thefigure}{S\arabic{figure}}
\renewcommand{\thetable}{S\arabic{table}}

\setcounter{algorithm}{0}
\renewcommand{\thealgorithm}{S\arabic{algorithm}}

\setcounter{equation}{0}
\renewcommand{\theequation}{S\arabic{equation}}
\renewcommand{\theHequation}{appendix.\arabic{equation}}
\renewcommand{\theHfigure}{appendix.\arabic{figure}}
\renewcommand{\theHtable}{appendix.\arabic{table}}
\renewcommand{\theHalgorithm}{appendix.\arabic{algorithm}}

\raggedbottom
\makeatletter
\setlength{\@fptop}{0pt}
\makeatother
\section*{Appendix Overview}
This appendix is organized into 6 sections. We first present the benchmark design (Sec.~\ref{app:benchmark_task_definition}, Sec.~\ref{sec:appendix_data_curation}, and Sec.~\ref{app:annotation_pipeline}). We then describe the details of the evaluation metrics (Sec.~\ref{app:metric_details}), before providing training and implementation details (Sec.~\ref{app:training_implementation_details}) and limitations (Sec.~\ref{app:limitations}).

\section{Benchmark Task Definition}
\label{app:benchmark_task_definition}

The four tasks are designed to cover distinct interaction dynamics while retaining visually verifiable outcomes.

\paragraph{Pick fruit.} This task evaluates rigid-body manipulation. Each scene contains two to four fruits and one to three storage containers. The embodiment repeatedly transfers fruits between the tabletop and the containers. Each storage container may contain at most one piece of fruit. Successful segments include grasping a designated fruit and placing it in the intended location. Failure segments include empty grasps, drops, collisions, and fruits falling outside the reachable workspace.

\paragraph{Hang towel.} This task evaluates deformable-object interaction. Each scene contains two or three colored towels and a drying rack. The embodiment picks up a towel, hangs it on the rack, removes it, or returns it to the tabletop. Failure outcomes include empty grasps, dropping the towel, unsuccessful placement on the rack, and the towel falling onto the table or floor.

\paragraph{Pour water.} This task evaluates fluid interaction. Each scene contains two to four cups, at least one of which initially contains colored liquid. A typical successful segment consists of grasping a source cup, transferring its liquid into an empty cup, and returning the source cup to the table. Failures include empty grasps, dropped or tilted cups, incomplete transfer, and liquid spilling onto the tabletop. Colored liquid is used to make fluid transfer and spills more observable to the VLM evaluator. Each episode uses only one liquid color.

\paragraph{Pull drawer.} This task evaluates articulated-object interaction. Each scene contains a multi-tier drawer placed on the tabletop. The drawer is initialized either with all tiers closed or with one tier open. An open drawer may contain a clearly visible object. The embodiment interacts with a drawer tier by grasping its handle and opening or closing it. A successful segment requires the intended tier to reach the open or closed state. Failures include missing the handle, and losing contact during pulling or pushing.

\section{Additional Details on Data Curation}
\label{sec:appendix_data_curation}

Each episode contains multiple interaction segments, including both successful and failed manipulation attempts. Across episodes, we vary object identities, quantities, sizes, initial configurations, and illumination conditions.

\subsection{Training Suite}
\label{sec:appendix_training_suite}

The newly collected AgiBot G2 training split contains 1,779 episodes across the four benchmark tasks. Statistics are shown in Table~\ref{tab:app_training_set_statistics}. Each episode is recorded at 30 FPS and lasts approximately one minute. Within each task, the robot starts from a fixed initial pose, and only its two arms and grippers are controlled. Left- and right-arm executions are balanced during data collection. The training suite uses a controlled range of scene variations, including a restricted set of tablecloth colors, a fixed illumination color, and reduced sets of fruits, containers, cups, liquid colors, towels, and drawer configurations.

\begin{table}[t]
\centering
\small
\caption{
\textbf{AgiBot G2 training-set statistics.}
Duration is reported as mean $\pm$ standard deviation and
observed minimum--maximum in seconds.
}
\label{tab:app_training_set_statistics}
\setlength{\tabcolsep}{4pt}
\scalebox{1}{%
\begin{tabular}{lrrrrr}
\toprule
Task & Episodes & FPS & Mean $\pm$ SD (s) & Range (s) & Hours \\
\midrule
Pick fruit  & 426 & 30 & $60.19 \pm 1.62$ & $56.27$--$67.10$ & 7.12 \\
Hang towel  & 391 & 30 & $60.57 \pm 1.73$ & $54.70$--$67.30$ & 6.58 \\
Pour water  & 523 & 30 & $61.82 \pm 3.05$ & $51.83$--$64.94$ & 8.98 \\
Pull drawer & 439 & 30 & $60.70 \pm 1.35$ & $55.30$--$66.77$ & 7.40 \\
\midrule
\textbf{Total}
& \textbf{1,779}
& \textbf{30}
& $\mathbf{60.88 \pm 2.21}$
& $\mathbf{51.83}$--$\mathbf{67.30}$
& \textbf{30.08} \\
\bottomrule
\end{tabular}%
}
\end{table}

\begin{table*}[t]
\centering
\small
\caption{
\textbf{Evaluation-set statistics.}
The evaluation suite contains 335 episodes from
human-operated and policy-rollout distributions.
Duration is reported as mean $\pm$ standard deviation and
observed minimum--maximum in seconds.
}
\label{tab:app_evaluation_set_statistics}
\setlength{\tabcolsep}{5pt}
\resizebox{\textwidth}{!}{%
\begin{tabular}{lllrrrrr}
\toprule
Embodiment
& Action source
& \makecell[l]{Task\\distribution}
& Episodes
& FPS
& Mean $\pm$ SD (s)
& Range (s)
& Hours \\
\midrule

AgiBot G2
& Human-operated
& \makecell[l]{Pick (25), Hang (25),\\Pour (25), Pull (25)}
& 100 & 30
& $118.70 \pm 3.88$
& $108.27$--$130.47$
& 3.30 \\

AgiBot G2
& Policy-rollout
& \makecell[l]{Pick (20), Hang (20),\\Pour (20)}
& 60 & 30
& $22.59 \pm 13.01$
& $4.93$--$60.00$
& 0.38 \\

YAM
& Human-operated
& \makecell[l]{Pick (25), Hang (25),\\Pull (25)}
& 75 & 30
& $122.90 \pm 4.80$
& $110.00$--$138.77$
& 2.56 \\

Human hand
& Human-operated
& \makecell[l]{Pick (25), Hang (25),\\Pour (25), Pull (25)}
& 100 & 15
& $125.35 \pm 2.28$
& $121.40$--$134.53$
& 3.48 \\

\midrule
\textbf{Total}
& --
& \makecell[l]{Pick (95), Hang (95),\\Pour (70), Pull (75)}
& \textbf{335}
& --
& $\mathbf{104.41 \pm 38.85}$
& $\mathbf{4.93}$--$\mathbf{138.77}$
& \textbf{9.72} \\
\bottomrule
\end{tabular}%
}
\end{table*}

\subsection{Human-Operated Distribution}
\label{sec:appendix_human_distribution}
We collect held-out human-operated evaluation episodes for the AgiBot G2, YAM, and human-hand embodiments. The robot trajectories are produced through teleoperation, while the human-hand trajectories are produced through direct human execution. As shown in Table~\ref{tab:app_evaluation_set_statistics}, the AgiBot G2 and human-hand splits each contain 100 episodes, with 25 episodes per task. The YAM split contains 75 episodes, with 25 episodes each for \textit{pick fruit}, \textit{hang towel}, and \textit{pull drawer}. Each episode lasts approximately two
minutes and contains multiple consecutive interaction
segments.

For AgiBot G2, the evaluation split expands the scene
variations used during training. It includes held-out fruit instances with new colors and sizes, as well as new container types, towel colors, liquid colors, and drawer configurations. We additionally introduce unseen tablecloth colors and vary the color, angle, and position of scene illumination. The training and evaluation splits therefore share the same task semantics and retain some overlapping appearance conditions while holding out selected objects, scene
configurations, and visual conditions. This design evaluates whether an ACWM generalizes learned action--outcome relationships to new visual configurations instead of reproducing memorized trajectories.

\subsection{Policy-Rollout Distribution}
\label{sec:appendix_policy_distribution}
We additionally collect a policy-rollout evaluation split on the physical AgiBot G2 robot. A $\pi_{0.5}$
policy~\citep{intelligence2025pi05} trained using our
curated training data interacts with the real robot in
closed loop. We record the resulting observations and
low-level actions for evaluation. This split contains 60 episodes, with 20 episodes each for \textit{pick fruit}, \textit{hang towel}, and \textit{pour water}.

\section{Additional Details on the Annotation Pipeline}
\label{app:annotation_pipeline}

\subsection{Physical Criteria Annotation and Evaluation}
\label{app:physical_prompting}

We formulate physical-criteria annotation and evaluation as a criteria-authoring and criteria-judging procedure. Let $v_0=\texttt{top\_head}$ denote the primary camera view. In the criteria authoring stage, GPT-5.5~\citep{openai2026gpt54} receives the episode's initial observation $O_{i,0}^{v_0}$ and the outcome keyframe $K_{i,n}^{v_0}$. The initial observation establishes a canonical inventory of the objects and their visual identities throughout the episode, while the outcome keyframe determines the current physical state. GPT-5.5 generates a set of visually verifiable criteria $\mathcal{C}_{i,n}=\left\{c_{i,n,j}\right\}_{j=1}^{J_{i,n}}$, where $J_{i,n}$ is the number of physical criteria of the segment $n$ of episode $i$. A revision pass normalizes entity names against the inventory created from the initial observation without changing the states asserted by the criteria. During the judging stage, GPT-5.5 receives the same initial-observation reference, a predicted outcome keyframe $\hat{K}_{i,n}^{v_0}$, the canonical object inventory, and $\mathcal{C}_{i,n}$. It answers each criterion with a binary score $y_{i,n,j}\in\{0,1\}$, where $y_{i,n,j}=1$ indicates that criterion $c_{i,n,j}$ is visibly satisfied in the predicted outcome keyframe. We use separate task-specific prompts since the relevant entities, physical states, and sources of visual ambiguity differ substantially across interaction dynamics. The complete authoring and judging prompts are shown in the boxes below.

\begin{promptbox}{Pick Fruit: Physical Criteria Authoring and Judging}[AUTHORING SYSTEM PROMPT]

You are an expert annotator for a robot manipulation benchmark.
SETTING: A robot with two grippers (hands) performs a fruit pick-and-place task. It picks up a fruit from a table, or a basket / fruit plate, and places it into a basket / fruit plate, or puts it down on the table.

YOU ARE GIVEN TWO IMAGES:
- Image 1 is the episode's FIRST frame: a clear, mostly unoccluded view. Use it to establish the GROUND TRUTH set of objects in the scene -- every fruit's type/color and every basket/plate's color, style, and location.
- Image 2 is the KEY FRAME you must annotate.

IDENTITY RULE (very important): in the key frame a fruit or container can look different because of occlusion, motion blur, or viewpoint -- e.g. a mango may look like a banana or an orange. ALWAYS take the identity (type/color) of every fruit and container from Image 1. If something in the key frame roughly matches an object from Image 1, treat it as that same object. NEVER introduce a fruit or container that is not present in Image 1.

YOUR JOB: Write the `success_criteria' for the KEY FRAME (Image 2) -- a list of concrete, visually verifiable conditions that hold in Image 2.

HOW TO WRITE GOOD CRITERIA:
- Keep every criterion as short and simple as possible -- one plain statement.
- Every criterion has a FRUIT (named exactly as identified in Image 1) as its subject. Do NOT write criteria about a basket/plate, the gripper, or anything else as the subject, and do not describe the container's own state.
- In the key frame each fruit is in exactly ONE of these states:
    * on the table,
    * in a specific basket / fruit plate, or
    * not visible in the key frame.
- ONLY write a criterion you are VISUALLY CERTAIN about. If a fruit is too occluded (e.g. mostly hidden by the gripper/arm) or its spatial state is otherwise ambiguous, OMIT it entirely rather than guess.
- If a fruit that exists in Image 1 is clearly absent / not visible in the key frame, say so explicitly, e.g. "The apple is not in the view." Use this only when you are sure it is not visible -- NOT when its state is merely hard to judge.
- Identify the fruit by its color/type from Image 1 (e.g. "the red apple", "the banana").
- When a fruit is in a container, name which container by its color, style, and location so different baskets/plates can be told apart. The container only locates the fruit.
- Each criterion is ONE short, self-contained English sentence.
- Write a criterion for each fruit whose state you can judge with high visual certainty. Do not pad or guess.

GOOD EXAMPLES:
- "The red apple is in the brown wicker basket on the top left."
- "The orange is in the white plate on the right."
- "The banana is on the table."
- "The mango is not in the view."

OUTPUT CONTRACT: Respond with STRICTLY a JSON array of strings and NOTHING else. No markdown, no code fences, no explanations, no object keys.
Example: ["...", "..."]

[AUTHORING USER PROMPT]

Image 1 is the episode's first frame (the reference for the identity of every fruit and container). Image 2 is the key frame to annotate.

Using Image 1 to fix the identity of each fruit and container, write the success_criteria that hold in the key frame (Image 2). Respond with ONLY a JSON array of strings.

(*@\promptdivider@*)

[JUDGING SYSTEM PROMPT]

You are an expert physical-status judge for a robot FRUIT pick-and-place benchmark. A robot moves fruits between a table and various baskets / fruit plates. You are shown ONE key frame plus a list of success criteria. Each criterion names a fruit and asserts WHERE it is: on the table, in a specific basket / fruit plate (named by color / style / location), grasped by the gripper (optionally hovering above the table or above a named container), or not visible in the frame.

Your job: for each criterion, return a binary 1 (clearly satisfied) or 0 (clearly not satisfied).

IDENTITY -- DISAMBIGUATE AGAINST THE OBJECT INVENTORY:
- You may be given TWO images: IMAGE 1 is the episode's FIRST frame and IMAGE 2 is the frame to JUDGE.
- The user message gives the COMPLETE set of fruits and containers in the scene.
- Do NOT demand a textbook shape/type match. Match each frame object to the CLOSEST inventory item, primarily by COLOR and then by rough shape / size.
- A criterion is satisfied when the object in the stated location best matches the criterion's named object among the inventory.
- Score 0 when the object best matches a different inventory item or when no matching object is present.
- Disambiguate containers by color, style, and location.

WHAT TO VERIFY:
- The fruit's location: on the table, in the named container, grasped by the gripper, or not visible.
- For grasped / hovering criteria, verify both the grasp and, when stated, whether it is above the table or the named container.
- For "not visible", verify that no matching inventory fruit is visible.

Reserve 0 for clear contradictions: a different fruit, wrong container, wrong grasp/hover location, present when it should be absent, or absent when it should be present. A shape/type mismatch alone is not a reason to fail.

Success criteria for this frame: for each, judge 1/0 whether the named fruit is in the described place. Match the object to the closest item in the inventory.

Return exactly:
{
  "aspect_scores": [0 or 1, ...],
  "reasoning": ["brief visible evidence", "..."],
  "explanation": "one sentence summary"
}

The aspect_scores array must contain exactly one integer per criterion, in the same order.
\end{promptbox}
\begin{promptbox}{Hang Towel: Physical Criteria Authoring and Judging}
[AUTHORING SYSTEM PROMPT]

You are an expert annotator for a robot manipulation benchmark.

SETTING: A robot with a gripper picks a towel up from the table and hangs / drapes it over a horizontal hanging ROD (a clothes-drying bar / pole), OR takes a towel off the rod and puts it down on the table. The scene's objects are TOWELS.

YOU ARE GIVEN TWO IMAGES:
- Image 1 is the episode's FIRST frame: a clear, mostly unoccluded view. Use it to establish the GROUND TRUTH set of towels in the scene -- every towel's COLOR (and its position, used only to tell same-colored towels apart).
- Image 2 is the KEY FRAME you must annotate.

IDENTITY RULE: in the key frame a towel can look different because of folding, draping, occlusion, the gripper, or viewpoint -- its shape is unreliable. ALWAYS take the identity (COLOR) of every towel from Image 1. If a towel in the key frame roughly matches a towel from Image 1 by color, treat it as that same towel. NEVER introduce a towel color that is not present in Image 1.

YOUR JOB: Write the `success_criteria' for the KEY FRAME (Image 2) -- a list of concrete, visually verifiable conditions that hold in Image 2.

HOW TO WRITE GOOD CRITERIA:
- Keep every criterion as short and simple as possible.
- Every criterion's SUBJECT is a TOWEL named by its COLOR. Do NOT write criteria about the rod, gripper, or table as the subject.
- When multiple towels share the same place, add each towel's RELATIVE position among those towels.
- In the key frame each towel is in exactly ONE state:
    * on the table,
    * hanging over the rod,
    * grasped by the gripper,
    * not in the view.
- ONLY write a criterion you are VISUALLY CERTAIN about. If a towel is too occluded or its state is ambiguous, OMIT it rather than guess.
- If a towel from Image 1 is clearly absent in the key frame, state that it is not in the view.
- "Hanging over the rod" means supported by or draped over the rod, not lying flat on the table.
- Each criterion is ONE short, self-contained sentence.
- Do not pad or guess.

GOOD EXAMPLES:
- "The red towel is on the table on the left."
- "The blue towel is on the table on the right."
- "The green towel is hanging over the rod on the left."
- "The yellow towel is hanging over the rod on the right."
- "The blue towel is grasped by the gripper."
- "The red towel is not in the view."

OUTPUT CONTRACT: Respond with STRICTLY a JSON array of strings and NOTHING else.
Example: ["...", "..."]

[AUTHORING USER PROMPT]

Image 1 is the episode's first frame (the reference for the COLOR identity of every towel). Image 2 is the key frame to annotate.

Using Image 1 to fix the identity of each towel, write the success_criteria that hold in the key frame (Image 2). Respond with ONLY a JSON array of strings.

(*@\promptdivider@*)

[JUDGING SYSTEM PROMPT]

You are an expert physical-status judge for a robot TOWEL hang / unhang benchmark. A robot picks a towel up from the table and hangs / drapes it over a horizontal hanging rod, or takes a towel off the rod and puts it down on the table.

You are shown one key frame plus success criteria. Each criterion names a towel by color and, when necessary, position, and asserts that it is on the table, hanging over the rod, grasped near the table, grasped near the rod, or not visible.

For each criterion, return 1 if it is clearly satisfied and 0 if it is clearly not satisfied.

IDENTITY:
- Use the first frame and complete towel inventory to resolve identity.
- Match primarily by COLOR and then by relative position.
- Do not demand a fixed towel shape, because folds, draping, occlusion, and viewpoint change its appearance.
- When two towels share a color, use relative position to distinguish them.

WHAT TO VERIFY:
- Whether the towel is on the table, hanging over the rod, grasped near the table, grasped near the rod, or absent.
- Interpret left/middle/right relative to other towels in the same place.
- "Hanging over the rod" means visibly supported by or draped over the rod.
- For a grasp criterion, verify both the grasp and, if specified, proximity to the table or rod.
- For "not visible", verify that no matching towel is visible.

Reserve 0 for clear contradictions: wrong towel, wrong location, present when it should be absent, or absent when it should be present. Shape or fold differences alone do not constitute failure.

Return exactly:
{
  "aspect_scores": [0 or 1, ...],
  "reasoning": ["brief visible evidence", "..."],
  "explanation": "one sentence summary"
}

Return exactly one score per criterion, in the same order.
\end{promptbox}
\begin{promptbox}{Pour Water: Physical Criteria Authoring and Judging}[AUTHORING SYSTEM PROMPT]

You are an expert annotator for a robot manipulation benchmark.

SETTING: A robot arm grasps a cup that holds colored water and pours the water into another cup. The scene's objects are CUPS, and the WATER is colored.

YOU ARE GIVEN TWO IMAGES:
- Image 1 is the episode's FIRST frame. Use it to establish every cup's COLOR and the COLOR of the water.
- Image 2 is the KEY FRAME you must annotate.

IDENTITY VS POSITION:
- A cup's stable identity is its COLOR. Take each cup's color and water color from Image 1.
- A cup's POSITION is dynamic. Read its current position from Image 2 rather than copying its initial position.
- Use Image 1 for identity and Image 2 for current position and state.
- If cup color cannot be reliably determined, name the cup by its relative position instead of guessing a color.

YOUR JOB: Write the `success_criteria' for the KEY FRAME.

HOW TO WRITE GOOD CRITERIA:
- First ground the key frame by comparing all cups and determining their relative arrangement.
- Write one criterion per cup.
- Each criterion describes:
    (a) the cup's pose / position, and
    (b) the water inside the cup.
- Positions are relative to other cups, using terms such as left, middle, right, front, behind, front-left, or back-right.
- Name each cup by color plus relative position when color is reliable.
- If color is unreliable, use relative position alone.
- Cup pose must be one of:
    * standing upright on the table,
    * grasped / held by the gripper,
    * tilted and pouring into another cup.
- Water content must be one of:
    * is empty,
    * contains <color> water.
- Combine cup pose and water content into one sentence.
- Only write states that are visually certain.
- Water content is binary. If colored liquid is clearly and obviously visible, write "contains <color> water"; otherwise treat the cup as empty.
- When pouring, identify both source and target cups.
- Add a spill criterion only for a CLEAR colored spill or puddle on the table.
- Do not annotate minor droplets, wet sheen, or ambiguous regions as spills.

GOOD EXAMPLES:
- "The dark gray cup in the middle is standing upright on the table and is empty."
- "The clear cup on the right is standing upright on the table and contains red water."
- "The blue cup on the left is grasped by the gripper and contains red water."
- "The blue cup on the left is tilted over the white cup on the right, pouring red water into it."
- "Red water is spilled on the table."

OUTPUT CONTRACT: Respond with STRICTLY a JSON array of strings and NOTHING else.
Example: ["...", "..."]

[AUTHORING USER PROMPT]

Image 1 is the episode's first frame and is the reference for each cup's COLOR identity, not its position. Image 2 is the key frame to annotate.

Use Image 1 only to fix cup color identity. Read each cup's CURRENT position and state from Image 2. If cup color is unclear, name it by position only. Respond with ONLY a JSON array of strings.

(*@\promptdivider@*)

[JUDGING SYSTEM PROMPT]

You are an expert physical-status judge for a robot water-pouring benchmark.

Each criterion usually describes both a cup's pose or position and the water inside that cup. For each criterion, return 1 only if every asserted component holds; otherwise return 0.

IDENTITY AND POSITION:
- Use Image 1 only to recall cup and water colors.
- Cup positions are dynamic and relative. Judge them from Image 2 by comparing cups to one another.
- Identify a cup primarily by relative position and pouring role. Cup color is only a weak secondary cue.
- Do not score 0 solely because cup color appears different under lighting, transparency, or colored liquid.
- Judge water state, cup pose, and pouring relation as the primary signals.

WHAT TO VERIFY:
- GRASP: which cup the gripper visibly contacts and holds.
- POUR RELATION: whether the source cup is tilted above and over the rim of the specified target.
- CUP POSE: upright on the table, grasped, or tilted.
- WATER CONTENT: empty or containing the stated colored water.
- SPILL: a clear colored spill or puddle on the table.

EMPTY VS CONTAINS WATER:
- If the criterion says the cup is empty, default to empty unless water is especially obvious.
- If the criterion says the cup contains water, accept it unless the cup is clearly empty.
- When liquid is clearly visible, its color must match the inventory.
- Do not infer water from droplets, residue, reflections, or ambiguous wet sheen.

For pouring, use the strongest visible cue: liquid in the target, a correct tilt-and-pour pose, a visible stream or droplets, or a clearly changed water level.

Judge cup appearance leniently, but judge the asserted state and spatial relation strictly.

Return exactly:
{
  "aspect_scores": [0 or 1, ...],
  "reasoning": ["brief visible evidence", "..."],
  "explanation": "one sentence summary"
}

Return exactly one score per criterion, in the same order.
\end{promptbox}
\begin{promptbox}{Pull Drawer: Physical Criteria Authoring and Judging}[AUTHORING SYSTEM PROMPT]

You are an expert annotator for a robot manipulation benchmark.

SETTING: A robot with two grippers OPENS or CLOSES a drawer. The drawers form a single vertical column of three drawers, and each drawer is a different color. A drawer is either OPEN or CLOSED, and an open drawer may contain an object.

YOU ARE GIVEN TWO IMAGES:
- Image 1 is the episode's FIRST frame. Use it to establish the color of each drawer and any object inside an open drawer.
- Image 2 is the KEY FRAME you must annotate.

YOUR JOB: Write the `success_criteria' for Image 2.

HOW TO WRITE GOOD CRITERIA:
- Identify each drawer by COLOR ONLY.
- Do not use top, middle, bottom, left, or right in criteria.
- If a drawer is open and empty, write: "The <color> drawer is open."
- If a drawer is open and contains an object, write two separate criteria:
    1. "The <color> drawer is open."
    2. "A <object> is inside the <color> drawer."
- Name an inside object by color plus rough kind.
- Do not over-specify its exact category.
- If more than one drawer is open, write one criterion per open drawer.
- If no drawer is open, write exactly: "All drawers are closed."
- Judge drawer open / closed state strictly.
- If the inside object is ambiguous, omit the object criterion rather than guessing.
- Each criterion must be one short, self-contained sentence.

GOOD EXAMPLES:
- "The dark green drawer is open."
- "A green fruit is inside the dark green drawer."
- "The red drawer is open."
- "All drawers are closed."

OUTPUT CONTRACT: Respond with STRICTLY a JSON array of strings and NOTHING else.
Example: ["...", "..."]

[AUTHORING USER PROMPT]

Image 1 is the episode's first frame and provides the color of each drawer and any object placed inside a drawer. Image 2 is the key frame to annotate.

Using Image 1 to fix the colors, write which colored drawer is open, what object is inside it as a separate criterion, or that all drawers are closed. Use drawer COLOR only and no position words. Respond with ONLY a JSON array of strings.

(*@\promptdivider@*)

[JUDGING SYSTEM PROMPT]

You are an expert physical-status judge for a robot drawer open / close benchmark.

The scene contains a vertical column of three differently colored drawers. An open drawer may contain an object.

For each criterion, return 1 when it is visibly satisfied and 0 when it is clearly not satisfied.

IDENTITY:
- Use Image 1 as the reference frame.
- Use the object inventory to align drawer and inside-object names.
- For an object inside a drawer, judge identity leniently by color and rough shape. Do not require an exact object category.

WHAT TO VERIFY:
- Judge drawer OPEN / CLOSED state strictly.
- Mark a drawer as open only with clear geometric evidence that the drawer body is pulled out, such as:
    * a visible inner cavity or interior volume,
    * exposed side rails or a clear gap,
    * clear forward protrusion from the cabinet frame.
- A visible handle, shadow, slight perspective difference, or gripper contact is not sufficient evidence of opening.
- If strong open cues are absent, treat the drawer as closed.
- For "All drawers are closed", return 1 when no drawer has strong open evidence.
- Evaluate drawer-state and object-inside criteria independently.
- For an inside-object criterion, require the same drawer to be visibly open and the object inside to match by color and rough shape.
- If the drawer is closed or its opening state is unclear, score the inside-object criterion as 0.

Reserve 0 for a wrong drawer state, a wrong drawer being open, a drawer being closed when it should be open, or a missing / clearly mismatched inside object.

Return exactly:
{
  "aspect_scores": [0 or 1, ...],
  "reasoning": ["brief visible evidence", "..."],
  "explanation": "one sentence summary"
}

Return exactly one score per criterion, in the same order.
\end{promptbox}

\paragraph{Ground-truth consistency check.} We verify the internal consistency of the complete authoring--judging pipeline by passing each real-world outcome keyframe used for criteria authoring to the physical-state judge. GPT-5.5 judges $100\%$ of the authored criteria as satisfied on these ground-truth keyframes. This consistency check provides high confidence that the VLM evaluator can reliably recognize the annotated physical states and apply the criteria consistently when computing \physical.

\subsection{Interaction Region Annotation}
\label{app:interaction_region_annotation}

Physical-state criteria are attached to outcome keyframes, while interaction regions are annotated frame-wise throughout each evaluation episode. We use SAM3~\citep{carion2026sam3} to generate temporal mask sequences for the manipulated objects and the robot arms, end effectors, or human hands. For camera view $v$ and source $s\in\{\mathrm{obj},\mathrm{arm}\}$, the annotation pipeline produces $L_i^{v,s}$ temporal mask tracks:
\begin{equation}
\mathcal{M}_i^{v,s}=\left\{\left(M_{i,t}^{v,s,m}\right)_{t=1}^{T_i}
\right\}_{m=1}^{L_i^{v,s}},
\label{eq:temporal_mask_tracks}
\end{equation}
where
\begin{equation}
M_{i,t}^{v,s,m}
\in
\{0,1\}^{H\times W}
\end{equation}
is the binary mask of track $m$ at frame $t$. The index $m$ identifies an individual entity track within source $s$. For example, different fruits or towels are represented by different object-mask tracks, while the left and right robot arms are represented by different arm-mask tracks. Object masks cover task-relevant entities whose states change under the actions, including fruits, cups, liquid regions, towels, drawer components, and other manipulated objects. Arm masks cover robot arms and grippers or, for the human-hand embodiment, the interacting hands. The masks are produced for every camera view. The masks for the main camera view are used for \spatial and \keyarea, whereas the masks for all camera views are used for \multiview, evaluating models that support multi-view outputs.
\section{Additional Details on \modelfull Metrics}
\label{app:metric_details}

This section provides the complete definitions and implementation details of the four evaluation dimensions introduced in Sec.~\ref{sec:metrics}.

\subsection{Physical State Correctness}
\label{app:physical_metric}

We use GPT-5.5~\citep{openai2026gpt54} as the VLM evaluator to assess the physical correctness for predicted outcome keyframes of each interaction segment. This metric is computed using the primary \texttt{top\_head} view.

\paragraph{Physical state criteria.} For interaction segment $n$ in episode $i$, we define a set of observable physical-state criteria:
\begin{equation}
\mathcal{C}_{i,n}
=
\left\{
c_{i,n,j}
\right\}_{j=1}^{J_{i,n}},
\label{eq:physical_criteria}
\end{equation}
where $J_{i,n}$ is the number of criteria associated with the segment's outcome keyframe. These criteria describe task-specific physical states, including object identity and location, grasp state, articulation state, containment, contact, and liquid transfer. Examples include ``the yellow fruit is held by the gripper'' and ``the blue towel is hanging over the rack''. As described in Sec.~\ref{app:physical_prompting}, these criteria are authored from the corresponding real-world outcome keyframe and verified through our ground-truth consistency check.

\paragraph{VLM evaluation.}
Let $v_0=\texttt{top\_head}$ denote the primary camera view.
For each interaction segment, the VLM receives:
\begin{itemize}
    \item the predicted outcome keyframe
    $\hat{K}_{i,n}^{v_0}$;
    \item the initial real observation
    $O_{i,0}^{v_0}$ as an object-identity reference;
    \item the episode-level object inventory; and
    \item the physical-state criteria
    $\mathcal{C}_{i,n}$.
\end{itemize}
We use task-specific judging prompts for \textit{pick fruit}, \textit{hang towel}, \textit{pour water}, and \textit{pull drawer}. The complete prompts are provided in Sec.~\ref{app:physical_prompting}. For each criterion $c_{i,n,j}$, the VLM returns a binary judgment $y_{i,n,j}\in\{0,1\}$, where $y_{i,n,j}=1$ indicates that the criterion is visibly satisfied in the predicted outcome keyframe.

\paragraph{Score aggregation.} The \physical score of an outcome keyframe is the mean of its criterion-level judgments:
\begin{equation}
S_{\mathrm{phys},i,n}=\frac{1}{J_{i,n}}\sum_{j=1}^{J_{i,n}}y_{i,n,j}.
\label{eq:physical_keyframe_score}
\end{equation}
The episode-level score, expressed on the common
$[0,100]$ scale, is then computed by averaging over its $N_i$ outcome keyframes:
\begin{equation}
S_{\mathrm{phys},i}=\frac{100}{N_i}\sum_{n=1}^{N_i}S_{\mathrm{phys},i,n}.
\label{eq:physical_episode_score}
\end{equation}
Then we average $S_{\mathrm{phys},i}$ over all the episodes to obtain the final score $S_{\mathrm{phys}}$. This formulation assigns equal weight to each interaction segment, irrespective of the number of criteria associated with its outcome keyframe. It also preserves partial correctness when only a subset of the physical-state criteria is satisfied.
\subsection{Interaction Spatial Alignment}
\label{app:interaction_spatial_alignment}

Interaction spatial alignment measures the spatial correspondence of manipulated objects and robot arms or human hands between real and predicted observations and calculates their spatial displacement. The metric evaluates the primary camera view
$v_0=\texttt{top\_head}$ at a temporal stride of 10 frames. For compactness, we omit the view superscript $v_0$ for masks from this view. For example:
\begin{equation}
M_{i,t}^{s,m}:=M_{i,t}^{v_0,s,m}.
\label{eq:primary_view_mask}
\end{equation}

\paragraph{Dense feature extraction.} For sampled frame $t$ of episode $i$, we first extract dense DINOv3~\citep{simeoni2025dinov3} feature maps from the real and predicted observations in the primary view:
\begin{equation}
F_{i,t}=\phi_{\mathrm{DINO}}\left(O_{i,t}^{v_0}\right),\qquad\hat{F}_{i,t}=\phi_{\mathrm{DINO}}\left(\hat{O}_{i,t}^{v_0}\right),
\end{equation}
where $\phi_{\mathrm{DINO}}$ is DINOv3 ViT-B/16 in our implementation, which produces spatial feature maps:
\begin{equation}
F_{i,t},
\hat{F}_{i,t}
\in
\mathbb{R}^{D_f\times H_p\times W_p},
\end{equation}
where $D_f$ is the feature dimension and $H_p\times W_p$ is the spatial patch grid. Each patch feature is $L_2$-normalized before matching. For source $s\in\{\mathrm{obj},\mathrm{arm}\}$, let $M_{i,t}^{s,m}\in\{0,1\}^{H\times W}$ denote the $m$-th ground-truth mask at frame $t$. Here, $m$ indexes individual mask tracks within the object or robot-arm source. Because $M_{i,t}^{s,m}$ is defined in pixel space, we resize it to the DINOv3 patch grid using nearest-neighbor interpolation:
\begin{equation}
\widetilde{M}_{i,t}^{s,m}
=
\operatorname{Resize}_{\mathrm{NN}}
\left(
M_{i,t}^{s,m},
H_p,
W_p
\right).
\end{equation}
The resized mask identifies a set of patch locations in the real observation:
\begin{equation}
\mathcal{Q}_{i,t}^{s,m}
=
\left\{
u
\;\middle|\;
\widetilde{M}_{i,t}^{s,m}(u)=1
\right\}.
\end{equation}
We use the corresponding real-observation features as template tokens:
\begin{equation}
\mathcal{Z}_{i,t}^{s,m}
=
\left\{
F_{i,t}(u)
\;\middle|\;
u\in\mathcal{Q}_{i,t}^{s,m}
\right\}.
\end{equation}
Each template token is then used to search over the entire feature map of the corresponding predicted observation. Masks containing fewer than two patch tokens are considered too small for reliable matching and are excluded.

\paragraph{Feature correspondence.}

For each template location $u\in\mathcal{Q}_{i,t}^{s,m}$, we find its nearest neighbor over all locations $u'$ in the complete predicted feature map:
\begin{equation}
\hat{u}
=
\argmax_{u'}
\operatorname{sim}
\left(
F_{i,t}(u),
\hat{F}_{i,t}(u')
\right),
\end{equation}
where cosine similarity is defined as
\begin{equation}
\operatorname{sim}(z,\hat{z})
=
\frac{
z^\top\hat{z}
}{
\|z\|_2\|\hat{z}\|_2
}.
\end{equation}
A correspondence $(u,\hat{u})$ is mutual if $u$ is also the nearest template token to $\hat{u}$ in the reverse direction. We retain mutual nearest-neighbor correspondences when at least three mutual matches are available. When fewer than three mutual matches are found, we use all forward nearest-neighbor matches. The subsequent confidence score still depends on the number and ratio of mutual matches, and therefore penalizes correspondences with weak mutual support.

\paragraph{Spatial displacement.} Let $x(u)\in\mathbb{R}^{2}$ denote the center of patch $u$ in the original image coordinate system. For each retained correspondence $(u_\nu,\hat{u}_\nu)$, we compute its displacement vector
\begin{equation}
\boldsymbol{\delta}_\nu
=
x(\hat{u}_\nu)-x(u_\nu)
\end{equation}
and displacement magnitude
\begin{equation}
d_\nu
=
\left\|
\boldsymbol{\delta}_\nu
\right\|_2.
\end{equation}
The spatial displacement for mask $M_{i,t}^{s,m}$ is the median displacement across retained correspondences:
\begin{equation}
d_{i,t}^{s,m}
=
\operatorname{median}_{\nu}
d_\nu.
\label{eq:isa_mask_displacement}
\end{equation}
Using the median reduces sensitivity to isolated incorrect feature matches. 

\paragraph{Feature confidence.} A small estimated displacement is meaningful only if the underlying feature correspondences are reliable. We characterize correspondence reliability using four signals. 

First, the mean cosine similarity over the retained matches is
\begin{equation}
\bar{c}
=
\frac{1}{R}
\sum_{\nu=1}^{R}
\operatorname{sim}
\left(
F_{i,t}(u_\nu),
\hat{F}_{i,t}(\hat{u}_\nu)
\right),
\end{equation}
where $R$ is the number of retained matches. 

Second, let $c_\nu^{(1)}$ and $c_\nu^{(2)}$ denote the highest and second-highest similarities obtained for template location $u_\nu$. The matching margin is the mean difference between the two predicted similarities:
\begin{equation}
\Delta
=
\frac{1}{R}
\sum_{\nu=1}^{R}
\left(
c_\nu^{(1)}-c_\nu^{(2)}
\right).
\end{equation}
A larger margin indicates that the best correspondence is more clearly distinguished from alternative matches.

Third, the mutual-match ratio is
\begin{equation}
r_{\mathrm{mut}}
=
\frac{
R_{\mathrm{mut}}
}{
\left|
\mathcal{Q}_{i,t}^{s,m}
\right|
},
\end{equation}
where $R_{\mathrm{mut}}$ is the number of mutual matches among all template tokens. 

Finally, we measure whether the matched features agree on a common geometric translation. We estimate the median translation
\begin{equation}
\boldsymbol{\delta}_{\mathrm{med}}
=\operatorname{median}_{\nu}
\boldsymbol{\delta}_{\nu}=
\left(
\operatorname{median}_{\nu}\delta_{\nu,x},
\operatorname{median}_{\nu}\delta_{\nu,y}
\right)
\end{equation}
and define the geometric inlier ratio as
\begin{equation}
r_{\mathrm{inl}}
=
\frac{1}{R}
\sum_{\nu=1}^{R}
\mathbf{1}
\left[
\left\|
\boldsymbol{\delta}_\nu
-
\boldsymbol{\delta}_{\mathrm{med}}
\right\|_2
\leq
r_{\mathrm{geo}}
\right].
\end{equation}
We use an inlier radius of $r_{\mathrm{geo}}=32$ pixels. We combine these 4 signals into a feature-confidence score:
\begin{equation}
c_{\mathrm{feat}}
=
\sigma
\left(
b
+
w_{\cos}\bar{c}
+
w_{\Delta}\Delta
+
w_{\mathrm{mut}}r_{\mathrm{mut}}
+
w_{\mathrm{inl}}r_{\mathrm{inl}}
\right),
\label{eq:isa_feature_confidence}
\end{equation}
where $\sigma(\cdot)$ is the sigmoid function. We use
\begin{equation}
b=-2,
\quad
w_{\cos}=3,
\quad
w_{\Delta}=8,
\quad
w_{\mathrm{mut}}=1.5,
\quad
w_{\mathrm{inl}}=1.5.
\end{equation}

\paragraph{Match support.}
A high feature-confidence score may still be unreliable if it
is supported by only a small number of mutual matches. We therefore define a match-support score:
\begin{equation}
c_{\mathrm{sup}}
=
\sqrt{
\min
\left(
\frac{r_{\mathrm{mut}}}{\rho},
1
\right)
\left(
1-
\exp
\left[
-\frac{R_{\mathrm{mut}}}{\kappa}
\right]
\right)
}.
\label{eq:isa_match_support}
\end{equation}
The first term measures mutual-match coverage, while the
second measures the absolute number of mutual matches.
We use
\begin{equation}
\rho=0.3,
\qquad
\kappa=8.
\end{equation}
The final correspondence confidence is
\begin{equation}
c
=
c_{\mathrm{feat}}
c_{\mathrm{sup}}^{\,\eta},
\end{equation}
where $\eta=1$ in our implementation.

\paragraph{Alignment score.}
We map correspondence confidence to a bounded quality score:
\begin{equation}
q_c
=\operatorname{clip}
\left(
\frac{
c-c_{\mathrm{bad}}
}{
c_{\mathrm{good}}-c_{\mathrm{bad}}
},
0,
1
\right).
\end{equation}
Similarly, we map the median pixel displacement to a bounded distance score:
\begin{equation}
q_d
=
\operatorname{clip}
\left(
\frac{
d_{\mathrm{bad}}-d_{i,t}^{s,m}
}{
d_{\mathrm{bad}}-d_{\mathrm{good}}
},
0,
1
\right).
\end{equation}
The mask-level ISA score is
\begin{equation}
A_{i,t}^{s,m}
=
100
q_d
q_c^{\gamma}.
\end{equation}
The confidence exponent $\gamma>1$ prevents an ambiguous feature match from receiving a high alignment score merely because its estimated displacement is small. We use
\begin{equation}
c_{\mathrm{bad}}=0.5,
\qquad
c_{\mathrm{good}}=0.8,
\end{equation}
\begin{equation}
d_{\mathrm{good}}=30\text{ pixels},
\qquad
d_{\mathrm{bad}}=150\text{ pixels},
\qquad
\gamma=3.
\end{equation}
\paragraph{Score aggregation.}
Let $\mathcal{V}_{i,t}^{s}$ denote the valid masks for source $s$ at frame $t$. We first average the mask-level scores within each source:
\begin{equation}
A_{i,t}^{s}
=
\frac{1}{
\left|
\mathcal{V}_{i,t}^{s}
\right|
}
\sum_{m\in\mathcal{V}_{i,t}^{s}}
A_{i,t}^{s,m}.
\end{equation}
Let $\mathcal{T}_{i}^{s}$ denote the sampled frames containing at least one valid mask for source $s$.
The episode-level ISA score is
\begin{equation}
A_i
=
\frac{
\displaystyle
\sum_{s}
\sum_{t\in\mathcal{T}_{i}^{s}}
A_{i,t}^{s}
}{
\displaystyle
\sum_{s}
\left|
\mathcal{T}_{i}^{s}
\right|
}.
\end{equation}
Thus, every valid source--frame pair contributes equally. When both sources are valid at every sampled frame, this is equivalent to averaging the object and robot-arm scores per frame and then averaging over time. The benchmark-level ISA score is the mean over evaluated episodes:
\begin{equation}
A
=
\frac{1}{N}
\sum_{i=1}^{N}
A_i.
\end{equation}
We evaluate the top-head view at a temporal stride of 10 frames.
\subsection{Key Interaction Area Fidelity}
\label{app:keyarea_metric}

Let $v_0=\texttt{top\_head}$ denote the primary view. For compactness, we omit the view superscript for masks from this view:
\begin{equation}
M_{i,t}^{s,m}
:=
M_{i,t}^{v_0,s,m}.
\label{eq:keyarea_primary_view_mask}
\end{equation}

Whole-frame visual metrics can be dominated by static background regions and may therefore overlook visual errors around manipulated objects, robot arms, and contact regions. We thus evaluate visual quality specifically within these action-relevant regions.

\paragraph{Interaction-region extraction.} For frame $t$ of episode $i$, source $s\in\{\mathrm{obj},\mathrm{arm}\}$, and mask instance $m$, let $M_{i,t}^{s,m}\in\{0,1\}^{H\times W}$ denote the corresponding ground-truth segmentation mask. Object masks identify manipulated objects, while robot-arm masks identify the robot body, end effector, or human hand. To include object boundaries and nearby contact information, we dilate each mask by $\delta$ pixels:
\begin{equation}
\bar{M}_{i,t}^{s,m}
=
\operatorname{Dilate}
\left(
M_{i,t}^{s,m};
\delta
\right).
\end{equation}
We use $\delta=5$ pixels in our experiments. Masks containing fewer than 10 foreground pixels before dilation are excluded. We then extract the bounding box of the dilated mask, expand it by a padding of 10 pixels, and ensure that both its height and width are at least 64 pixels. Let $B_{i,t}^{s,m}$ denote the resulting crop operator. The cropped and background-masked ground-truth and predicted observations are
\begin{equation}
X_{i,t}^{s,m}
=
B_{i,t}^{s,m}
\left(
O_{i,t}
\odot
\bar{M}_{i,t}^{s,m}
\right),\quad
\hat{X}_{i,t}^{s,m}
=
B_{i,t}^{s,m}
\left(
\hat{O}_{i,t}
\odot
\bar{M}_{i,t}^{s,m}
\right).
\end{equation}
Pixels inside the dilated interaction region retain their RGB values, while pixels outside the region are set to zero. Cropping prevents the shared black background from dominating the perceptual measurements.

\paragraph{Masked PSNR.} We compute PSNR~\citep{huynh2008scopepsnr} directly over pixels inside the dilated mask:
\begin{equation}
p_{i,t}^{s,m}
=
10
\log_{10}
\left(
\frac{1}{
\operatorname{MSE}_{M}
\left(
\hat{O}_{i,t},
O_{i,t};
\bar{M}_{i,t}^{s,m}
\right)
}
\right).
\end{equation}
For observations normalized to $[0,1]$, the masked mean squared error is
\begin{equation}
\operatorname{MSE}_{M}
\left(
\hat{O},
O;
M
\right)
=
\frac{
\sum_{x,y}
M(x,y)
\left\|
\hat{O}(x,y)-O(x,y)
\right\|_2^2
}{
3\sum_{x,y}M(x,y)
}.
\end{equation}
This calculation excludes all pixels outside the interaction region.

\paragraph{Masked LPIPS.} We compute LPIPS~\citep{zhang2018unreasonablelpips} between the cropped and background-masked observations:
\begin{equation}
l_{i,t}^{s,m}
=
\operatorname{LPIPS}
\left(
\hat{X}_{i,t}^{s,m},
X_{i,t}^{s,m}
\right).    
\end{equation}

\paragraph{Masked V-JEPA similarity.} We additionally use V-JEPA 2 features~\citep{assran2025vjepa} to measure representation-level similarity between the same masked crops:
\begin{equation}
v_{i,t}^{s,m}
=
\cos
\left(
\phi_{\mathrm{VJEPA}}
\left(
\hat{X}_{i,t}^{s,m}
\right),
\phi_{\mathrm{VJEPA}}
\left(
X_{i,t}^{s,m}
\right)
\right).    
\end{equation}
We use the V-JEPA 2 ViT-B encoder and compute cosine similarity between the pooled representations of the real and predicted masked crops.

\paragraph{Metric normalization.} The three measurements have different ranges and directions. For a higher-is-better metric, we define
\begin{equation}
\mathcal{N}_{+}(x;a,b)
=
\operatorname{clip}
\left(
\frac{x-a}{b-a},
0,
1
\right).
\end{equation}
For a lower-is-better metric, we define
\begin{equation}
\mathcal{N}_{-}(x;a,b)
=
\operatorname{clip}
\left(
\frac{b-x}{b-a},
0,
1
\right).
\end{equation}
We use the empirical normalization ranges $[a_p,b_p]=[8,30]$ for PSNR, $[a_l,b_l]=[0.1,0.35]$ for LPIPS, and $[a_v,b_v]=[0.97,1.0]$ for V-JEPA similarity.

\paragraph{Score aggregation.} Each metric is normalized before any averaging. The quality score for mask $m$ is the geometric mean of the
three normalized measurements:
\begin{equation}
q_{i,t}^{s,m}
=
100
\left[
\mathcal{N}_{+}
\left(
p_{i,t}^{s,m};
a_p,b_p
\right)
\mathcal{N}_{-}
\left(
l_{i,t}^{s,m};
a_l,b_l
\right)
\mathcal{N}_{+}
\left(
v_{i,t}^{s,m};
a_v,b_v
\right)
\right]^{\frac{1}{3}}.
\end{equation}
The geometric mean requires the prediction to perform well across pixel-level fidelity, perceptual similarity, and representation-level similarity. A low score in any one component therefore reduces the combined mask-level score.

Let $\mathcal{M}_{i,t}^{s}$ denote the valid masks belonging to source $s$ at frame $t$. We first average the mask-level scores within each source:
\begin{equation}
q_{i,t}^{s}
=
\frac{1}{
\left|
\mathcal{M}_{i,t}^{s}
\right|
}
\sum_{m\in\mathcal{M}_{i,t}^{s}}
q_{i,t}^{s,m}.
\end{equation}
The object and robot-arm scores are then averaged with equal weight at each frame:
\begin{equation}
q_{i,t}
=
\frac{1}{2}
\left(
q_{i,t}^{\mathrm{obj}}
+
q_{i,t}^{\mathrm{arm}}
\right).
\end{equation}
Finally, the episode-level key interaction area fidelity score is the mean over its sampled frames:
\begin{equation}
Q_i
=
\frac{1}{
\left|
\mathcal{T}_i
\right|
}
\sum_{t\in\mathcal{T}_i}
q_{i,t}.
\end{equation}
This frame-first aggregation gives equal weight to the object and robot-arm regions at every time step. We evaluate the top-head view at a temporal stride of 10 frames.
\subsection{Multi-View Action Coherence}
\label{app:multiview_metric}

For models that support multi-view output, a predicted rollout may remain aligned with reality in one view while producing an incorrect robot pose or motion in another. We therefore evaluate robot-execution alignment across three synchronized camera views: $\mathcal{V}=\{\texttt{top\_head},\texttt{hand\_left},\texttt{hand\_right}\}$.

\paragraph{Per-view robot-arm alignment.} We evaluate only the robot-arm source because manipulated objects are frequently occluded, absent, or only partially visible in the auxiliary hand-camera views. Robot-arm motion provides a more consistently observable reference across the three views.

For view $v\in\mathcal{V}$, mask track $m$, and sampled frame
$t$, we compute the mask-level alignment score $A_{i,t}^{v,\mathrm{arm},m}$ using the procedure described in Sec.~\ref{app:interaction_spatial_alignment}. Let
\begin{equation}
\mathcal{I}_{i,t}^{v}
=
\left\{
m
\;\middle|\;
A_{i,t}^{v,\mathrm{arm},m}
\text{ is valid}
\right\}
\end{equation}
denote the valid robot-arm mask indices in view $v$ at frame $t$. The available per-view alignment score is
\begin{equation}
A_{i,t}^{v}
=
\frac{1}{
|\mathcal{I}_{i,t}^{v}|
}
\sum_{m\in\mathcal{I}_{i,t}^{v}}
A_{i,t}^{v,\mathrm{arm},m},
\qquad
|\mathcal{I}_{i,t}^{v}|>0.
\label{eq:multiview_per_view}
\end{equation}

\paragraph{Cross-view aggregation.}
The frame-level \multiview score is the average of the three view-level robot-arm alignment scores:
\begin{equation}
M_{i,t}
=
\frac{1}{
|\mathcal{V}|
}
\sum_{v\in\mathcal{V}}
\widetilde{A}_{i,t}^{v}.
\label{eq:multiview_frame_score}
\end{equation}
A missing or poorly aligned view directly reduces the score because every view receives equal weight. Let $\mathcal{T}_i$ denote the frame indices. The episode-level score is
\begin{equation}
S_{\mathrm{mv},i}
=
\frac{1}{
|\mathcal{T}_i|
}
\sum_{t\in\mathcal{T}_i}
M_{i,t}.
\label{eq:multiview_episode_score}
\end{equation}

\section{Training and Implementation Details}
\label{app:training_implementation_details}

\begin{table*}[t]
\centering
\caption{
\textbf{Training configurations for the evaluated ACWMs and policy.}
\textit{Pretrained embodiment(s)} denotes the robot or embodiment
domains covered by the official checkpoint used for initialization.
Batch denotes the effective global batch size across eight NVIDIA
H100 GPUs.
The reported step corresponds to the checkpoint used for evaluation.
}
\label{tab:baseline_training}
\resizebox{\textwidth}{!}{%
\begin{tabular}{lccccccc}
\toprule
Model &
Pretrained Embodiment(s) &
Step &
Batch &
Optimizer &
Peak LR &
Weight Decay &
Precision \\
\midrule

IRASim
& WidowX (Bridge)
& 100K
& 32
& AdamW
& $1{\times}10^{-4}$
& 0
& FP32 \\

Ctrl-World
& Franka Panda
& 250K
& 32
& AdamW
& $1{\times}10^{-5}$
& $1{\times}10^{-2}$
& BF16 \\

DreamDojo
& Human, Unitree G1, Fourier GR-1, AgiBot, YAM
& 100K
& 16
& Fused AdamW
& $5{\times}10^{-6}$
& $1{\times}10^{-1}$
& BF16 \\

Cosmos 3
& AgiBot, Franka Panda, Google Robot, WidowX, UMI, UR
& 200K
& 8
& AdamW
& $2{\times}10^{-5}$
& $5{\times}10^{-2}$
& BF16 \\

GE-Sim 2.0
& AgiBot G1
& 23K
& 64
& 8-bit AdamW
& $1{\times}10^{-4}$
& $5{\times}10^{-5}$
& BF16 \\

$\pi_{0.5}$ policy
& Franka Panda
& 51K
& 256
& AdamW
& $2.5{\times}10^{-5}$
& $1{\times}10^{-10}$
& BF16 \\
\bottomrule
\end{tabular}
}
\end{table*}

\begin{table*}[t]
\centering
\small
\setlength{\tabcolsep}{5pt}
\renewcommand{\arraystretch}{1.15}

\def\twolinecell#1#2{%
  \begin{tabular}[c]{@{}c@{}}
    #1\\
    #2
  \end{tabular}%
}

\caption{
\textbf{Architectural and interface configurations.}
\textit{Generated Frames} denotes the number of future frames produced
per model invocation, while \textit{History Frames} denotes the number
of previous RGB frames used as visual context.
The \textit{Output} column indicates whether the model natively
generates synchronized multi-view observations.
}
\label{tab:baseline_architecture}

\resizebox{\textwidth}{!}{%
\begin{tabular}{@{}lcccccc@{}}
\toprule
Model
& Size
& Action Conditioning
& Generated Frames
& History Frames
& Output
& Resolution \\
\midrule

IRASim
& 0.7B
& \twolinecell{Relative EEF/gripper;}
                  {frame-level AdaLN}
& 15
& 1
& Video
& $320{\times}256$ \\
\addlinespace[2pt]

Ctrl-World
& 1.5B
& \twolinecell{Absolute EEF/gripper;}
                  {frame-wise cross-attention}
& 5
& 6 (sparse)
& Multi-view video
& $384{\times}256$ \\
\addlinespace[2pt]

DreamDojo
& 2B
& \twolinecell{Block-relative robot actions;}
                  {4-action chunks via AdaLN}
& 12
& 1
& Video
& $640{\times}480$ \\
\addlinespace[2pt]

Cosmos 3
& 16B
& \twolinecell{Relative EEF/gripper/hand pose;}
                  {action-token conditioning}
& 16
& 1
& Video
& $832{\times}480$ \\
\addlinespace[2pt]

GE-Sim 2.0
& 2B
& \twolinecell{Absolute EEF/gripper;}
                  {channel-concat trajectory + ray maps}
& 25
& 4 (sparse)
& \twolinecell{Multi-view video,}
                  {proprioception, reward}
& $512{\times}384$ \\
\addlinespace[2pt]

JointControlVideo
& 14B
& \twolinecell{Absolute hand pose;}
                  {channel-concat depth-aware heatmaps}
& 80
& 1
& Video
& $832{\times}480$ \\
\addlinespace[2pt]

OSCAR
& 2B
& \twolinecell{Absolute hand pose;}
                  {channel-concat rendered skeleton}
& 80
& 1
& Video
& $640{\times}480$ \\

\bottomrule
\end{tabular}%
}
\end{table*}

We conduct training experiments on 8 NVIDIA H100 GPUs, with training configurations shown in Table~\ref{tab:baseline_training} and architectural configurations shown in Table~\ref{tab:baseline_architecture}.

\paragraph{IRASim.} We initialize IRASim-XL from the released Bridge frame-Ada checkpoint. We extend the original 7D relative single-arm end-effector representation to a 14D dual-arm action interface. Each action is embedded independently and injected into the spatial transformer blocks. Training uses a per-GPU batch size of 2 with two-step gradient accumulation, resulting in a global batch size of 32 across 8 GPUs. We use AdamW with a learning rate of $1\times10^{-4}$, zero weight decay, gradient clipping at 0.1, and an EMA decay of 0.9999. The model generates 15 future frames per autoregressive chunk at a resolution of $320\times256$, conditioned on 1 clean history frame.

\paragraph{Ctrl-World.} We extend the original 7D absolute single-arm end-effector representation to a 14D dual-arm action interface. Each action is projected by a three-layer MLP and supplied as the cross-attention conditioning of the SVD spatial transformer. Training uses a per-GPU batch size of 4 without gradient accumulation, giving a global batch size of 32 across 8 GPUs. We use AdamW with a learning rate of $1\times10^{-5}$ and bfloat16 precision. The model generates 5 future frames per autoregressive chunk at a resolution of $384\times256$, conditioned on 6 sparsely sampled history frames.

\paragraph{DreamDojo.} We initialize DreamDojo from its 2B AgiBot post-training checkpoint. We follow the official embodiment-specific action-conditioning interface. Robot actions are converted into block-relative representations, grouped into four-action chunks to match the temporal compression of the video tokenizer, and projected into the adaptive normalization
layers of the DiT backbone. Training uses a per-GPU batch size of 2, resulting in a global batch size of 16 across 8 GPUs. We use fused AdamW with a learning rate of $5\times10^{-6}$, weight decay of 0.1, and bfloat16 precision. The model generates 12 future frames per autoregressive chunk at a resolution of $640\times480$, conditioned on one preceding frame.

\paragraph{Cosmos 3.} We initialize Cosmos 3 Nano from the official 16B checkpoint. We replace the pretrained 29D AgiBot action interface with a 20D dual-arm representation comprising, for each arm, a 3D relative translation, a 6D relative rotation, and a scalar gripper state. The resulting actions are quantile-normalized and padded to the model's unified 64D action space. Training uses a per-GPU batch size of 1 without gradient accumulation, resulting in a global batch size of 8 across 8 GPUs. We use AdamW with a learning rate of $2\times10^{-5}$ and bfloat16 precision. The model generates 16 future frames per autoregressive chunk at a resolution of $832\times480$, conditioned on one preceding frame.

\paragraph{GE-Sim 2.0.} We initialize GE-Sim 2.0 from its 2B AgiBot post-training backbone, which jointly generates synchronized top-head, left-hand, and right-hand views. GE-Sim 2.0 conditions on actions geometrically. The absolute end-effector poses and gripper states of both arms are projected into every camera view to render pixel-aligned trajectory maps. The camera parameters are provided as Plücker ray maps. The ABC-130K training data contain only joint positions. We therefore recover the end-effector poses by forward kinematics on the public YAM URDF and calibrate the cameras of each episode from its videos. The wrist-camera mounts and focal lengths are fitted by aligning the robot's own fingertips in the image with their forward-kinematics positions. For the top camera, we reconstruct all views with VGGT-$\Omega$~\citep{wang2026vggt} and align the reconstructed wrist-camera centers to the calibrated ones with a similarity transform. Training uses a per-GPU batch size of 8 without gradient accumulation, resulting in a global batch size of 64 across 8 GPUs. We use 8-bit AdamW with a constant learning rate of $1\times10^{-4}$ after a 200-step linear warm-up, a weight decay of $5\times10^{-5}$, and bfloat16 precision. The model generates frames with a resolution of $512\times384$ per view and predicts 25 future frames per autoregressive chunk, conditioned on 4 sparsely sampled memory frames.


\paragraph{$\pi_{0.5}$ Policy.} We initialize $\pi_{0.5}$ from its official DROID checkpoint and fine-tune it on our AgiBot training suite using 8 NVIDIA H100 GPUs. Training uses AdamW with a global batch size of 256 and bfloat16 precision. The learning rate is linearly warmed up to
$2.5\times10^{-5}$ over the first 1K steps and subsequently decayed to $2.5\times10^{-6}$.
We use the checkpoint at 51K optimization steps to collect the policy-rollout distribution split.

\section{Limitations and Future Work}
\label{app:limitations}


\paragraph{Dataset coverage.} \modelfull currently spans three embodiments and four manipulation tasks, covering diverse interaction dynamics and execution outcomes. We envision extending the benchmark to additional embodiments and environments.

\paragraph{Adaptation across embodiments.} The three embodiments have different levels of training support: AgiBot uses a dedicated training set, YAM relies on public data with a larger domain gap, and human-hand predictions are evaluated zero-shot. To account for these differences, we organize the leaderboard into three embodiment-specific tracks, enabling separate evaluation and analysis within each embodiment.

\end{document}